\documentclass[11pt]{article}
\usepackage{psfrag,graphicx,natbib,epsf,mathrsfs,euscript,amsfonts,amsmath,latexsym,amssymb,mathrsfs}
\usepackage[active]{srcltx}
\usepackage{color}

\usepackage{float}
\usepackage{multirow}

\newtheorem{fact}{Fact}
\newtheorem{theorem}{Theorem}

\newtheorem{remark}{Remark}

\newtheorem{algorithm}{Algorithm}

\usepackage{amsmath}
\usepackage{amssymb}
\usepackage{enumerate}
\usepackage{subfig}
\usepackage{verbatim}
\usepackage{bbm}
\usepackage{color}
\usepackage{graphicx}
\usepackage{algorithm}
\usepackage[noend]{algpseudocode}
\usepackage[parfill]{parskip}

\makeatletter
\def\BState{\State\hskip-\ALG@thistlm}
\makeatother

\begin{document}
\title{A Closer Look at the Worst-case Behavior of Multi-armed Bandit Algorithms
\thanks{to appear in NeurIPS, 2021 (\textbf{spotlight}). Link: \texttt{http://www.columbia.edu/\string~ak4076/papers/NIPS-21-MAB.pdf}}
}
\author{
    Anand Kalvit\textsuperscript{\rm 1} and Assaf Zeevi\textsuperscript{\rm 2} \\
    Columbia University \\
    \texttt{\{\textsuperscript{\rm 1}akalvit22,\textsuperscript{\rm 2}assaf\}@gsb.columbia.edu}
}
\date{}
\maketitle

\begin{abstract}
One of the key drivers of complexity in the classical (stochastic) multi-armed bandit (MAB) problem is the difference between mean rewards in the top two arms, also known as the instance gap. The celebrated Upper Confidence Bound (UCB) policy is among the simplest optimism-based MAB algorithms that naturally {adapts} to this gap: for a horizon of play $n$, it achieves optimal $\mathcal{O}\left( \log n \right)$ regret in instances with ``large'' gaps, and a near-optimal $\mathcal{O}\left( \sqrt{n\log n} \right)$ minimax regret when the gap can be arbitrarily ``small.'' This paper provides new results on the \textit{arm-sampling} behavior of UCB, leading to several important insights. Among these, it is shown that arm-sampling rates under UCB are asymptotically deterministic, \textit{regardless} of the problem complexity. This discovery facilitates new sharp asymptotics and a novel alternative proof for the $\mathcal{O}\left( \sqrt{n\log n} \right)$ minimax regret of UCB. Furthermore, the paper also provides the first complete process-level characterization of the MAB problem under UCB in the conventional \emph{diffusion scaling}. Among other things, the ``small'' gap worst-case lens adopted in this paper also reveals profound distinctions between the behavior of UCB and Thompson Sampling, such as an \emph{incomplete learning} phenomenon characteristic of the latter.
\end{abstract}


\section{Introduction}

\textbf{Background and motivation.} The MAB paradigm provides a succinct abstraction of the quintessential \emph{exploration} vs. \textit{exploitation} trade-offs inherent in many sequential decision making problems. This has origns in clinical trial studies dating back to $1933$ \citep{thompson1933} which gave rise to the earliest known MAB heuristic, Thompson Sampling (see \cite{agrawal2012analysis}). Today, the MAB problem manifests itself in various forms with applications ranging from dynamic pricing and online auctions to packet routing, scheduling, e-commerce and matching markets among others (see \cite{bubeck2012regret} for a comprehensive survey of different formulations). In the canonical stochastic MAB problem, a decision maker (DM) pulls one of $K$ \emph{arms} sequentially at each time $t\in\{1,2,...\}$, and receives a random payoff drawn according to an arm-dependent distribution. The DM, oblivious to the statistical properties of the arms, must balance exploring new arms and exploiting the best arm played thus far in order to maximize her cumulative payoff over the horizon of play. This objective is equivalent to minimizing the \textit{regret} relative to an oracle with perfect ex ante knowledge of the optimal arm (the one with the highest mean reward). The classical stochastic MAB problem is fully specified by the tuple $\left( \left(\mathcal{P}_i\right)_{1 \leqslant i \leqslant K}, n \right)$, where $\mathcal{P}_i$ denotes the distribution of rewards associated with the $i^{\text{th}}$ arm, and $n$ the horizon of play. 

The statistical complexity of regret minimization in the stochastic MAB problem is governed by a key primitive called the \emph{gap}, denoted by $\Delta$, which accounts for the difference between the top two arm mean rewards in the problem. For a ``well-separated'' or ``large gap'' instance, i.e., a fixed $\Delta$ bounded away from $0$, the seminal paper \cite{lai} showed that the order of the smallest achievable regret is logarithmic in the horizon. There has been a plethora of subsequent work involving algorithms which can be fine-tuned to achieve a regret arbitrarily close to the optimal rate discovered in \cite{lai} (see \cite{audibert2009exploration,garivier2016etc,garivier2011kl,agrawal2017near, moss}, etc., for a few notable examples). On the other hand, no algorithm can achieve an expected regret smaller than $C\sqrt{n}$ for a fixed $n$ (the constant hides dependence on the number of arms) \emph{uniformly} over all problem instances (also called \emph{minimax} regret); see, e.g., \cite{lattimore2020bandit}, Chapter~15. The \emph{saddle-point} in this minimax formulation occurs at a gap that satisfies $\Delta \asymp {1}/{\sqrt{n}}$. This has a natural interpretation: approximately $1/\Delta^2$ samples are required to distinguish between two distributions with means separated by $\Delta$; at the ${1}/{\sqrt{n}}$-scale, it becomes statistically impossible to distinguish between samples from the top two arms within $n$ rounds of play. If the gap is smaller, despite the increased difficulty in the hypothesis test, the problem becomes ``easier'' from a regret perspective. Thus, $\Delta\asymp {1}/{\sqrt{n}}$ is the statistically ``hardest'' scale for regret minimization. A number of popular algorithms achieve the $\sqrt{n}$ minimax-optimal rate (modulo constants), see, e.g., \cite{moss,agrawal2017near}, and many more do this within poly-logarithmic factors in $n$. Many of these are variations of the celebrated upper confidence bound algorithms, e.g., UCB1 \citep{auer2002}, that achieve a minimax regret of $\mathcal{O}\left( \sqrt{n\log n} \right)$, and at the same time also deliver the logarithmic regret achievable in the instance-dependent setting of \cite{lai}.

A major driver of the regret performance of an algorithm is its arm-sampling characteristics. For example, in the instance-dependent (large gap) setting, optimal regret guarantees imply that the fraction of time the optimal arm(s) are played approaches $1$ in probability, as $n$ grows large. However, this fails to provide any meaningful insights as to the distribution of arm-pulls for smaller gaps, e.g., the $\Delta \asymp {1}/{\sqrt{n}}$ ``small gap'' that governs the ``worst-case'' instance-independent setting. 

\begin{figure}[htbp]
\centering
\subfloat[$q=0.5$]{\includegraphics[scale=0.345]{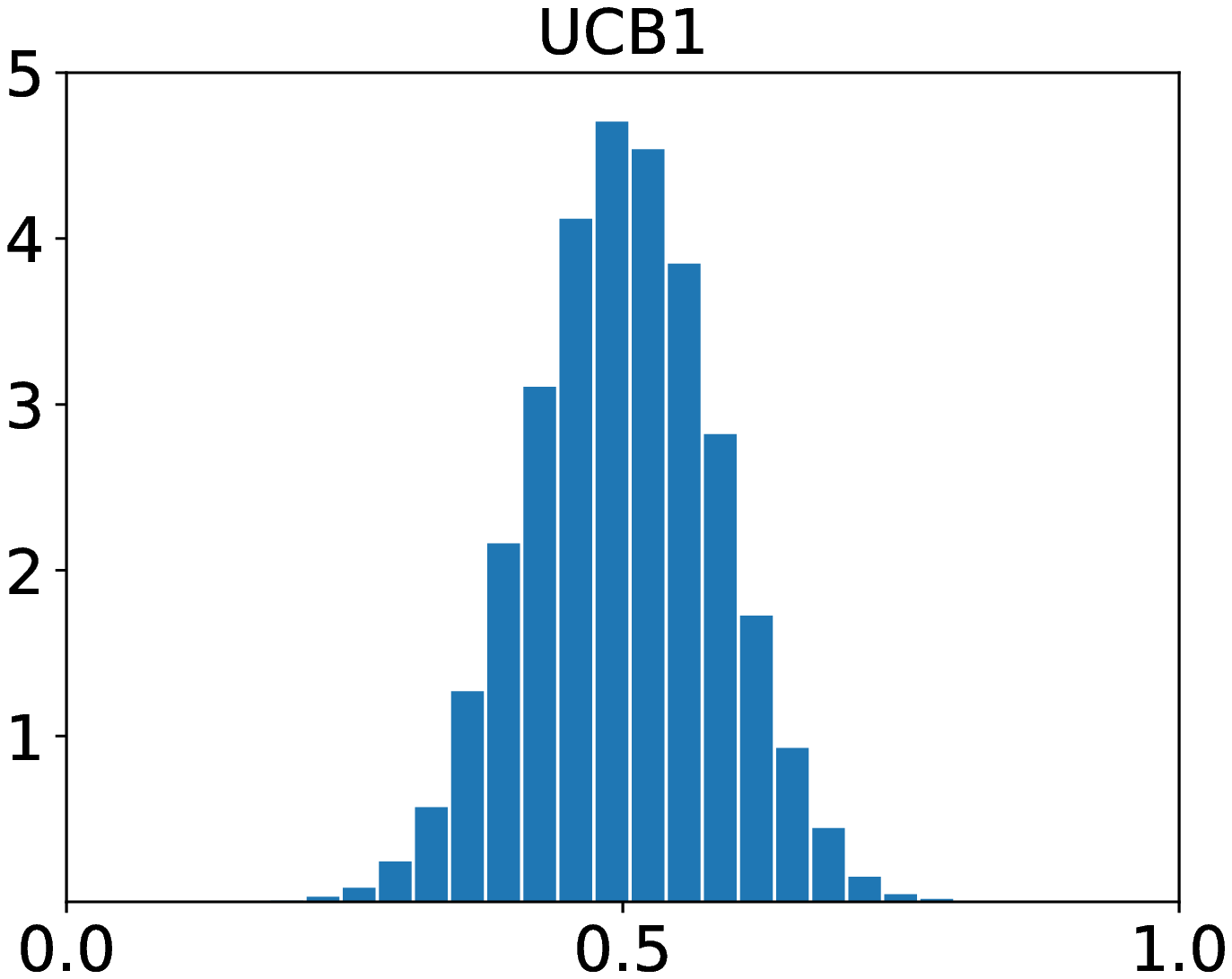}}\hfill
\subfloat[$q=0.5$]{\includegraphics[scale=0.345]{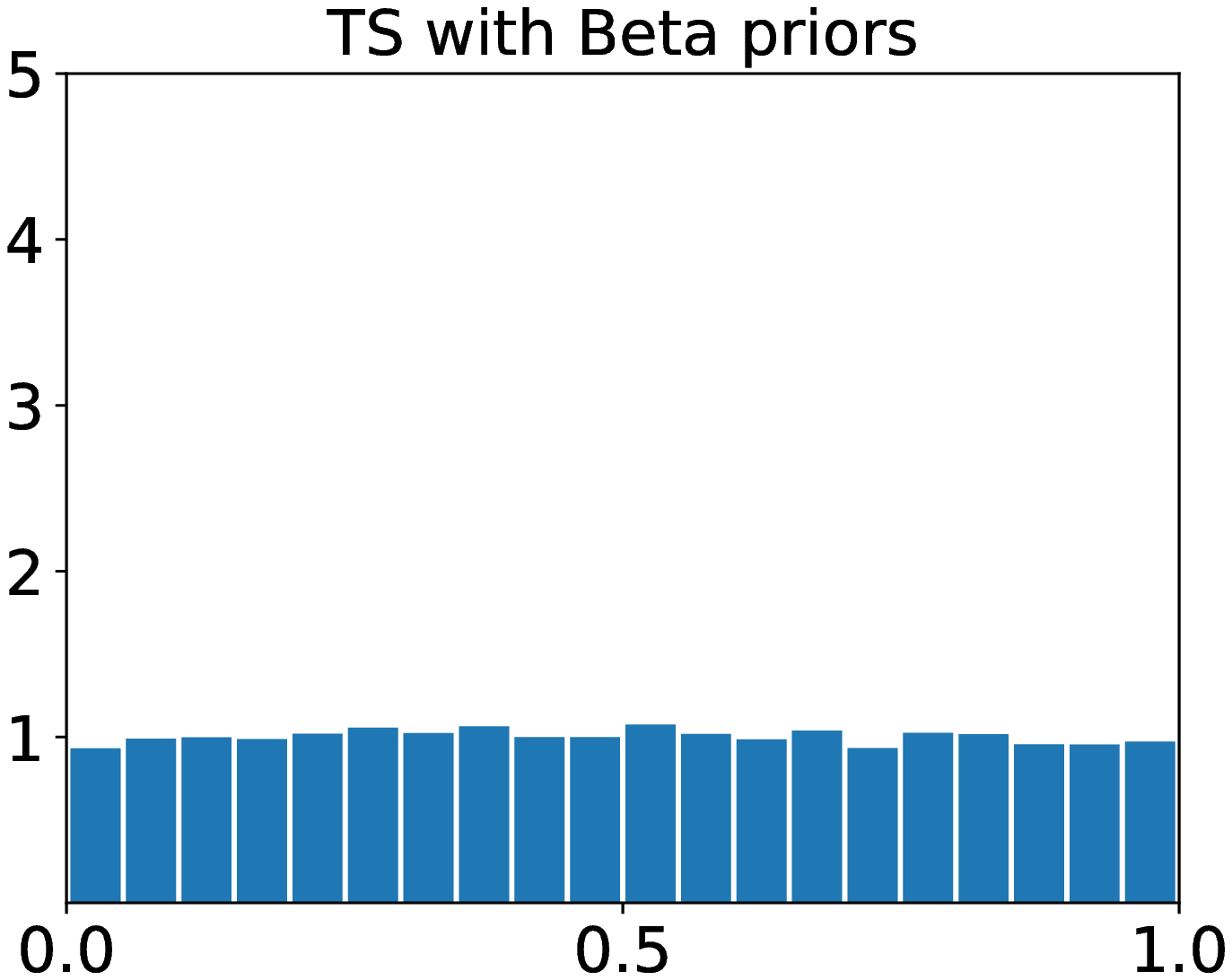}}\hfill
\subfloat[$q=0$]{\includegraphics[scale=0.345]{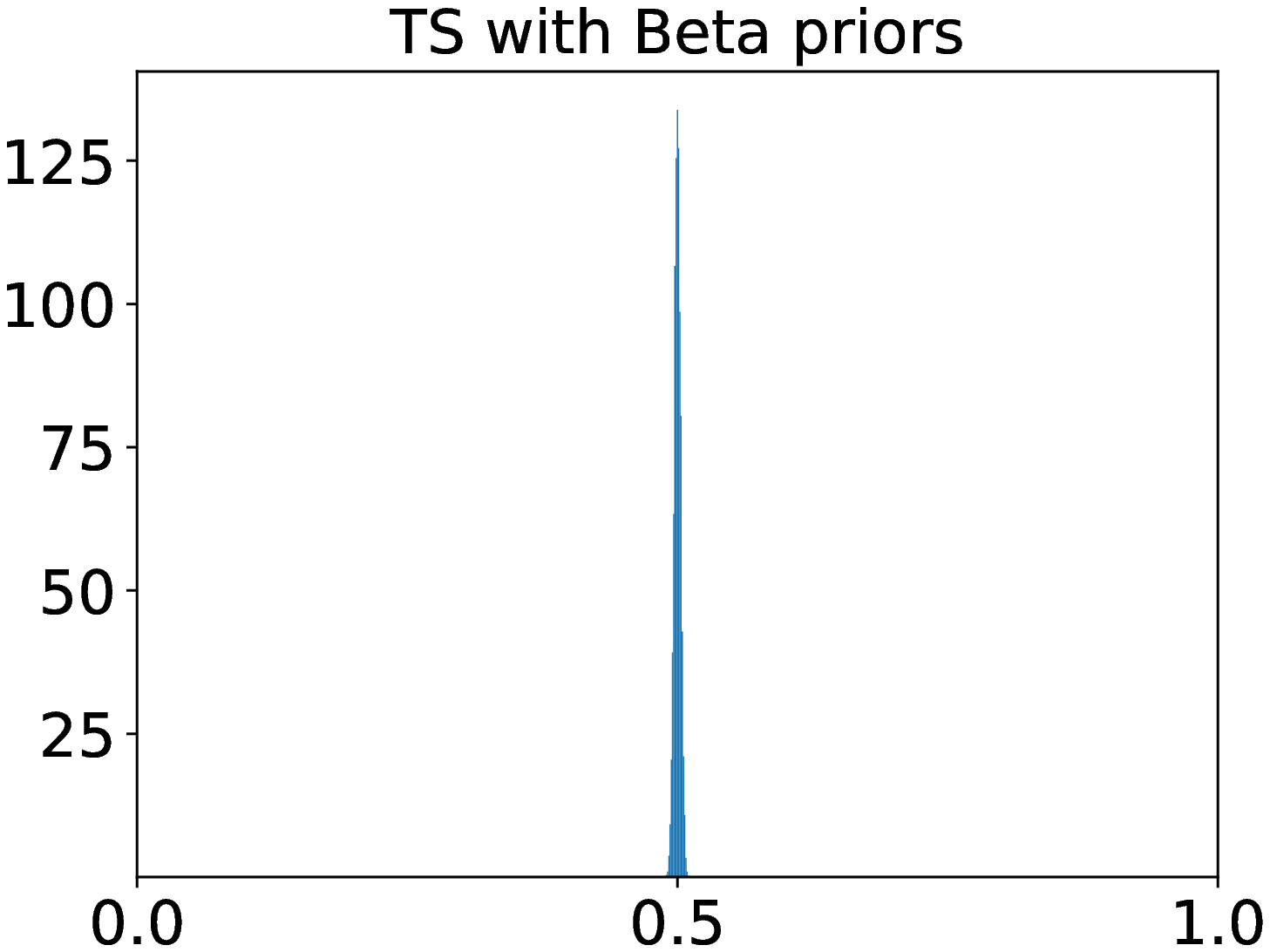}}
\caption{A two-armed bandit with arms having Bernoulli$(q)$ rewards each: Histograms show the empirical (probability) distribution of $N_1\left(n\right)/n$ for $n=\text{10,000}$ pulls, plotted using $\aleph=\text{20,000}$ experiments. Algorithms: UCB1 \citep{auer2002} and TS with Beta priors \citep{agrawal2012analysis}.}  
\label{fig:overview}
\textbf{[\emph{Concentration} under UCB and \emph{Incomplete Learning} under TS]}
\end{figure}

\textbf{An illustrative numerical example involving ``small gap.''} Consider an A/B testing problem (e.g., a vaccine clinical trial) where the experimenter is faced with two competing objectives: first, to estimate the efficacy of each alternative with the best possible precision given a budget of samples, and second, keeping the overall cost of the experiment low. This is a fundamentally hard task and algorithms incurring a low cumulative cost typically spend little time exploring sub-optimal alternatives, resulting in a degraded estimation precision (see, e.g., \cite{audibert2010best}). In other words, algorithms tailored for (cumulative) regret minimization may lack \emph{statistical power} \citep{villar2015multi}. While this trade-off is unavoidable in ``well-separated'' instances, numerical evidence suggests a plausible resolution in instances with ``small'' gaps as illustrated below. For example, such a situation might arise in trials conducted using two similarly efficacious vaccines (abstracted away as $\Delta\approx 0$). To illustrate the point more vividly, consider the case where $\Delta$ is exactly $0$ (of course, this information is not known to the experimenter). This setting is numerically illustrated in Figure~\ref{fig:overview}, which shows the empirical distribution of $N_1(n)/n$ (the fraction of time arm~1 is played until time $n$) in a two-armed bandit with $\Delta=0$, under two different algorithms.

\begin{figure}[h!]
\centering
\subfloat[Distribution of $N_1(n)/n$]{\includegraphics[scale=0.34]{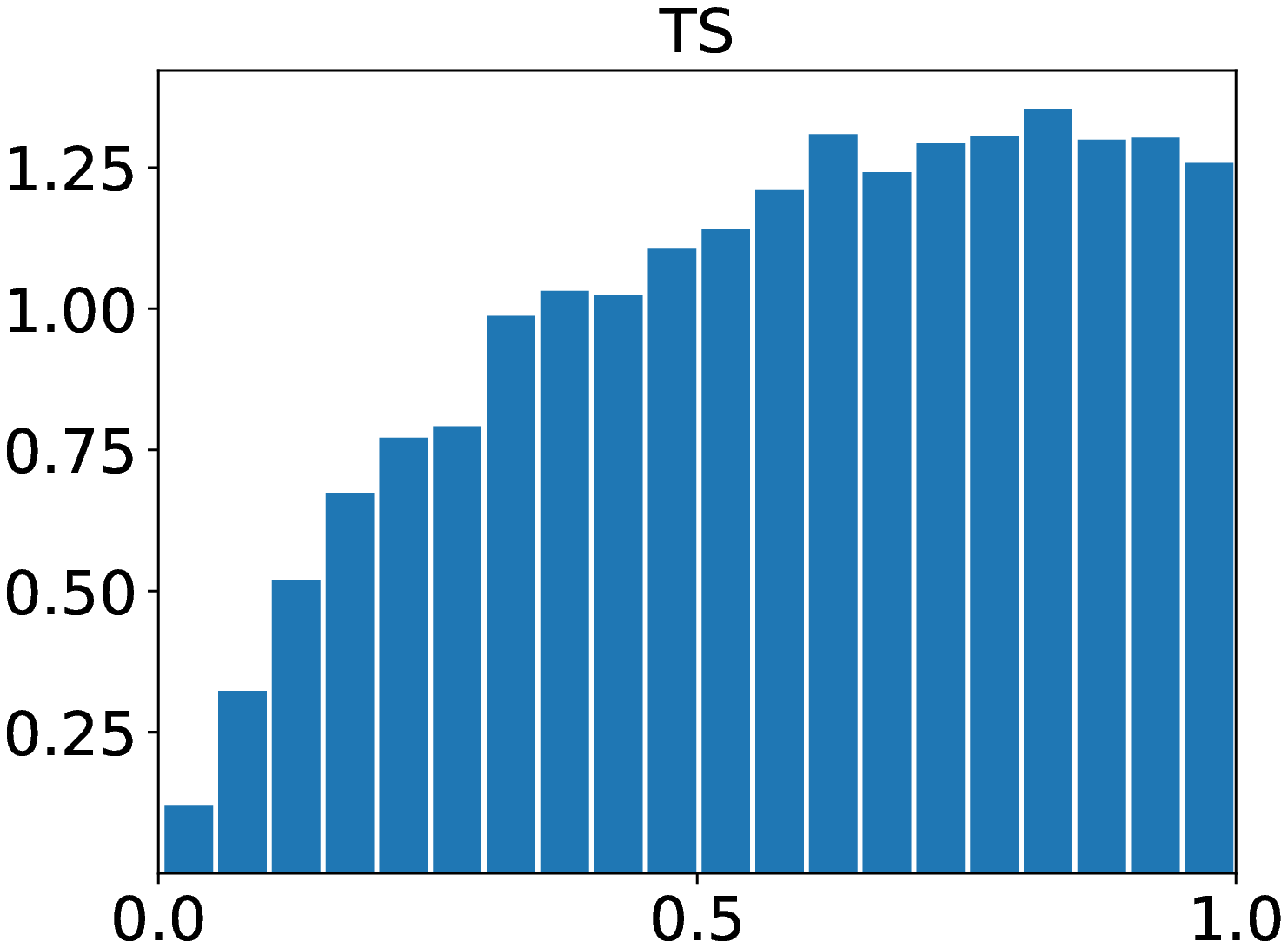}}\hfill
\subfloat[Departure from CLT]{\includegraphics[scale=0.34]{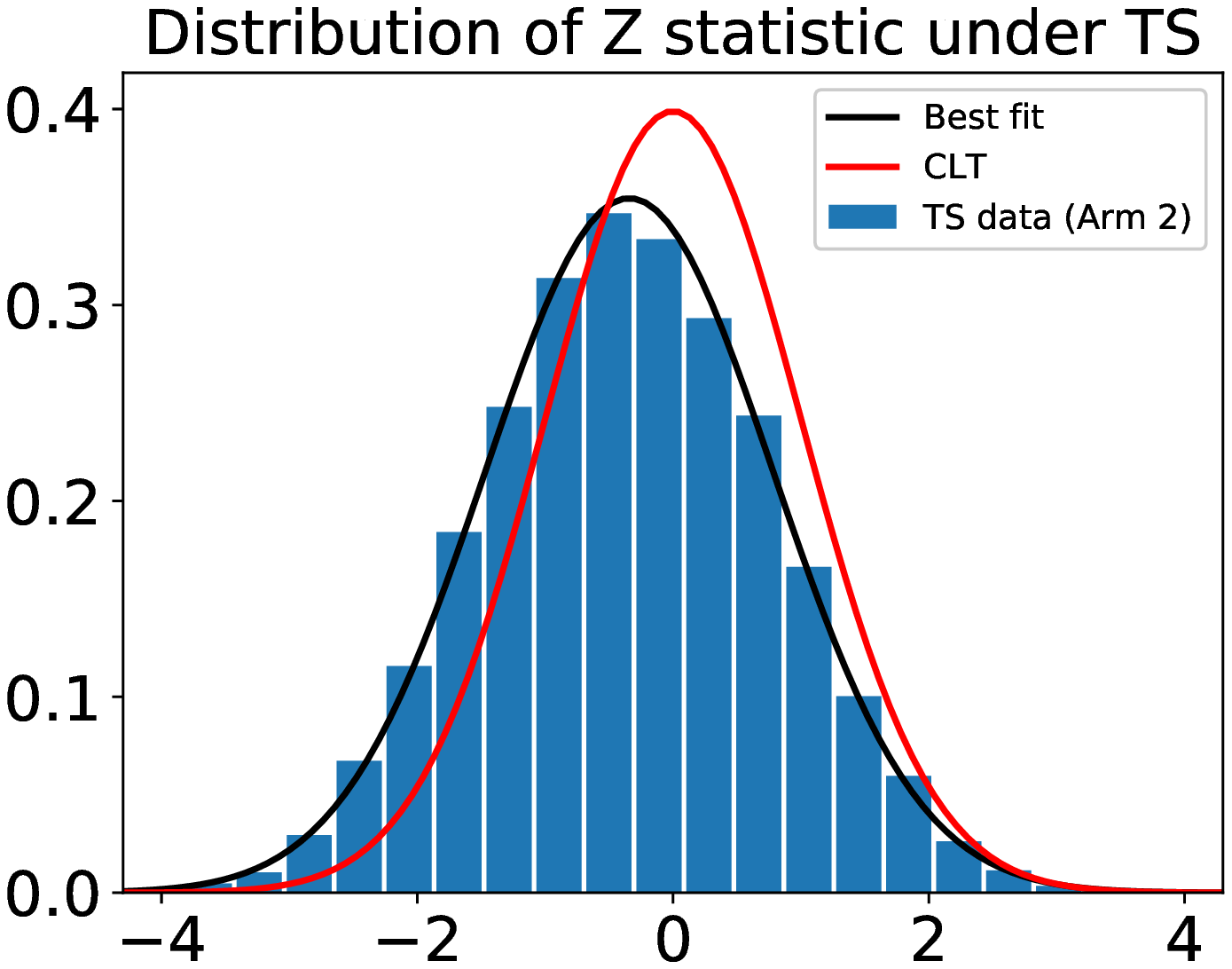}}\hfill
\subfloat[Asymptotic normality per CLT]{\includegraphics[scale=0.34]{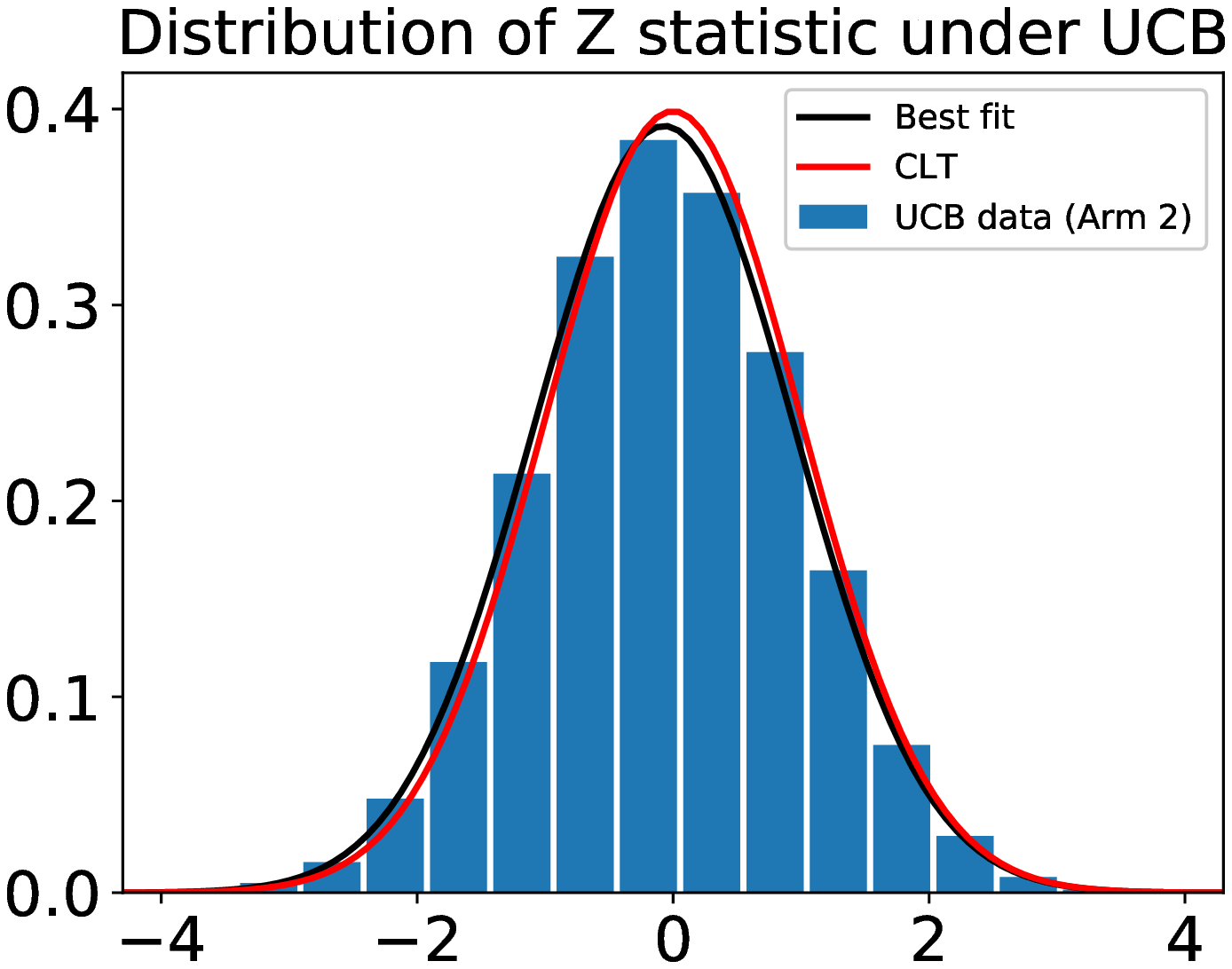}}
\caption{A two-armed bandit with $\Delta=0$: Arm~1 returns constant $0.5$, arm~2 returns Bernoulli$(0.5)$. In (a), the histogram shows the empirical (probability) distribution of $N_1(n)/n$. Algorithms: TS (Algorithm~2 in \cite{agrawal2017near}), UCB (UCB1 in \cite{auer2002}). All histograms have $n=\text{10,000}$ pulls, plotted using $\aleph=\text{20,000}$ experiments.}  
\label{fig:weak-limit4}
\textbf{[Failure of CLT under TS and asymptotic-normality under UCB]}
\end{figure}

A desirable property of the outcome in this setting is to have a \emph{linear allocation} of the sampling budget \emph{per arm} on almost every sample-path of the algorithm, as this leads to ``complete learning:'' an algorithm's ability to discern statistical indistinguishability of the arm-means, and induce a ``balanced'' allocation in that event. However, despite the simplicity of the zero-gap scenario, it is far from obvious whether the aforementioned property may be satisfied for standard bandit algorithms such as UCB and Thompson Sampling. Indeed, Figure~\ref{fig:overview} exhibits a striking difference between the two. The concentration around $1/2$ observable in Figure~\ref{fig:overview}(a) indicates that UCB results in an approximately ``balanced'' sample-split, i.e., the allocation is roughly $n/2$ per arm for large $n$ (and this is observed for ``most'' sample-paths). In fact, we will later see that the ``bell curve'' in Figure~\ref{fig:overview}(a) eventually collapses into the Dirac measure at $1/2$ (Theorem~\ref{thm:rates}). On the other hand, under Thompson Sampling, the allocation of samples across arms \emph{may} be arbitrarily ``imbalanced'' despite the arms being statistically identical, as seen in Figure~\ref{fig:overview}(b) (see, for contrast, Figure~\ref{fig:overview}(c), where the allocation is perfectly ``balanced''). Namely, the distribution of the posterior \emph{may} be such that arm~1 is allocated anywhere from almost no sampling effort all the way to receiving almost the entire sampling budget, as Figure~\ref{fig:overview}(b) suggests. Non-degeneracy of arm-sampling rates is observable also under the more widely used version of the algorithm that is based on Gaussian priors and Gaussian likelihoods (Algorithm~2 in \cite{agrawal2017near}); see Figure~\ref{fig:weak-limit4}(a). Such behavior can be detrimental for ex post causal inference in the general A/B testing context, and the vaccine testing problem referenced earlier. This is demonstrated via an instructional example of a two-armed bandit with one deterministic reference arm (aka the ``one-armed'' bandit paradigm) discussed below, and numerically illustrated later in Figure~\ref{fig:weak-limit4}.

\textbf{A numerical example illustrating inference implications.} Consider a model where arm~1 returns a constant reward of $0.5$, while arm~2 yields Bernoulli$(0.5)$ rewards. In this setup, the estimate of the gap $\Delta$ (\emph{average treatment effect} in causal inference parlance) after $n$ rounds of play is given by $\hat{\Delta} = \bar{X}_2(n)-0.5$, where $\bar{X}_2(n)$ denotes the empirical mean reward of arm~2 at time $n$. The $\mathcal{Z}$ statistic associated with this gap estimator is given by $\mathcal{Z}=2\sqrt{N_2(n)}\hat{\Delta}$, where $N_2(n)$ is the visitation count of arm~2 at time $n$. In the absence of any sample-adaptivity in the arm~2 data, results from classical statistics such as the Central Limit Theorem (CLT) would posit an asymptotically Normal distribution for $\mathcal{Z}$. However, since the algorithms that play the arms are adaptive in nature, e.g., UCB and Thompson Sampling (TS), asymptotic-normality may no longer be guaranteed. Indeed, the numerical evidence in Figure~\ref{fig:weak-limit4}(b) strongly points to a significant departure from asymptotic-normality of the $\mathcal{Z}$ statistic under TS. Non-normality of the $\mathcal{Z}$ statistic can be problematic for inferential tasks, e.g., it can lead to statistically unsupported inferences in the binary hypothesis test $\mathcal{H}_0: \Delta=0$ vs. $\mathcal{H}_1: \Delta\neq 0$ performed using confidence intervals constructed as per the conventional CLT approximation. In sharp contrast, our work shows that UCB satisfies a certain ``balanced'' sampling property (such as that in Figure~\ref{fig:overview}(a)) in instances with ``small'' gaps, formally stated as Theorem~\ref{thm:rates}, that drives the $\mathcal{Z}$ statistic towards asymptotic-normality in the aforementioned binary hypothesis testing example (asymptotic-normality being a consequence of Theorem~\ref{thm:diffusion}). Furthermore, since the $\sqrt{n}$-normalized ``stochastic'' regret (defined in \eqref{eqn:stochastic_regret} in \S\ref{sec:formulation}) equals $-\left(\sqrt{N_2(n)/(4n)}\right)\mathcal{Z}$, it follows that this too, satisfies asymptotic-normality under UCB (Theorem~\ref{thm:diffusion}, in conjunction with Theorem~\ref{thm:rates}). These properties are evident in Figure~\ref{fig:weak-limit4}(c) below, and signal reliability of ex post causal inference (under classical assumptions like validity of CLT) from ``small gap'' data collected by UCB vis-\`a-vis TS. The veracity of inference under TS may be doubtful even in the limit of infinite data, as Figure~\ref{fig:weak-limit4}(b) suggests.

\textbf{Additional applications involving ``small'' gap.} Another example reinforcing the need for a better theoretical understanding of the ``small'' gap sampling behavior of traditional bandit algorithms involves the problem of online allocation of homogeneous tasks to a pool of agents; a problem faced by many online platforms and matching markets. In such settings, \emph{fairness} and \emph{incentive} expectations on part of the platform necessitate ``similar'' agents to be routed a ``similar'' volume of assignments by the algorithm, on almost every {sample-path} (and not merely in expectation). In light of the numerical evidence reported in Figure~\ref{fig:overview} and Figure~\ref{fig:weak-limit4}(a), this can be quite sensitive to the choice of the deployed algorithm.

While traditional literature has focused primarily on the expected regret minimization problem for the stochastic MAB model, there has been recent interest in finer-grain properties of popular MAB algorithms in terms of their {arm-sampling behavior}. As already discussed, this has significance from the standpoint of ex post causal inference using data collected adaptively by bandit algorithms (see, e.g., \cite{zhang2020inference, hadad2019confidence,wager2021diffusion}, etc., and references therein for recent developments), algorithmic fairness in the broader context of fairness in machine learning (see \cite{mehrabi2019survey} for a survey), as well as novel formulations of the MAB problem such as \cite{kalvit2020finite}. Below, we discuss extant literature relevant to our line of work.

\textbf{Previous work.} The study of ``well-separated'' instances, or the \emph{large gap} regime, is supported by rich literature. For example, \cite{audibert2009exploration} provides high-probability bounds on arm-sampling rates under a parametric family of UCB algorithms. However, as the gap diminishes, leading to the so called \emph{small gap} regime, the aforementioned bounds become vacuous. The understanding of arm-sampling behavior remains relatively under-studied here even for popular algorithms such as UCB and Thompson Sampling. This regime is of special interest in that it also covers the classical \emph{diffusion scaling}\footnote{This is a standard technique for performance evaluation of stochastic systems, commonly used in the operations research and mathematics literature, see, e.g., \cite{glynn1990diffusion}.}, where $\Delta \asymp {1}/{\sqrt{n}}$, which as discussed earlier, corresponds to instances that statistically constitute the ``worst-case'' for hypothesis testing and regret minimization. Recently, a partial diffuion-limit characterization of the arm-sampling distribution under a version of Thompson Sampling with horizon-dependent prior variances\footnote{Assumed to be vanishing in $n$; standard versions of the algorithm involve fixed (positive) prior variances.}, was provided in \cite{wager2021diffusion} as a solution to a certain stochastic differential equation (SDE). The numerical solution to said SDE was observed to have a non-degenerate distribution on $[0,1]$. Similar numerical observations on non-degeneracy of the arm-sampling distribution also under standard versions of Thompson Sampling were reported in \cite{deshpande2017accurate,kalvit2020finite}, among others, albeit limited only to the special case of $\Delta=0$, and absent a theoretical explanation for the aforementioned observations. More recently, \cite{fan2021diffusion} provide a theoretical characterization of the diffusion limit also for a standard version of Thompson Sampling (one where prior variances are fixed as opposed to vanishing in $n$). However, while these results are illuminating in their own right, they provide limited insight as to the actual distribution of arm-pulls as $n\to\infty$, the primary object of focus in this paper. Thus, outside of the so called ``easy'' problems, where $\Delta$ is bounded away from $0$ by an absolute constant, theoretical understanding of the arm-sampling behavior of bandit algorithms remains an open area of research. 


\textbf{Contributions.} In this paper, we provide the first complete asymptotic characterization of arm-sampling distributions under canonical UCB (Algorithm~\ref{alg:UCB}) as a function of the gap $\Delta$ (Theorem~\ref{thm:rates}). This gives rise to a fundamental insight: arm-sampling rates are asymptotically deterministic under UCB regardless of the hardness of the instance. We also provide the first theoretical explanation for an ``incomplete learning'' phenomenon under Thompson Sampling (Algorithm~\ref{alg:TS}) alluded to in Figure~\ref{fig:overview}, as well as a sharp dichotomy between Thompson Sampling and UCB evident therein (Theorem~\ref{thm:TS-BP-00}). This result earmarks an ``instability'' of Thompson Sampling in terms of the limiting arm-sampling distribution. As a sequel to Theorem~\ref{thm:rates}, we provide the first \emph{complete} characterization of the worst-case performance of canonical UCB (Theorem~\ref{cor}). One consequence is that the $\mathcal{O}\left( \sqrt{n\log n} \right)$ minimax regret of UCB is strictly unimprovable in a precise sense. Moreover, our work also leads to the first process-level characterization of the two-armed bandit problem under canonical UCB in the classical {diffusion limit}, according to which a suitably normalized cumulative reward process converges in law to a Brownian motion with fully characterized drift and infinitesimal variance (Theorem~\ref{thm:diffusion}). To the best of our knowledge, this is the first such characterization of UCB-type algorithms. Theorem~\ref{thm:diffusion} facilitates a complete distribution-level characterization of UCB's diffusion-limit regret, thereby providing sharp insights as to the problem's minimax complexity. Such distribution-level information may also be useful for a variety of inferential tasks, e.g., construction of confidence intervals (see the binary hypothesis testing example referenced in Figure~\ref{fig:weak-limit4}(c)), among others. We believe our results may also present new design considerations, in particular, how to achieve, loosely speaking, the ``best of both worlds'' for Thompson Sampling, by addressing its ``small gap'' instability. Lastly, we note that our proof techniques are markedly different from the conventional methodology adopted in MAB literature, e.g., \cite{audibert2009exploration,bubeck2012regret,agrawal2017near}, and may be of independent interest in the study of related learning algorithms.

{\bf Organization of the paper.} A formal description of the model and the canonical UCB algorithm is provided in \S\ref{sec:formulation}. All theoretical propositions are stated in \S\ref{sec:main}, along with a high-level overview of their scope and proof sketch; detailed proofs and ancillary results are relegated to the appendices. Finally, concluding remarks and open problems are presented in \S\ref{sec:conclusion}.

\section{The model and notation}
\label{sec:formulation}

The technical development in this paper will focus on the two-armed problem purely for expositional reasons; we remark on extensions to the general $K$-armed setting at the end in \S\ref{sec:conclusion}. The two-armed setting encapsulates the core statistical complexity of the MAB problem in the ``small gap'' regime, as well as concisely highlighting the key novelties in our approach. Before describing the model formally, we introduce the following asymptotic conventions.

\textbf{Notation.} We say $f(n)=o\left( g(n) \right)$ or $g(n) = \omega \left( f(n) \right)$ if $\lim_{n\to\infty}\frac{f(n)}{g(n)} = 0$. Similarly, $f(n)=\mathcal{O}\left( g(n) \right)$ or $g(n) = \Omega \left( f(n) \right)$ if $\limsup_{n\to\infty}\left\lvert \frac{f(n)}{g(n)} \right\rvert \leqslant C$ for some constant $C$. If $f(n) = \mathcal{O}\left( \left( g(n) \right) \right)$ and $f(n) = \Omega\left( \left( g(n) \right) \right)$ hold simultaneously, we say $f(n) = \Theta\left( g(n) \right)$, or $f(n) \asymp g(n)$, and we write $f(n)\sim g(n)$ in the special case where $\lim_{n\to\infty}\frac{f(n)}{g(n)}=1$. If either sequence $f(n)$ or $g(n)$ is random, and one of the aforementioned ratio conditions holds in probability, we use the subscript $p$ with the corresponding Landau symbol. For example, $f(n)=o_p\left( g(n) \right)$ if ${f(n)}/{g(n)} \xrightarrow{p} 0$ as $n\to\infty$. Lastly, the notation `$\Rightarrow$' will be used for weak convergence.

\textbf{The model.} The arms are indexed by $\{1,2\}$. Each arm~$i\in\{1,2\}$ is characterized by a reward distribution $\mathcal{P}_i$ supported on $[0,1]$ with mean $\mu_i$. The difference between the two mean rewards, aka the \emph{gap}, is given by $\Delta = \left\lvert \mu_1 - \mu_2 \right\rvert$; as discussed earlier, this captures the \emph{hardness} of an instance. The sequence of rewards associated with the \emph{first} $m$ pulls of arm~$i$ is denoted by $\left( X_{i,j} \right)_{1 \leqslant j \leqslant m}$. The rewards are assumed to be i.i.d. in time, and independent across arms.\footnote{These assumptions can be relaxed in the spirit of \cite{auer2002}; our results also extend to sub-Gaussian rewards.} The number of pulls of arm~$i$ up to (and including) time $t$ is denoted by $N_i(t)$. A policy $\pi := \left( \pi_t \right)_{t\in\mathbb{N}}$ is an adapted sequence that prescribes pulling an arm $\pi_t\in\mathcal{S}$ at time $t$, where $\mathcal{S}$ denotes the probability simplex on $\{1,2\}$. The natural filtration at time $t$ is given by $\mathcal{F}_t := \sigma\left\lbrace \left( \pi_s \right)_{s \leqslant t},\ \left(\left( X_{i,j} \right)_{j \leqslant N_i(t)} : i=1,2\right) \right\rbrace$. The stochastic regret of policy $\pi$ after $n$ plays, denoted by $R_n^{\pi}$, is given by
\begin{align}
R_n^{\pi} := \sum_{t=1}^{n}\left[ \max\left( \mu_1, \mu_2 \right) - X_{\pi_t, N_{\pi_t}(t)} \right]. \label{eqn:stochastic_regret}
\end{align}
The decision maker is interested in the problem of minimizing the \emph{expected regret}, given by
\begin{align}
\inf_{\pi\in\Pi} \mathbb{E}R_n^{\pi}, \notag
\end{align}
where $\Pi$ is the set of policies satisfying the non-anticipation property $\pi_t : \mathcal{F}_{t-1}\to \mathcal{S},\ 1 \leqslant t \leqslant n$, and the expectation is w.r.t. the randomness in reward realizations as well as possible randomness in the policy $\pi$. In this paper, we will focus primarily on the canonical UCB policy given by Algorithm~\ref{alg:UCB} below. This policy is parameterized by an exploration coefficient $\rho$, which controls its arm-exploring rate. The standard UCB1 policy \citep{auer2002} corresponds to Algorithm~\ref{alg:UCB} with $\rho=2$; the effect of $\rho$ on the expected and high-probability regret bounds of the algorithm is well-documented in \cite{audibert2009exploration} for problems with a ``large gap.'' In what follows, $\bar{X}_i(t-1)$ denotes the empirical mean reward from arm~$i\in\{1,2\}$ at time $t-1$, i.e., $\bar{X}_i(t-1):= \frac{\sum_{j=1}^{N_i(t-1)}X_{i,j}}{N_i(t-1)}$. 

\begin{algorithm}
\caption{The canonical UCB policy for two-armed bandits.}
\label{alg:UCB}
\begin{algorithmic}[1]
\State \textbf{Input:} Exploration coefficient $\rho\in\mathbb{R}_+$.
\State At $t=1,2$, play each arm~$i\in\{1,2\}$ once.
\For{$t \in\{3,4,...\}$}
\State Play arm $\pi_t\in\arg\max_{i\in\{1,2\}} \left( \bar{X}_{i}(t-1) + \sqrt{\frac{\rho\log(t-1)}{N_i(t-1)}} \right)$.
\EndFor
\end{algorithmic}
\end{algorithm}

\section{Main results}
\label{sec:main}

Algorithm~\ref{alg:UCB} is known to achieve $\mathbb{E}R_n^{\pi} = \mathcal{O}\left( \log n \right)$ in the instance-dependent setting, and $\mathbb{E}R_n^{\pi} = \mathcal{O}\left( \sqrt{n\log n} \right)$ in the ``small gap'' minimax setting. The primary focus of this paper is on the distribution of arm-sampling rates, i.e., ${N_i(n)}/{n}$, $i\in\{1,2\}$. Our main results are split across two sub-sections; \S\ref{subsec:1} examines the behavior of UCB (Algorithm~\ref{alg:UCB}) as well as another popular bandit algorithm, Thompson Sampling (specified in Algorithm~\ref{alg:TS}). \S\ref{subsec:2} is dedicated to results on the (stochastic) regret of Algorithm~\ref{alg:UCB} under the $\Delta \asymp \sqrt{\left(\log n\right)/{n}}$ ``worst-case'' gap and the $\Delta \asymp {1}/{\sqrt{n}}$ ``diffusion-scaled'' gap. 

\subsection{Asymptotics of arm-sampling rates}
\label{subsec:1}

\begin{theorem}[Arm-sampling rates under UCB]
\label{thm:rates}

Let $i^{\ast}\in\arg\max \left\lbrace \mu_i : i=1,2 \right\rbrace$ with ties broken arbitrarily. Then, the following results hold for arm~$i^\ast$ as $n\to\infty$ under Algorithm~\ref{alg:UCB} initialized with $\rho>1$:
\begin{enumerate}[(I)]

\item ``Large gap:'' If $\Delta = \omega\left( \sqrt{\frac{\log n}{n}} \right)$, then 
$$\frac{N_{i^{\ast}}(n)}{n}\xrightarrow{p} 1.$$

\item ``Small gap:'' If $\Delta = o\left( \sqrt{\frac{\log n}{n}} \right)$, then 
$$\frac{N_{i^{\ast}}(n)}{n}\xrightarrow{p} \frac{1}{2}.$$

\item ``Moderate gap:'' If $\Delta \sim \sqrt{\frac{\theta\log n}{n}}$ for some fixed $\theta\geqslant 0$, then 
$$\frac{N_{i^{\ast}}(n)}{n}\xrightarrow{p} \lambda_\rho^{\ast}(\theta),$$
where the limit is the unique solution (in $\lambda$) to 
\begin{align}
\frac{1}{\sqrt{1-\lambda}} - \frac{1}{\sqrt{\lambda}} = \sqrt{\frac{\theta}{\rho}}, \label{eqn:limit_eqn}
\end{align}
and is monotone increasing in $\theta$, with $\lambda_\rho^{\ast}(0) = 1/2$ and $\lambda_\rho^{\ast}(\theta) \to 1$ as $\theta\to\infty$.
\end{enumerate}

\end{theorem}

\begin{remark}[Permissible values of $\rho$ in Algorithm~\ref{alg:UCB}]
\label{remark}

For $\rho>1$, the expected regret of the policy $\pi$ given by Algorithm~\ref{alg:UCB} is bounded as $\mathbb{E}R_n^{\pi} \leqslant C\rho \left(\frac{\log n}{\Delta} + \frac{\Delta}{\rho-1} \right)$ for some absolute constant $C>0$; the upper bound becomes vacuous for $\rho \leqslant 1$ (see \cite{audibert2009exploration}, Theorem~7). We therefore restrict Theorem~\ref{thm:rates} to $\rho>1$ to ensure that $\mathbb{E}R_n^{\pi}$ remains non-trivially bounded for all $\Delta$.

\end{remark}

\textbf{Discussion and intuition.} Theorem~\ref{thm:rates} essentially asserts that the sampling rates $N_i(n)/n$, $i\in\{1,2\}$ are asymptotically deterministic in probability under canonical UCB; $\Delta$ only serves to determine the value of the limiting constant. The ``moderate'' gap regime offers a continuous interpolation from instances with zero gaps to instances with ``large'' gaps as $\theta$ sweeps over $\mathbb{R}_+$ in that $\lambda_\rho^{\ast}(\theta)$ increases monotonically from $1/2$ at $\theta=0$ to $1$ at $\theta=\infty$, consistent with intuition. The special case of $\theta=0$ is numerically illustrated in Figure~\ref{fig:overview}(a). The tails of $N_{i^\ast}(n)/n$ decay polynomially fast near the end points of the interval $[0,1]$ with the best possible rate approaching $\mathcal{O}\left( n^{-3} \right)$, occurring for $\theta=0$. However, as $N_{i^\ast}(n)/n$ approaches its limit, convergence becomes slower and is dominated by fatter $\Theta\left( \sqrt{\frac{\log\log n}{\log n}} \right)$ tails. The behavior of $N_{i^\ast}(n)/n$ in this regime is regulated by the $\mathcal{O}\left( \sqrt{n\log\log n} \right)$ envelope of the zero-drift random walk process that drives the algorithm's ``stochastic'' regret (defined in \eqref{eqn:stochastic_regret}); for precise details, refer to the full proof in Appendix~\ref{sec:large_gap},\ref{sec:small_gap},\ref{sec:moderate_gap}. Since Theorem~\ref{thm:rates} is of fundamental importance to all forthcoming results on UCB, we provide a high-level overview of its proof below. 

\textbf{Proof sketch.} To provide the most intuitive explanation, we pivot to the special case where the arms have \emph{identical} reward distributions, and in particular, $\Delta=0$. The natural candidate then for the limit of the empirical sampling rate is $1/2$. On a high level, the proof relies on polynomially decaying bounds in $n$ for $\epsilon$-deviations of the form $\mathbb{P}\left( \left\lvert \frac{N_{1}(n)}{n} - \frac{1}{2} \right\rvert \geqslant \epsilon \right)$ derived using the standard trick for bounding the number of pulls of any arm on a given sample-path, to wit, for any $u,n\in\mathbb{N}$, $N_{1}(n)$ can be bounded above by $u + \sum_{t=u+1}^{n}\mathbbm{1}\left\lbrace \pi_t = 1,\ N_{1}(t-1) \geqslant u \right\rbrace$, \emph{path-wise}. Setting $u=\left\lceil \left( 1/2 + \epsilon \right)n \right\rceil$ in this expression, one can subsequently show via an analysis involving careful use of the policy structure together with appropriate Chernoff bounds that with high probability (approaching $1$ as $n\to\infty$), $N_1(n)/n \leqslant 1/2 + \varepsilon_\rho$ for some $\varepsilon_\rho\in\left( 0, 1/2 \right)$ that depends only on $\rho$. An identical result would naturally hold also for the other arm by symmetry arguments, and therefore we arrive at a meta-conclusion that $N_i(n)/n \geqslant 1/2 - \varepsilon_\rho > 0$ for both arms~$i\in\{1,2\}$ with high probability (approaching $1$ as $n\to\infty$). It is noteworthy that said conclusion cannot be arrived at for an arbitrary $\epsilon>0$ (in place of $\varepsilon_\rho$) since the polynomial upper bounds on $\mathbb{P}\left( \left\lvert \frac{N_{1}(n)}{n} - \frac{1}{2} \right\rvert \geqslant \epsilon \right)$ derived using the aforementioned path-wise upper bound on $N_1(n)$, become vacuous if $u$ is set ``too close'' to $n/2$, i.e., if $\epsilon$ is ``near'' $0$. Extension to the full generality of $\epsilon>0$ is achieved via a refined analysis that uses the Law of the Iterated Logarithm (see \cite{durrett2019probability}, Theorem~8.5.2), together with the previous meta-conclusion, to obtain fatter $\mathcal{O}\left( \sqrt{\frac{\log\log n}{\log n}} \right)$ tail bounds when $\epsilon$ is near $0$. Here, it is imperative to point out that the ``$\log n$'' appearing in the denominator is essentially from the $\sqrt{\rho\log t}$ optimistic bias term of UCB (see Algorithm~\ref{alg:UCB}), and therefore the convergence will, as such, hold also for other variations of the policy that have ``less aggressive'' $\omega\left( \log\log t \right)$ exploration functions vis-\`a-vis $\log t$. However, this will be achieved at the expense of the policy's expected regret performance, as noted in Remark~\ref{remark}. We also note that the extremely slow $\mathcal{O}\left( \sqrt{\frac{\log\log n}{\log n}} \right)$ convergence is not an artifact of our analysis, but in fact, supported by the numerical evidence in Figure~\ref{fig:overview}(a), suggestive of a plausible non-convergence (to $1/2$) in the limit. We believe such observations in previous works likely led to incorrect folk conjectures ruling out the existence of a deterministic limit under UCB \`a la Theorem~\ref{thm:rates} (see, e.g., \cite{deshpande2017accurate} and references therein). The proof for a general $\Delta$ in the ``small'' and ``moderate'' gap regimes is skeletally similar to that for $\Delta=0$, albeit guessing a candidate limit for ${N_{i^\ast}(n)}/{n}$ is non-trivial; a closed-form expression for $\lambda_\rho^{\ast}(\theta)$ is provided in Appendix~\ref{sec:expressions}. Full details of the proof of Theorem~\ref{thm:rates} are provided in Appendix~\ref{sec:large_gap},\ref{sec:small_gap},\ref{sec:moderate_gap}. {\color{white}...} \hfill $\square$ 

\begin{remark}[Possible generalizations of Theorem~\ref{thm:rates}]

A simple extension to the $K$-armed setting is stated below as Theorem~\ref{thm:zero-gap-K-arm}. The behavior of UCB policies is largely governed by their optimistic bias terms. While this paper only considers the canonical UCB policy with $\sqrt{\rho\log t}$ bias, results in the style of Theorem~\ref{thm:rates} will continue to hold also for smaller $\omega\left(\sqrt{\rho\log\log t}\right)$ bias terms, driven by the $\mathcal{O}\left( \sqrt{t\log\log t} \right)$ envelope of the ``small gap'' regret process (governed by the Law of the Iterated Logarithm), as discussed earlier. We believe this observation will be useful when examining more complicated UCB-inspired policies such as KL-UCB \citep{garivier2011kl}, \cite{dmed,imed}, etc. 

\end{remark}

\begin{theorem}[Asymptotic sampling rate of optimal arms under UCB]
\label{thm:zero-gap-K-arm}
Fix $K\in\mathbb{N}$, and consider a $K$-armed model with arms indexed by $[K]:=\{1,...,K\}$. Let $\mathcal{I}\subseteq[K]$ be the set of optimal arms, i.e., arms with mean $\max_{i\in[K]}\mu_i$. If $\mathcal{I}\neq[K]$, define $\Delta_{\min}:= \max_{i\in[K]}\mu_i - \max_{i\in[K]\backslash\mathcal{I}}\mu_i$. Then, there exists a finite $\rho_0>1$ that depends only on $\left\lvert \mathcal{I} \right\rvert$, such that the following results hold for any arm $i\in\mathcal{I}$ as $n\to\infty$ under the $K$-armed version of Algorithm~\ref{alg:UCB} initialized with $\rho \geqslant \rho_0$:
\begin{enumerate}[(I)]

\item If $\mathcal{I}=[K]$, then
\begin{align*}
\frac{N_i(n)}{n} \xrightarrow{p} \frac{1}{K}.
\end{align*}

\item If $\mathcal{I}\neq[K]$ and optimal arms are ``well-separated,'' i.e., $\Delta_{\min}=\omega\left( \sqrt{\frac{\log n}{n}} \right)$, then
\begin{align*}
\frac{N_i(n)}{n} \xrightarrow{p} \frac{1}{|\mathcal{I}|}.
\end{align*}
\end{enumerate}

\end{theorem}

\textbf{Discussion.} The main observation here is that if the set of optimal arms is ``sufficiently separated'' from the sub-optimal arms, then classical UCB policies eventually allocate the sampling effort over the set of optimal arms \emph{uniformly}, in probability. This is a desirable property to have from a fairness standpoint, and also markedly different from the instability and imbalance exhibited by Thompson Sampling in Figure~\ref{fig:overview}(b), \ref{fig:overview}(c) and \ref{fig:weak-limit4}(a). We remark that the $\rho \geqslant \rho_0$ condition is only necessary for tractability of the proof, and conjecture the result to hold, in fact, for any $\rho>1$, akin to the result for the two-armed setting (Theorem~\ref{thm:rates}). We also conjecture analogous results for ``small gap'' and ``moderate gap'' regimes, in the spirit of Theorem~\ref{thm:rates}; proofs, however, can be unwieldy in the general $K$-armed setting. The full proof of Theorem~\ref{thm:zero-gap-K-arm} is provided in Appendix~\ref{proof:K-arm}.

\textbf{What about Thompson Sampling?} Results such as those discussed above for other popular adaptive algorithms like Thompson Sampling\footnote{This is the version based on Gaussian priors and Gaussian likelihoods, not the classical version based on Beta priors and Bernoulli likelihoods which has a minimax regret of $\mathcal{O}\left( \sqrt{n\log n} \right)$ \citep{agrawal2017near}.} are only arable in ``well-separated'' instances where $N_{i^\ast}(n)/n \xrightarrow{p}1$ as $n\to\infty$ follows as a trivial consequence of its $\mathcal{O}\left( \sqrt{n} \right)$ minimax regret bound \citep{agrawal2017near}. For smaller gaps, theoretical understanding of the distribution of arm-pulls under Thompson Sampling remains largely absent even for its most widely-studied variants. In this paper, we provide a first result in this direction: Theorem~\ref{thm:TS-BP-00} formalizes a revealing observation for classical Thompson Sampling (Algorithm~\ref{alg:TS}) in instances with zero gap, and elucidates its instability in view of the numerical evidence reported in Figure~\ref{fig:overview}(b) and \ref{fig:overview}(c). This result also offers an explanation for the sharp contrast with the statistical behavior of canonical UCB (Algorithm~\ref{alg:UCB}) \`a la Theorem~\ref{thm:rates}, also evident from Figure~\ref{fig:overview}(a). In what follows, rewards are assumed to be Bernoulli, and $S_i$ (respectively $F_i$) counts the number of successes/1's (respectively failures/0's) associated with arm~$i\in\{1,2\}$.

\begin{algorithm}
\caption{Thompson Sampling for the two-armed Bernoulli bandit.}
\label{alg:TS}
\begin{algorithmic}[1]
\State \textbf{Initialize:} Number of successes (1's) and failures (0's) for each arm~$i\in\{1,2\}$, $\left( S_i, F_i \right) = (0,0)$.
\For{$t \in\{1,2,...\}$}
\State Sample for each $i\in\{1,2\}$, $\mathcal{T}_i\sim\text{Beta}\left(S_i+1, F_i+1 \right)$.
\State Play arm $\pi_t\in\arg\max_{i\in\{1,2\}} \mathcal{T}_i$ and observe reward $r_t\in\{0,1\}$.
\State Update success-failure counts: $S_{\pi_t} \gets S_{\pi_t} + r_t$, $F_{\pi_t} \gets F_{\pi_t} + 1-r_t$.
\EndFor
\end{algorithmic}
\end{algorithm}

\begin{theorem}[Incomplete learning under Thompson Sampling]
\label{thm:TS-BP-00}
In a two-armed model where both arms yield rewards distributed as Bernoulli$(q)$, the following holds under Algorithm~\ref{alg:TS} as $n\to\infty$:
\begin{enumerate}[(I)]
\item If $q=0$, then $$\frac{N_1(n)}{n} \Rightarrow \frac{1}{2}.$$
\item If $q=1$, then $$\frac{N_1(n)}{n} \Rightarrow \text{Uniform distribution on}\ [0,1].$$
\end{enumerate}
\end{theorem}

\textbf{Proof sketch.} The proof of Theorem~\ref{thm:TS-BP-00} relies on a careful application of two subtle properties of the Beta distribution (Fact~\ref{fact:1} and Fact~\ref{fact:2}), stated and proved in Appendix~\ref{sec:aux},\ref{facts}. For part (I), we invoke symmetry to deduce $\mathbb{E}N_1(n) = n/2$, and use Fact~\ref{fact:1} to show that the standard deviation of $N_1(n)$ is sub-linear in $n$, thus proving the stated assertion in (I). More elaborately, Fact~\ref{fact:1} states for the reward configuration in (I) that the probability of playing arm~1 after it has already been played $n_1$ times, and arm~2 $n_2$ times, equals $\left( n_2+1 \right)/(n_1+n_2+2)$. This probability is smaller than $1/2$ if $n_1 > n_2$, which provides an intuitive explanation for the fast convergence of $N_1(n)/n$ to $1/2$ observed in Figure~\ref{fig:overview}(c). In fact, we conjecture that the result in (I) holds also with probability $1$ based on the aforementioned ``self-balancing'' property. The conclusion in part (II) hinges on an application of Fact~\ref{fact:2} to show the \emph{stronger} result: $N_1(n)$ is \emph{uniformly distributed} over $\{0,1,...,n\}$ for any $n\in\mathbb{N}$. Contrary to Fact~\ref{fact:1}, Fact~\ref{fact:2} states that quite the opposite is true for the reward configuration in (II): the probability of playing arm~1 after it has already been played $n_1$ times, and arm~2 $n_2$ times, equals $\left( n_1+1 \right)/(n_1+n_2+2)$, which is greater than $1/2$ when $n_1 > n_2$. That is, the posterior distributions evolve in such a way that the algorithm is ``deceived'' into incorrectly believing one of the arms (arm~2 in this case) to be inferior. This leads to large sojourn times between successive visitations of arm~2 on such a sample-path, thereby resulting in a perpetual ``imbalance'' in the sample-counts. This provides an intuitive explanation for the non-degeneracy observed in Figure~\ref{fig:overview}(b) and \ref{fig:weak-limit4}(a), which additionally, also indicates that such behavior, in fact, persists also for general (non-deterministic) reward distributions, as well as under the Gaussian prior-based version of the algorithm. Full proof of Theorem~\ref{thm:TS-BP-00} is provided in Appendix~\ref{appendix:thm5}. \hfill $\square$ 

\textbf{More on ``incomplete learning.''} The zero-gap setting is a special case of the ``small gap'' regime where canonical UCB guarantees a $\left(1/2,1/2\right)$ sample-split in probability (Theorem~\ref{thm:rates}). On the other hand, Theorem~\ref{thm:TS-BP-00} suggests that second order factors such as the mean signal strength (magnitude of the mean reward) could significantly affect the nature of the resulting sample-split under Thompson Sampling. Note that even though the result only presupposes deterministic $0/1$ rewards, the aforementioned claim is, in fact, borne out by the numerical evidence in Figure~\ref{fig:overview}(b) and \ref{fig:overview}(c). The sampling distribution seemingly flattens rapidly from the Dirac measure at $1/2$ to the Uniform distribution on $[0,1]$ as the mean rewards move away from $0$. This uncertainty in the limiting sampling behavior has non-trivial implications for a  variety of application areas of such learning algorithms. For instance, a uniform distribution of arm-sampling rates on $[0,1]$ indicates that the sample-split could be arbitrarily imbalanced along a sample-path, despite, as in the setting of Theorem~\ref{thm:TS-BP-00}, the two arms being statistically identical; this phenomenon is typically referred to as ``incomplete learning'' (see \cite{rothschild1974two,mclennan1984price} for the original context). Non-degeneracy in the limiting distribution is also observable numerically up to diffusion-scale gaps of $\mathcal{O}\left( {1}/{\sqrt{n}} \right)$ under other versions of Thompson Sampling (see \cite{wager2021diffusion} for examples); our focus on the more extreme zero-gap setting simplifies the illustration of these effects.

\textbf{A brief survey of Thompson Sampling.} While extant literature does not provide any explicit result for Thompson Sampling characterizing its arm-sampling behavior in instances with ``small'' and ``moderate'' gaps, there has been recent work on its analysis in the $\Delta\asymp 1/\sqrt{n}$ regime under what is known as the \emph{diffusion approximation} lens (see \cite{wager2021diffusion,fan2021diffusion}). Cited works, however, study Thompson Sampling primarily under the assumption that the prior variance associated with the mean reward of any arm vanishes in the horizon of play at an ``appropriate'' rate.\footnote{The only result applicable to the case of non-vanishing prior variances is Theorem~4.2 of \cite{fan2021diffusion}.} Such a scaling, however, is not ideal for optimal regret performance in typical MAB instances. Indeed, the versions of Thompson Sampling optimized for regret performance in the MAB problem are based on fixed (non-vanishing) prior variances, e.g., Algorithm~\ref{alg:TS} and its Gaussian prior-based counterpart \citep{agrawal2017near}. On a high level, \cite{wager2021diffusion,fan2021diffusion} establish that as $n\to\infty$, the pre-limit $\left( N_i(nt)/n \right)_{t\in[0,1]}$ under Thompson Sampling converges weakly to a ``diffusion-limit'' stochastic process on $t\in[0,1]$. Recall from earlier discussion that $\Delta \asymp 1/\sqrt{n}$ is covered under the ``small gap'' regime; consequently, it follows from Theorem~\ref{thm:rates} that the analogous limit for UCB is, in fact, the deterministic process $t/2$. In sharp contrast, the diffusion-limit process under Thompson Sampling may at best be characterizable only as a solution (possibly non-unique) to an appropriate stochastic differential equation or ordinary differential equation driven by a suitably (random) time-changed Brownian motion. Consequently, the diffusion limit under Thompson Sampling is more difficult to interpret vis-\`a-vis UCB, and it is much harder to obtain lucid insights as to the nature of the distribution of $N_i(n)/n$ as $n\to\infty$.

\subsection{Beyond arm-sampling rates}
\label{subsec:2}

This part of the paper is dedicated to a more fine-grained analysis of the ``stochastic'' regret of UCB (defined in \eqref{eqn:stochastic_regret} in \S\ref{sec:formulation}). Results are largely facilitated by insights on the sampling behavior of UCB in instances with ``small'' gaps, attributable to Theorem~\ref{thm:rates}; however, we believe they are of interest in their own right. We commence with an application of Theorem~\ref{thm:rates} which provides the first complete characterization of the worst-case (minimax) performance of UCB. A full diffusion-limit characterization of the two-armed bandit problem under UCB is provided thereafter in Theorem~\ref{thm:diffusion}.

\begin{theorem}[Minimax regret complexity of UCB]
\label{cor}
In the ``moderate gap'' regime referenced in Theorem~\ref{thm:rates} where $\Delta\sim\sqrt{\frac{\theta\log n}{n}}$, the (stochastic) regret of the policy $\pi$ given by Algorithm~\ref{alg:UCB} initialized with $\rho>1$ satisfies as $n\to\infty$,
\begin{align}
\frac{R_n^\pi}{\sqrt{n\log n}} \Rightarrow \sqrt{\theta}\left( 1 - \lambda_\rho^{\ast}(\theta) \right) =: h_\rho(\theta), \label{eqn:worst-regret}
\end{align}
where $\lambda_\rho^{\ast}(\theta)$ is the (unique) solution to \eqref{eqn:limit_eqn}.
\end{theorem}

To the best of our knowledge, this is the first statement predicating an \emph{algorithm-specific} achievability result (sharp asymptotic for regret) that is distinct from the general ${\Omega}\left(\sqrt{n}\right)$ information-theoretic lower bound by a horizon-dependent factor.\footnote{The closest being \cite{agrawal2017near}, which established matching $\Theta\left( \sqrt{Kn\log K} \right)$ upper and lower bounds for the minimax \emph{expected} regret of the Gaussian prior-based Thompson Sampling algorithm in the $K$-armed problem.}

\textbf{Discussion.} The behavior of $h_\rho(\theta)$ is numerically illustrated in Figure~\ref{fig} below.
\begin{figure}[h!]
\centering
\includegraphics[scale=0.25]{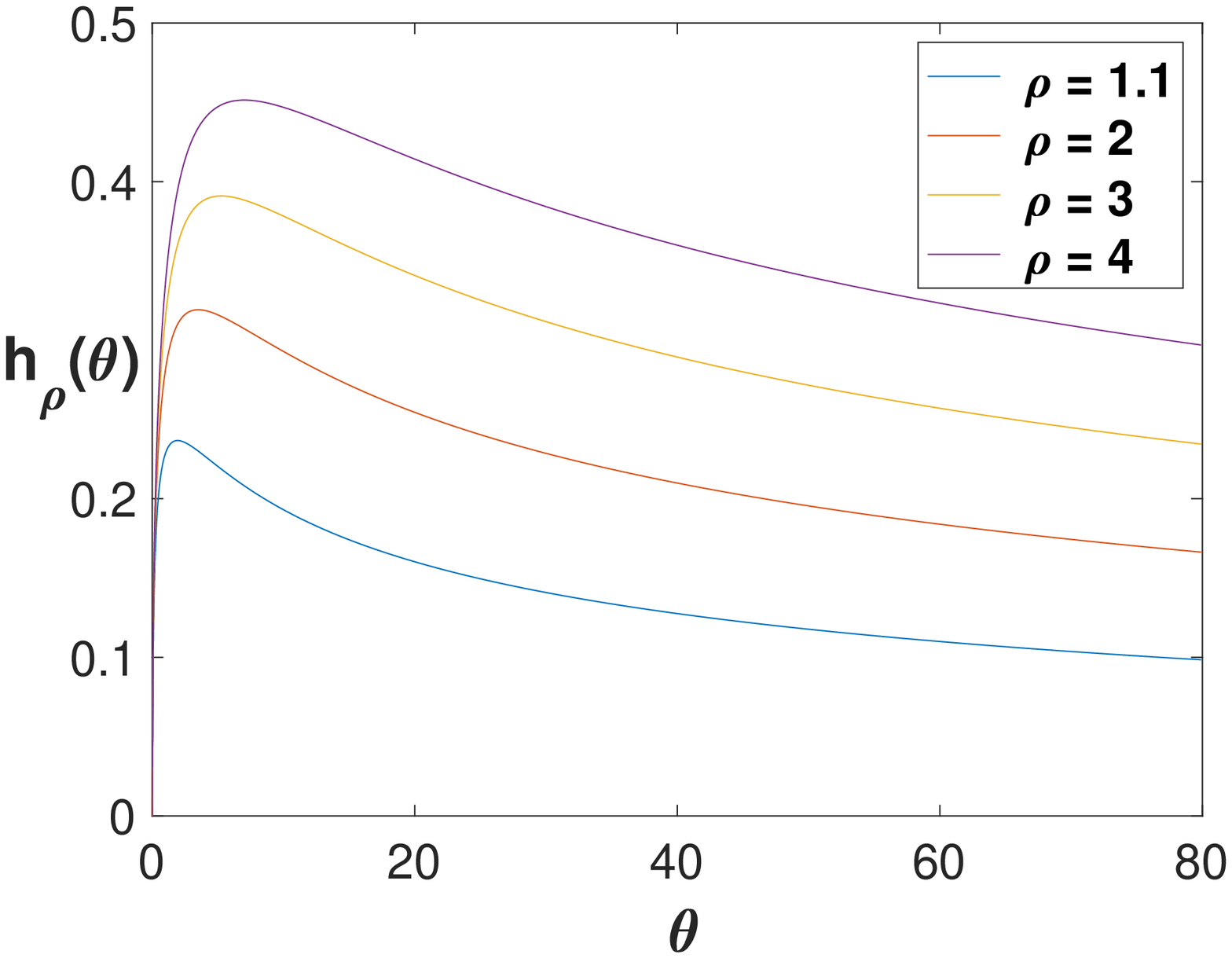} \hfill
\includegraphics[scale=0.25]{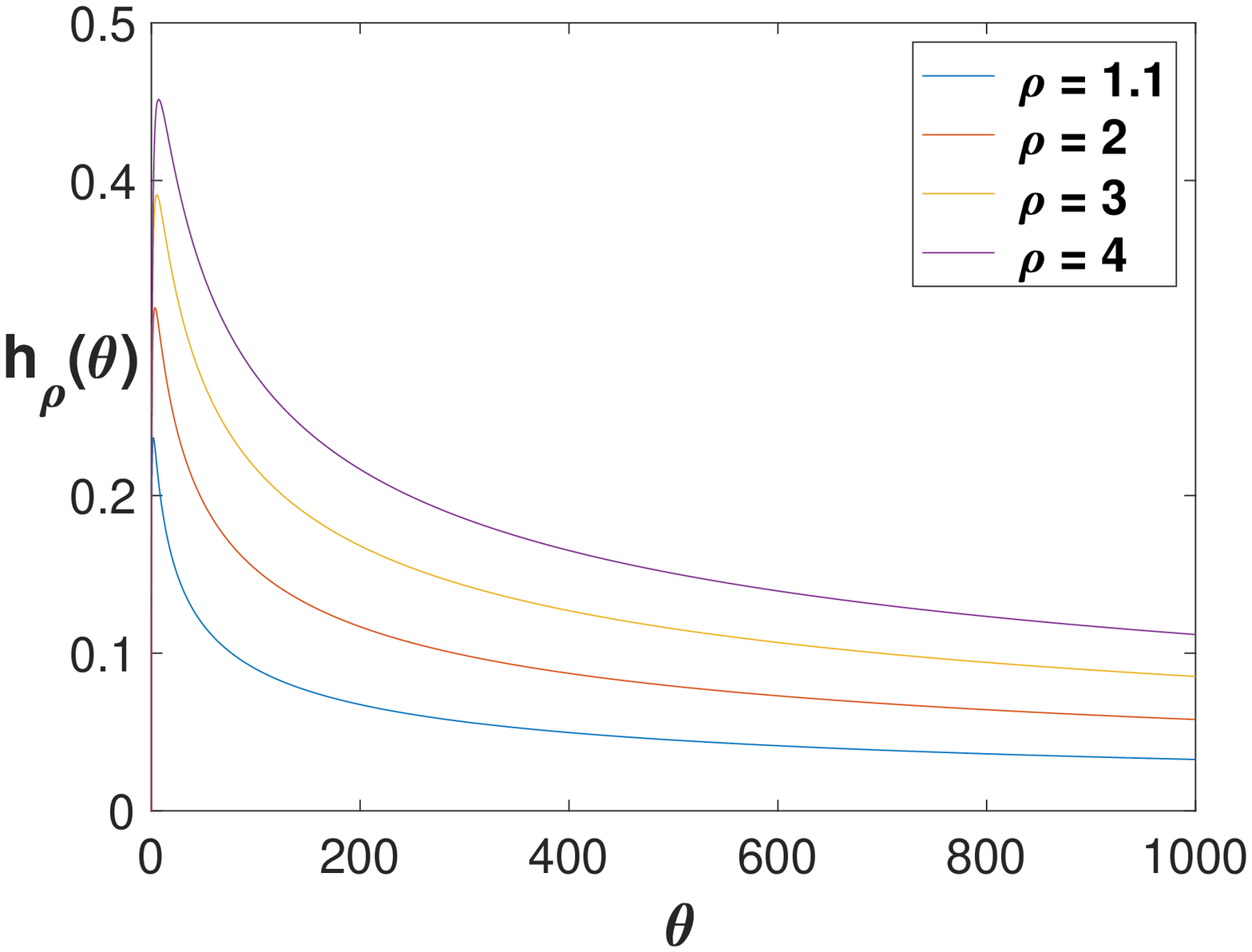} \hfill
\includegraphics[scale=0.25]{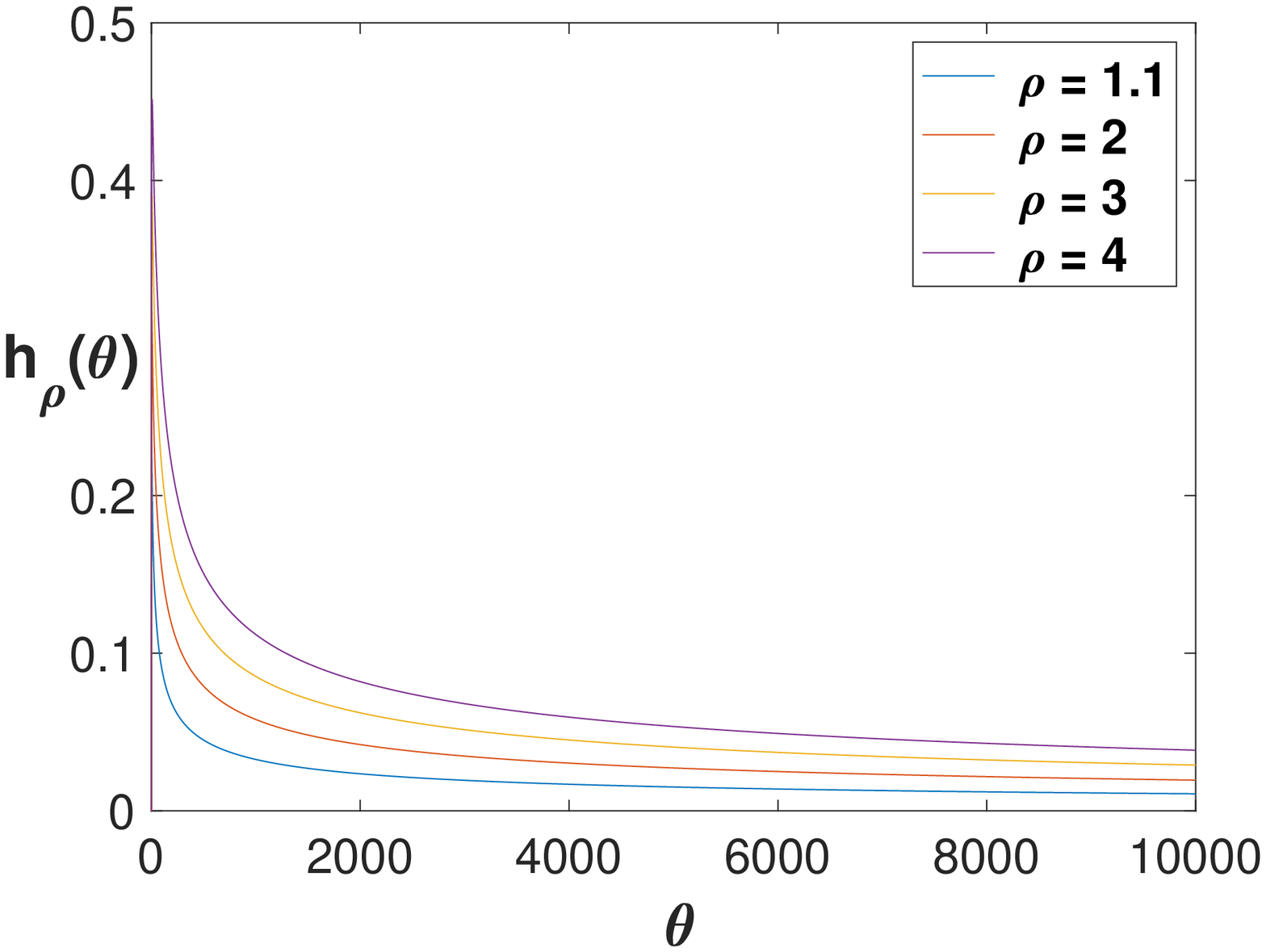}
\caption{$\mathbf{h_\rho(\theta)}$ vs. $\mathbf{\theta}$ for different values of the exploration coefficient $\rho$ in Algorithm~\ref{alg:UCB}. The graphs exhibit a unique global maximizer $\theta^\ast_\rho$ for each $\rho$. The $\left( \theta^\ast_\rho, h_\rho\left( \theta^\ast_\rho \right) \right)$ pairs for $\rho\in\{1.1,2,3,4\}$ in that order are: $(1.9, 0.2367), (3.5,0.3192), (5.3,0.3909), (7,0.4514)$.}
\label{fig}
\end{figure}
A closed-form expression for $\lambda_\rho^{\ast}(\theta)$ and $h_\rho(\theta)$ is provided in Appendix~\ref{sec:expressions}. For a fixed $\rho$, the function $h_\rho(\theta)$ is observed to be uni-modal in $\theta$ and admit a global maximum at a unique $\theta_\rho^{\ast} := \arg\sup_{\theta\geqslant 0}h_\rho(\theta)$, bounded away from $0$. Thus, Theorem~\ref{cor} establishes that the worst-case (instance-independent) regret admits the sharp asymptotic $R^\pi_n \sim h_\rho\left( \theta^\ast_\rho \right)\sqrt{n\log n}$. In standard bandit parlance, this substantiates that the $\mathcal{O}\left( \sqrt{n\log n} \right)$ worst-case (minimax) performance guarantee of canonical UCB cannot be improved in terms of its horizon-dependence. In addition, the result also specifies the precise asymptotic constants achievable in the worst-case setting. This can alternately be viewed as a direct approach to proving the $\mathcal{O}\left( \sqrt{n\log n} \right)$ performance bound for UCB vis-\`a-vis conventional minimax analyses such as those provided in \cite{bubeck2012regret}, among others.


\textbf{Proof sketch.} On a high level, note that when $\Delta=\sqrt{{\left( \theta\log n \right)}/{n}}$ (one possible instance in the ``moderate gap'' regime), we have that $\mathbb{E}R_n^\pi = \Delta \mathbb{E}\left[ n - N_{i^\ast}(n) \right] = \sqrt{{\left( \theta\log n \right)}/{n}} \mathbb{E}\left[ n - N_{i^\ast}(n) \right] \sim \sqrt{\theta}\left( 1 - \lambda_{\rho}^\ast(\theta) \right)\sqrt{n\log n} = h_\rho(\theta)\sqrt{n\log n}$, where the conclusion on asymptotic-equivalence follows using Theorem~\ref{thm:rates}, together with the fact that convergence in probability, coupled with uniform integrability of $N_{i^\ast}(n)/n$, implies convergence in mean. The desired statement that the ``stochastic'' regret $R_n^\pi$ also admits the same sharp $h_\rho(\theta)\sqrt{n\log n}$ asymptotic, can be arrived at via a finer analysis using its definition in \eqref{eqn:stochastic_regret}, together with Theorem~\ref{thm:rates}. In other regimes of $\Delta$, viz., $o\left( \sqrt{\left( \log n \right)/ n }\right)$ ``small'' and $\omega\left( \sqrt{\left( \log n \right)/ n} \right)$ ``large'' gaps, we already know that $R_n^\pi = o_p \left( \sqrt{n\log n} \right)$. The assertion for ``small'' gaps is obvious from $\mathbb{E}R_n^\pi \leqslant \Delta n = o\left( \sqrt{n\log n} \right)$, followed by Markov's inequality, while that for ``large'' gaps uses the well-known result that Algorithm~\ref{alg:UCB} with $\rho>1$ has its regret bounded as $\mathbb{E}R_n^{\pi} \leqslant C\rho\left( {\left(\log n\right)}/{\Delta} + {1}/{\left(\rho-1\right)}\right)$ for some absolute constant $C$ when $\Delta \leqslant 1$ (see \cite{audibert2009exploration}, Theorem~7), together with Markov's inequality. Thus, it must be the case that the multiplicative constant $\sup_{\theta \geqslant 0}h_\rho(\theta)$ obtained in the ``moderate'' gap regime indeed corresponds to the worst-case performance (over all possible values of $\Delta$) of the algorithm. We also remark that while we expect Theorem~\ref{thm:rates} to hold also for smaller values of $\rho$, such a result can only be achieved at the expense of the algorithm's expected regret in the ``large gap'' regime (see Remark~\ref{remark}). In such cases, therefore, the worst-case performance of the algorithm will no longer occur for instances with ``moderate'' gaps, but in fact, will be shifted to the ``large gap'' regime where the policy will incur linear regret. Full proof of Theorem~\ref{cor} is provided in Appendix~\ref{sec:cor}. \hfill $\square$

\textbf{Towards diffusion asymptotics.} Diffusion scaling is a standard tool for performance evaluation of stochastic systems that is widely used in the operations research literature, with origins in queuing theory (see \cite{glynn1990diffusion} for a survey). Under this scaling, time is accelerated linearly in $n$, space contracted by a factor of $\sqrt{n}$, and a sequence of systems indexed by $n$ is considered. In our problem, the $n^{\text{th}}$ such system refers to an instance of the two-armed bandit with: $n$ as the horizon of play; a gap that vanishes in the horizon as $\Delta={c}/{\sqrt{n}}$ for some fixed $c$; and fixed reward variances given by $\sigma_1^2,\sigma_2^2$. This is a natural scaling for MAB experiments in that it ``preserves'' the hardness of the learning problem as $n$ sweeps over the sequence of systems. Recall also from previous discussion that the ``hardest'' information-theoretic instances have a $\Theta\left( {1}/{\sqrt{n}} \right)$ gap; in short, the diffusion limit is an appropriate asymptotic lens for observing interesting process-level behavior in the MAB problem. However, despite the aforementioned reasons, the diffusion limit behavior of bandit algorithms remains poorly understood and largely unexplored. A recent foray was made in \cite{wager2021diffusion}, however, standard bandit algorithms such as the ones discussed in this paper remain outside the ambit of their work due to the nature of assumptions underlying their weak convergence analysis; similar results are provided also in \cite{fan2021diffusion} for the widely used Gaussian prior-based version of Thompson Sampling. In Theorem~\ref{thm:diffusion} stated next, we provide the first full characterization of the diffusion-limit regret performance of canonical UCB, which is based on showing that the cumulative reward processes associated with the arms, when appropriately re-centered and re-scaled, converge in law to independent Brownian motions.

\begin{theorem}[Diffusion asymptotics for canonical UCB]
\label{thm:diffusion}

Suppose that the mean reward of arm~$i\in\{1,2\}$ is given by $\mu_i = \mu + \theta_i/\sqrt{n}$, where $n$ is the horizon of play, and $\mu, \theta_1, \theta_2\geqslant 0$ are fixed constants, and reward variances are $\sigma_1^2, \sigma_2^2$. Define $\Delta_0 := \left\lvert \theta_1 - \theta_2 \right\rvert$. Denote the cumulative reward earned from arm~$i$ until time $m$ by $S_{i,m} := \sum_{j=1}^{N_i(m)}X_{i,j}$, and let $\tilde{S}_{i,m} := S_{i,m} - \mu N_i(m)$. Then, the following process-level convergences hold under the policy $\pi$ given by Algorithm~\ref{alg:UCB} initialized with $\rho>1$:
\begin{align*}
\text{(I)}\ &\left( \frac{\tilde{S}_{1,\left\lfloor nt \right\rfloor}}{\sqrt{n}}, \frac{\tilde{S}_{2,\left\lfloor nt \right\rfloor}}{\sqrt{n}} \right) \Rightarrow \left( \frac{\theta_1t}{2} + \frac{\sigma_1}{\sqrt{2}} B_1(t), \frac{\theta_2t}{2} + \frac{\sigma_2}{\sqrt{2}} B_2(t) \right), \\
\text{(II)}\ &\left( \frac{R^{\pi}_{\left\lfloor nt \right\rfloor}}{\sqrt{n}} \right) \Rightarrow \left( \frac{\Delta_0t}{2} + \sqrt{\frac{\sigma_1^2+\sigma_2^2}{2}}\tilde{B}(t) \right),
\end{align*}
where the process-level convergence is over $t\in[0,1]$, and $B_1(t)$ and $B_2(t)$ are independent standard Brownian motions in $\mathbb{R}$, and $\tilde{B}(t) := -\sqrt{\frac{\sigma_1^2}{\sigma_1^2+\sigma_2^2}}B_1(t) - \sqrt{\frac{\sigma_2^2}{\sigma_1^2+\sigma_2^2}}B_2(t)$.

\end{theorem}

\textbf{Proof sketch.} Note that if the arms are played $\left\lfloor n/2 \right\rfloor$ times each \emph{independently} over the horizon of play $n$ (resulting in $N_i(n) = \left\lfloor n/2 \right\rfloor$, $i\in\{1,2\}$), part (I) of the stated assertion would immediately follow from Donsker's Theorem (see \cite{billingsley2013convergence}, Section 14). However, since the sequence of plays, and hence also the eventual allocation $\left( N_1(n), N_2(n) \right)$, is determined \emph{adaptively} by the policy, the aforementioned convergence may no longer be true. Here, the result hinges crucially on the observation from Theorem~\ref{thm:rates} that for any arm~$i\in\{1,2\}$ as $n\to\infty$, ${N_i(n)}/{n}\xrightarrow{p}{1}/{2}$ under UCB when $\Delta \asymp {1}/{\sqrt{n}}$ (diffusion-scaled gaps are covered under the ``small gap'' regime). This observation facilitates a standard ``random time-change'' argument $t\gets N_i(\left\lfloor nt \right\rfloor)/n$, $i\in\{1,2\}$, which followed upon by an application of Donsker's Theorem, leads to the stated assertion in (I). This has the profound implication that for diffusion-scaled gaps, a two-armed bandit under UCB is, in fact, well-approximated by a classical system with \emph{independent} samples (sample-interdependence due to the adaptive nature of the policy is washed away in the limit). The conclusion in (II) follows after a direct application of the Continuous Mapping Theorem (see \cite{billingsley2013convergence}, Theorem~2.7) to (I). \hfill $\square$ 

\textbf{Discussion.} An immediate observation following Theorem~\ref{thm:diffusion} is that the normalized regret ${R_n^\pi}/{\sqrt{n}}$ is asymptotically Gaussian with mean ${\Delta_0}/{2}$ and variance $\left(\sigma_1^2 + \sigma_2^2\right)/2$ under canonical UCB. Apart from aiding in obvious inferential tasks such as the construction of confidence intervals (see, e.g., the binary hypothesis testing example referenced in Figure~\ref{fig:weak-limit4}(c) where asymptotic-normality of the gap estimator $\hat{\Delta}$ follows as a consequence of Theorem~\ref{thm:diffusion}, in conjunction with the Continuous Mapping Theorem), etc., such information provides new insights as to the problem's minimax complexity as well. This is because the $\Delta \asymp 1/\sqrt{n}$-scale is known to be the information-theoretic ``worst-case'' for the problem; the smallest achievable regret in this regime must therefore, be asymptotically dominated by that under UCB, i.e., $\Delta_0\sqrt{n}/2$. It is also noteworthy that while the diffusion limit in Theorem~\ref{thm:diffusion} does not itself depend on the exploration coefficient $\rho$ of the algorithm, the rate at which the system converges to said limit indeed depends on $\rho$. Theorem~\ref{thm:diffusion} will continue to hold only as long as $\rho=\omega\left( \left(\log\log n \right)/\log n\right)$; for smaller $\rho$, the convergence of ${N_i(n)}/{n}$ to ${1}/{2}$ may no longer be true (refer to the proof of Theorem~\ref{thm:rates} in the ``small'' gap regime in Appendix~\ref{sec:small_gap}).

\textbf{Comparison with Thompson Sampling.} The closest related works in extant literature are \cite{wager2021diffusion,fan2021diffusion}, which establish weak convergence results for the Gaussian prior-based Thompson Sampling algorithm. Cited works show that the resulting limit process may at best be characterizable as a solution to an appropriate stochastic differential equation or ordinary differential equation driven by a suitably (random) time-changed Brownian motion. Consequently, the diffusion limit under Thompson Sampling offers less clear insights as to the limiting distribution of arm-pulls vis-\`a-vis UCB. The principal complexity here that impedes a lucid characterization of Thompson Sampling in the style of Theorem~\ref{thm:diffusion} stems from its instability and non-degeneracy of the arm-sampling distribution; see the ``incomplete learning'' phenomenon referenced in Theorem~\ref{thm:TS-BP-00}, and the numerical illustration in Figure~\ref{fig:overview}(b), 1(c) and \ref{fig:weak-limit4}(a).

\section{Concluding remarks and open problems}
\label{sec:conclusion}

Theorem~\ref{thm:zero-gap-K-arm} extends Theorem~\ref{thm:rates} to the $K$-armed setting in the special case where all arms are either optimal, or the optimal arms are ``well-separated'' from the inferior arms. We believe the statement of Theorem~\ref{thm:rates} can, in fact, be fully generalized to the $K$-armed setting, including extensions for appropriately redefined ``small'' and ``moderate'' gaps. The $K$-armed problem under UCB is of interest in its own right: we postulate a division of sampling effort within and across ``clusters'' of ``similar'' arms, determined by their relative sub-optimality gaps in the spirit of Theorem~\ref{thm:rates}. We expect that similar generalizations are possible also for Theorem~\ref{cor} and Theorem~\ref{thm:diffusion}. For Thompson Sampling, on the other hand, things are less obvious even in the two-armed setting. For example, in spite of compelling numerical evidence (refer, e.g., to Figure~\ref{fig:overview}(b)) suggesting a plausibly non-degenerate distribution of arm-sampling rates for bounded rewards in $[0,1]$ with means away from $0$, the proof of Theorem~\ref{thm:TS-BP-00} relies heavily on the rewards being deterministic $0/1$, and cannot be extended to the general case. In addition, similar results are conjectured also for the more widely used Gaussian prior-based version of the algorithm. Such results may shed light on several ``small gap'' performance metrics of Thompson Sampling, including the stochastic behavior of its normalized minimax regret, which could prove useful in a variety of application areas. However, since technical difficulty is a major impediment in the derivation of such results, this currently remains an open problem.

\newpage

\appendix

{\bf Addendum.} General organization:

\begin{enumerate}

\item Appendix~\ref{sec:expressions} provides closed-form expressions for the $\lambda^\ast_{\rho}(\theta)$ and $h_\rho(\theta)$ functions that appear in Theorem~\ref{thm:rates} and Theorem~\ref{cor}.
\item Appendix~\ref{sec:aux} states three ancillary results that will be used in other proofs.
\item Appendix~\ref{sec:large_gap} provides the proof of Theorem~\ref{thm:rates} in the ``large gap'' regime.
\item Appendix~\ref{sec:small_gap} provides the proof of Theorem~\ref{thm:rates} in the ``small gap'' regime.
\item Appendix~\ref{sec:moderate_gap} provides the proof of Theorem~\ref{thm:rates} in the ``moderate gap'' regime.
\item Appendix~\ref{proof:K-arm} provides the proof of Theorem~\ref{thm:zero-gap-K-arm}. 
\item Appendix~\ref{appendix:thm5} provides the proof of Theorem~\ref{thm:TS-BP-00}.
\item Appendix~\ref{sec:cor} provides the proof of Theorem~\ref{cor}.
\item Appendix~\ref{sec:diffusion} provides the proof of Theorem~\ref{thm:diffusion}.
\item Appendix~\ref{facts} provides proofs for the ancillary results stated in Appendix~\ref{sec:aux}.

\end{enumerate}

\textit{Note.} In the proofs that follow, $\left\lceil \cdot \right\rceil$ has been used to denote the ``ceiling operator,'' i.e., $\left\lceil x \right\rceil = \inf \left\lbrace \nu\in\mathbb{N} : \nu \geqslant x \right\rbrace$ for any $x\in\mathbb{R}$. Similarly, $\left\lfloor \cdot \right\rfloor$ denotes the ``floor operator,'' i.e., $\left\lfloor x \right\rfloor = \sup \left\lbrace \nu\in\mathbb{N} : \nu \leqslant x \right\rbrace$ for any $x\in\mathbb{R}$.

\section{Closed-form expressions for $\lambda^\ast_{\rho}(\theta)$ and $h_\rho(\theta)$}
\label{sec:expressions}

$\lambda_\rho^{\ast}(\theta)$ is given by:
\begin{align}
\lambda_\rho^{\ast}(\theta) = \frac{1}{2} + \sqrt{\frac{1}{4} - \frac{1}{\left( 1 + \sqrt{1 + \frac{\theta}{\rho}}\right)^2}}. \label{eqn:limit_solution_lambda}
\end{align}

$h_\rho(\theta)$ is given by:
\begin{align}
h_\rho(\theta) = \sqrt{\theta}\left( \frac{1}{2} - \sqrt{\frac{1}{4} - \frac{1}{\left( 1 + \sqrt{1 + \frac{\theta}{\rho}}\right)^2}}\right). \label{eqn:limit_solution_h}
\end{align}

\section{Auxiliary results}
\label{sec:aux}

We will frequently use the following version of the Chernoff-Hoeffding inequality \citep{hoeffding} in our proofs:

\begin{fact}[Chernoff-Hoeffding bound]
\label{hoeffding}
Suppose that $\left\lbrace Y_{i,j} : i\in\{1,2\}, j\in\mathbb{N} \right\rbrace$ is a collection of independent, zero-mean random variables such that $\forall\ i\in\{1,2\}$, $j\in\mathbb{N}$, $Y_{i,j}\in\left[ c_i, 1+c_i \right]$ almost surely, for some fixed $c_1,c_2\leqslant 0$. Then, for any $m_1,m_2\in\mathbb{N}$ and $\alpha>0$,
\begin{align*}
\mathbb{P}\left( \frac{\sum_{j=1}^{m_1}Y_{1,j}}{m_1} - \frac{\sum_{j=1}^{m_2}Y_{2,j}}{m_2} \geqslant \alpha \right) \leqslant \exp \left( \frac{-2\alpha^2m_1m_2}{m_1+m_2} \right).
\end{align*}
\end{fact}

\textbf{Proof.}
Let $[n]:=\{1,...,n\}$ for $n\in\mathbb{N}$. The Chernoff-Hoeffding inequality in its standard form states that for independent, zero-mean, bounded random variables $\left\lbrace Z_j : j\in[n]\right\rbrace$ with $Z_j\in\left[ a_j, b_j \right]\ \forall\ j\in[n]$, the following holds for any $\varepsilon>0$,
\begin{align}
\mathbb{P}\left( \sum_{j=1}^{n}Z_j \geqslant \varepsilon n\right) \leqslant \exp\left( \frac{-2\varepsilon^2n^2}{\sum_{j=1}^{n}\left(b_j-a_j\right)^2} \right). \label{eqn:chernoff}
\end{align}
The desired form of the inequality can be obtained by making the following substitutions in \eqref{eqn:chernoff}: $n \gets m_1 + m_2$; $Z_j \gets \frac{Y_{1,j}}{m_1}$, $a_j \gets \frac{c_1}{m_1}$, $b_j \gets \frac{1+c_1}{m_1}$ for $j\in\left[ m_1 \right]$; $Z_j \gets \frac{-Y_{2,j-m_1}}{m_2}$, $a_j \gets \frac{-(1+c_2)}{m_2}$, $b_j \gets \frac{-c_2}{m_2}$ for $j\in\left[m_1+m_2\right]\backslash\left[ m_1 \right]$; and $\varepsilon\gets\frac{\alpha}{m_1+m_2}$, in that order. \hfill $\square$

In addition, we will use in the proof of Theorem~\ref{thm:TS-BP-00} the following two properties of the Beta distribution:

\begin{fact}
\label{fact:1}
If $\theta_k,\tilde{\theta}_k$ are Beta$(1,k+1)$-distributed, and $\theta_k, \tilde{\theta}_l$ are independent $\forall\ k,l\in\mathbb{N}\cup\{0\}$, then
\begin{align}
\mathbb{P}\left( \theta_k > \tilde{\theta}_l \right) = \frac{l+1}{k+l+2}\ \ \ \ {\text{for any}}\ k,l\in\mathbb{N}\cup\{0\}. \notag
\end{align}
\end{fact}

\begin{fact}
\label{fact:2}
If $\theta_k,\tilde{\theta}_k$ are Beta$(k+1,1)$-distributed, and $\theta_k,\tilde{\theta}_l$ are independent $\forall\ k,l\in\mathbb{N}\cup\{0\}$, then
\begin{align}
\mathbb{P}\left( \theta_k > \tilde{\theta}_l \right) = \frac{k+1}{k+l+2}\ \ \ \ {\text{for any}}\ k,l\in\mathbb{N}\cup\{0\}. \notag
\end{align}
\end{fact}

The proofs for Fact~\eqref{fact:1} and Fact~\eqref{fact:2} are elementary, and provided in Appendix~\ref{facts}.

\section{Proof of Theorem~\ref{thm:rates} in the ``large gap'' regime}
\label{sec:large_gap}

The proof is straightforward in this regime. We know that for $\rho>1$, $\mathbb{E}R_n^{\pi} \leqslant C\rho\left(\frac{\log n}{\Delta} + \frac{\Delta}{\rho-1} \right)$ for some absolute constant $C$ (see \cite{audibert2009exploration}, Theorem~7). Since $\mathbb{E}R_n^\pi=\Delta \mathbb{E}\left[ n - N_{i^\ast}(n)\right]$, it follows that $\mathbb{E}\left[ \frac{n - N_{i^\ast}(n)}{n} \right] = o(1)$ in the ``large gap'' regime. Using Markov's inequality, we then conclude that $\frac{n - N_{i^\ast}(n)}{n} = o_p(1)$, or equivalently, $\lim_{n\to\infty}\frac{N_{i^\ast}(n)}{n} = 1$. Results for ``small'' and ``moderate'' gaps are provided separately in Appendix~\ref{sec:small_gap} and Appendix~\ref{sec:moderate_gap} respectively. \hfill $\square$

\section{Proof of Theorem~\ref{thm:rates} in the ``small gap'' regime}
\label{sec:small_gap}

Without loss of generality, suppose that arm~1 is optimal, i.e., $\mu_1 \geqslant \mu_2$. We will show that for any $\epsilon>0$, it follows that $\lim_{n\to\infty}\mathbb{P}\left( \frac{N_1(n)}{n} \geqslant \frac{1}{2} + \epsilon \right) = 0$. Then, since arm~2 is inferior, an identical result would naturally hold for it as well. Combining the two would prove our assertion as desired. To this end, pick an arbitrary $\epsilon\in\left(0,1/2\right)$, define $u(n) := \left\lceil \left( \frac{1}{2} + \epsilon \right)n \right\rceil$, and consider the following:

\begin{align}
N_1(n) &\leqslant u(n) + \sum_{t=u(n)+1}^{n}\mathbbm{1}\left\lbrace \pi_t = 1,\ N_1(t-1) \geqslant u(n) \right\rbrace\ \ \ \ \ \ \ \ \ \ \ \ \ \ \ \ \ \ \ \ \ \ \ \ \text{(this is always true)} \notag \\
&= u(n) + \sum_{t=u(n)}^{n-1}\mathbbm{1}\left\lbrace \pi_{t+1} = 1,\ N_1(t) \geqslant u(n) \right\rbrace \notag \\
&\leqslant u(n) + \sum_{t=u(n)}^{n-1}\mathbbm{1}\left\lbrace \pi_{t+1} = 1,\ N_1(t) \geqslant u(t) \right\rbrace \notag \\
&\leqslant u(n) + \sum_{t=u(n)}^{n-1}\mathbbm{1}\left\lbrace \bar{X}_1(t) - \bar{X}_2(t) \geqslant \sqrt{\rho\log t}\left( \frac{1}{\sqrt{N_2(t)}} - \frac{1}{\sqrt{N_1(t)}} \right),\ N_1(t) \geqslant u(t) \right\rbrace \notag \\
&= u(n) + \underbrace{\sum_{t=u(n)}^{n-1}\mathbbm{1}\left\lbrace \bar{Y}_1(t) - \bar{Y}_2(t) \geqslant \sqrt{\rho\log t}\left( \frac{1}{\sqrt{N_2(t)}} - \frac{1}{\sqrt{N_1(t)}} \right) - \Delta,\ N_1(t) \geqslant u(t) \right\rbrace}_{=: Z(n)}, \label{eqn:1}
\end{align}
where $\bar{Y}_i(t) := \frac{\sum_{j=1}^{N_i(t)}Y_{i,j}}{N_i(t)}$ with $Y_{i,j} := X_{i,j} - \mu_i$, $i\in\{1,2\}, j\in\mathbb{N}$. Clearly, $Y_{i,j}$'s are independent, zero-mean, and $Y_{i,j} \in \left[ -\mu_i, 1-\mu_i \right]\ \forall\ i\in\{1,2\}, j\in\mathbb{N}$.

\subsection{An almost sure lower bound on the arm-sampling rates}

As a meta-result, we will first show that ${N_i(n)}/{n}$, for both arms~$i\in\{1,2\}$, is bounded away from $0$ by a positive constant, almost surely. To this end, consider $n$ large enough such that for the $\epsilon$ selected earlier, we have $\Delta < \sqrt{\frac{\rho\log n}{n}}\left( \frac{1}{\sqrt{1/2-\epsilon}} - \frac{1}{\sqrt{1/2+\epsilon}} \right)$; this is possible since $\Delta = o\left( \sqrt{\frac{\log n}{n}} \right)$ in the ``small gap'' regime. Working with a large enough $n$ will allow us to use the Chernoff-Hoeffding bound (Fact~\ref{hoeffding}) in step $(\star)$ in the forthcoming analysis. Observe from \eqref{eqn:1} that
\begin{align}
&\mathbb{E}Z(n) \notag \\
=\ &\sum_{t=u(n)}^{n-1}\mathbb{P}\left( \bar{Y}_1(t) - \bar{Y}_2(t) \geqslant \sqrt{\rho\log t}\left( \frac{1}{\sqrt{N_2(t)}} - \frac{1}{\sqrt{N_1(t)}} \right) - \Delta,\ N_1(t) \geqslant u(t) \right) \notag \\
=\ &\sum_{t=u(n)}^{n-1}\sum_{m=u(t)}^{t-1}\mathbb{P}\left( \frac{\sum_{j=1}^{m}Y_{1,j}}{m} - \frac{\sum_{j=1}^{t-m}Y_{2,j}}{t-m} \geqslant \sqrt{\rho\log t}\left( \frac{1}{\sqrt{t-m}} - \frac{1}{\sqrt{m}} \right) - \Delta,\ N_1(t) = m \right) \notag \\
\underset{\mathrm{(\star)}}{\leqslant}\ &\sum_{t=u(n)}^{n-1}\sum_{m=u(t)}^{t-1}\mathbb{P}\left( \frac{\sum_{j=1}^{m}Y_{1,j}}{m} - \frac{\sum_{j=1}^{t-m}Y_{2,j}}{t-m} \geqslant \sqrt{\rho\log t}\left( \frac{1}{\sqrt{t-m}} - \frac{1}{\sqrt{m}} \right) - \Delta\right) \notag \\
\underset{\mathrm{(\dag)}}{\leqslant}\ &\sum_{t=u(n)}^{n-1}\sum_{m=u(t)}^{t-1}\exp\left[ -2\rho\left(1-2\sqrt{\frac{m}{t}\left( 1-\frac{m}{t} \right)}\right)\log t\right]\exp\left[ 4\Delta\sqrt{\rho t\log t}\left( \sqrt{\frac{m}{t}} - \sqrt{1-\frac{m}{t}} \right) \sqrt{\frac{m}{t}\left( 1 - \frac{m}{t} \right)} \right] \notag \\
\underset{\mathrm{(\ddag)}}{\leqslant}\ &\sum_{t=u(n)}^{n-1}\sum_{m=u(t)}^{t-1}\exp\left[ -2\rho\left(1-\sqrt{1-4\epsilon^2}\right)\log t\right]\exp\left[ 4\Delta\sqrt{\rho t\log t}\left( \sqrt{\frac{m}{t}} - \sqrt{1-\frac{m}{t}} \right) \sqrt{\frac{m}{t}\left( 1 - \frac{m}{t} \right)} \right] \notag \\
\leqslant\ &\sum_{t=u(n)}^{n-1}\sum_{m=u(t)}^{t-1}\exp\left[ -2\rho\left(1-\sqrt{1-4\epsilon^2}\right)\log t\right]\exp\left[ 4\Delta\sqrt{\rho t\log t} \right] \notag \\
\leqslant\ &\sum_{t=u(n)}^{n-1}\sum_{m=u(t)}^{t-1}\exp\left[ -2\rho\left(1-\sqrt{1-4\epsilon^2}\right)\log t\right]\exp\left[ 4\Delta\sqrt{\rho n\log n} \right] \notag \\
\leqslant\ &\exp\left[ 4\Delta\sqrt{\rho n\log n} \right]\sum_{t=u(n)}^{n-1}t^{-\left( 2\rho-1-2\rho\sqrt{1-4\epsilon^2}\right)} \notag \\
=\ &\exp\left[ o\left( 4\sqrt{\rho}\log n \right) \right]\sum_{t=u(n)}^{n-1}t^{-\left( 2\rho-1-2\rho\sqrt{1-4\epsilon^2}\right)}\ \ \ \ \ \ \ \ \ \ \left( \because \Delta = o\left( \sqrt{\frac{\log n}{n}} \right) \right) \notag \\
\leqslant\ &n^{\frac{1}{2}-\epsilon}\sum_{t=u(n)}^{n-1}t^{-\left( 2\rho-1-2\rho\sqrt{1-4\epsilon^2}\right)}, \label{eqn:2}
\end{align}
where $(\dag)$ follows after an application of the Chernoff-Hoeffding bound (Fact~\ref{hoeffding}), $(\ddag)$ since $\frac{m}{t}\left( 1 - \frac{m}{t} \right) \leqslant \frac{1}{4} - \epsilon^2$ on the interval $\left\lbrace m : u(t) \leqslant m \leqslant t-1 \right\rbrace$, and the last inequality in \eqref{eqn:2} holds for $n$ large enough. Now consider an arbitrary $\delta>0$. Then,
\begin{align}
\mathbb{P}\left( N_1(n) - u(n) \geqslant \delta n\right) &\leqslant \mathbb{P}\left( Z(n) \geqslant \delta n\right) & \mbox{(using \eqref{eqn:1})} \notag \\
&\leqslant \frac{\mathbb{E}Z(n)}{\delta n} & \mbox{(Markov's inequality)} \notag \\
&\leqslant \left(\frac{n^{-\left( \frac{1}{2}+\epsilon\right)}}{\delta}\right)\sum_{t=u(n)}^{n-1}t^{-\left( 2\rho-1-2\rho\sqrt{1-4\epsilon^2}\right)} & \mbox{(using \eqref{eqn:2})} \notag \\
\implies \mathbb{P}\left( \frac{N_1(n)}{n} \geqslant \frac{1}{2}+\epsilon+\delta + \frac{1}{n} \right) &\leqslant \left(\frac{n^{-\left( \frac{1}{2}+\epsilon\right)}}{\delta}\right)\sum_{t=\left\lceil \frac{n}{2} \right\rceil}^{n-1}t^{-\left( 2\rho-1-2\rho\sqrt{1-4\epsilon^2}\right)}. \label{eqn:temp}
\end{align}
Define $g(\rho,\epsilon) := \frac{1}{2} + \epsilon + 2\rho-1-2\rho\sqrt{1-4\epsilon^2}$. Since $\rho>1$ is fixed, and $\epsilon\in\left(0,1/2\right)$ is arbitrary, it is possible to push $\epsilon$ close to ${1}/{2}$ to ensure that $g(\rho,\epsilon) > 2$. Therefore, $\exists\ \epsilon_\rho\in\left( 0, 1/2 \right)$ s.t. $g(\rho,\epsilon) > 2$ for $\epsilon \geqslant \epsilon_\rho$. Plugging in $\epsilon = \epsilon_\rho$ in \eqref{eqn:temp}, we obtain
\begin{align*}
\mathbb{P}\left( \frac{N_1(n)}{n} \geqslant \frac{1}{2}+\epsilon_\rho+\delta + \frac{1}{n} \right) \leqslant \left(\frac{2^{2\rho-1}}{\delta}\right)n^{-\left( g\left(\rho,\epsilon_\rho\right) - 1 \right)}.
\end{align*}
Note that since $\epsilon_\rho<1/2$, $\exists\ \epsilon^\prime_{\rho}<1/2$ s.t. $\epsilon_\rho + 1/n < \epsilon^\prime_{\rho}$ for $n$ large enough, i.e., the following holds for all $n$ large enough:
\begin{align*}
\mathbb{P}\left( \frac{N_1(n)}{n} \geqslant \frac{1}{2}+\epsilon^\prime_\rho+\delta \right) \leqslant \left(\frac{2^{2\rho-1}}{\delta}\right)n^{-\left( g\left(\rho,\epsilon_\rho\right) - 1 \right)}.
\end{align*}
Finally, since $\delta>0$ is arbitrary, and $g\left(\rho,\epsilon_\rho\right) > 2$, it follows from the Borel-Cantelli Lemma that 
\begin{align*}
\limsup_{n\to\infty} \frac{N_1(n)}{n} \leqslant \frac{1}{2} + \epsilon^\prime_\rho < 1\ \ \ \text{w.p. $1$}.
\end{align*}
By assumption, arm~2 is inferior; the above result thus holds, in fact, for both the arms (An almost identical proof can be replicated for rigor). Therefore, we conclude
\begin{align}
\liminf_{n\to\infty} \frac{N_i(n)}{n} \geqslant \frac{1}{2} - \epsilon^\prime_\rho > 0\ \ \ \text{w.p. $1$}\ \ \ \forall\ i\in\{1,2\}. \label{eqn:3}
\end{align}

\subsection{Closing the loop}
\label{subsec:closing_the_loop}

In this part of the proof, we will leverage \eqref{eqn:3} to finally show that ${N_i(n)}/{n} = {1}/{2} + o_p(1)$ for $i\in\{1,2\}$. To this end, recall from \eqref{eqn:1} that
\begin{align}
&\mathbb{E}Z(n) \notag \\
=\ &\sum_{t=u(n)}^{n-1}\mathbb{P}\left( \bar{Y}_1(t) - \bar{Y}_2(t) \geqslant \sqrt{\rho\log t}\left( \frac{1}{\sqrt{N_2(t)}} - \frac{1}{\sqrt{N_1(t)}} \right) - \Delta,\ N_1(t) \geqslant u(t) \right) \notag \\
\leqslant\ &\sum_{t=u(n)}^{n-1}\mathbb{P}\left( \bar{Y}_1(t) - \bar{Y}_2(t) \geqslant \sqrt{\frac{\rho\log t}{t}}\left( \frac{1}{\sqrt{\frac{1}{2}-\epsilon}} - \frac{1}{\sqrt{\frac{1}{2}+\epsilon}} \right) - \Delta,\ N_1(t) \geqslant u(t) \right) \notag \\
\leqslant\ &\sum_{t=u(n)}^{n-1}\mathbb{P}\left( \bar{Y}_1(t) - \bar{Y}_2(t) \geqslant \sqrt{\frac{\rho\log t}{t}}\left( \frac{1}{\sqrt{\frac{1}{2}-\epsilon}} - \frac{1}{\sqrt{\frac{1}{2}+\epsilon}} \right) - \Delta \right) \notag \\
=\ &\sum_{t=u(n)}^{n-1}\mathbb{P}\left( \underbrace{\sqrt{\frac{t}{\rho\log t}} \left( \bar{Y}_1(t) - \bar{Y}_2(t) \right)}_{=:W_t} \geqslant \sqrt{\frac{2}{1-2\epsilon}} - \sqrt{\frac{2}{1+2\epsilon}} - \Delta\sqrt{\frac{t}{\rho\log t}} \right) \notag \\
\leqslant\ &\sum_{t=u(n)}^{n-1}\mathbb{P}\left( W_t \geqslant \frac{1}{\sqrt{1-2\epsilon}} - \frac{1}{\sqrt{1+2\epsilon}} \right), \label{eqn:Wt0}
\end{align}
for $n$ large enough; the last inequality following since $\Delta = o\left( \sqrt{\frac{\log n}{n}} \right)$ and $u(n) > n/2$. Now,
\begin{align}
&\left\lvert W_t \right\rvert \notag \\
\leqslant\ &\sqrt{\frac{t}{\rho\log t}} \left( \left\lvert\frac{\sum_{j=1}^{N_1(t)}Y_{1,j}}{N_1(t)}\right\rvert + \left\lvert\frac{\sum_{j=1}^{N_2(t)}Y_{2,j}}{N_2(t)}\right\rvert \right) \notag \\
=\ &\sqrt{\frac{2t}{\rho\log t}} \left( \sqrt{\frac{\log \log N_1(t)}{N_1(t)}}\left\lvert\frac{\sum_{j=1}^{N_1(t)}Y_{1,j}}{\sqrt{2N_1(t)\log \log N_1(t)}}\right\rvert + \sqrt{\frac{\log \log N_2(t)}{N_2(t)}}\left\lvert\frac{\sum_{j=1}^{N_2(t)}Y_{2,j}}{\sqrt{2N_2(t)\log \log N_2(t)}}\right\rvert \right) \notag \\
\leqslant\ &\sqrt{\frac{2t}{\rho\log t}} \left( \sqrt{\frac{\log \log t}{N_1(t)}}\left\lvert\frac{\sum_{j=1}^{N_1(t)}Y_{1,j}}{\sqrt{2N_1(t)\log \log N_1(t)}}\right\rvert + \sqrt{\frac{\log \log t}{N_2(t)}}\left\lvert\frac{\sum_{j=1}^{N_2(t)}Y_{2,j}}{\sqrt{2N_2(t)\log \log N_2(t)}}\right\rvert \right) \notag \\
=\ &\sqrt{\frac{2\log\log t}{\rho\log t}} \left( \sqrt{\frac{t}{N_1(t)}}\left\lvert\frac{\sum_{j=1}^{N_1(t)}Y_{1,j}}{\sqrt{2N_1(t)\log \log N_1(t)}}\right\rvert + \sqrt{\frac{t}{N_2(t)}}\left\lvert\frac{\sum_{j=1}^{N_2(t)}Y_{2,j}}{\sqrt{2N_2(t)\log \log N_2(t)}}\right\rvert \right). \label{eqn:Wt}
\end{align}

We know that $N_i(t)$, for both arms~$i\in\{1,2\}$, can be lower bounded \emph{path-wise} by a deterministic monotone increasing function of $t$, say $f(t)$, that grows to $+\infty$ as $t\to\infty$. This is a trivial consequence of the structure of canonical UCB (Algorithm~\ref{alg:UCB}), and the fact that the rewards are uniformly bounded. We therefore have for any $i\in\{1,2\}$ that
\begin{align*}
\left\lvert\frac{\sum_{j=1}^{N_i(t)}Y_{i,j}}{\sqrt{2N_i(t)\log \log N_i(t)}}\right\rvert \leqslant \sup_{m\geqslant f(t)} \left\lvert\frac{\sum_{j=1}^{m}Y_{i,j}}{\sqrt{2m\log \log m}}\right\rvert.
\end{align*}

For a fixed arm~$i\in\{1,2\}$, $\left\lbrace Y_{i,j} : j\in\mathbb{N} \right\rbrace$ is a collection of i.i.d. random variables with $\mathbb{E}Y_{i,1} = 0$ and $\text{Var}\left( Y_{i,1} \right) = \text{Var}\left( X_{i,1} \right) \leqslant 1$. Also, $f(t)$ is monotone increasing and coercive in $t$. Therefore, the \emph{Law of the Iterated Logarithm} (see \cite{durrett2019probability}, Theorem~8.5.2) implies
\begin{align}
\limsup_{t\to\infty} \left\lvert\frac{\sum_{j=1}^{N_i(t)}Y_{i,j}}{\sqrt{2N_i(t)\log \log N_i(t)}}\right\rvert \leqslant 1\ \ \ \text{w.p.}\ 1\ \ \forall\ i\in\{1,2\}. \label{eqn:LIL}
\end{align}

Using \eqref{eqn:3}, \eqref{eqn:Wt} and \eqref{eqn:LIL}, we conclude that
\begin{align}
\lim_{t\to\infty} W_t = 0\ \ \ \text{w.p.}\ 1. \label{eqn:Wt_as_0}
\end{align}

Now consider an arbitrary $\delta>0$. Then,
\begin{align}
\mathbb{P}\left( N_1(n) - u(n) \geqslant \delta n \right) &\leqslant \mathbb{P}\left( Z(n) \geqslant \delta n\right) & \mbox{(using \eqref{eqn:1})} \notag \\
&\leqslant \frac{\mathbb{E}Z(n)}{\delta n} & \mbox{(Markov's inequality)} \notag \\
&\leqslant \frac{1}{\delta n}\sum_{t=u(n)}^{n-1}\mathbb{P}\left( W_t \geqslant \frac{1}{\sqrt{1-2\epsilon}} - \frac{1}{\sqrt{1+2\epsilon}} \right). & \mbox{(using \eqref{eqn:Wt0})} \notag \\
&\leqslant \frac{1}{\delta}\sup_{t > n/2}\mathbb{P}\left( W_t \geqslant \frac{1}{\sqrt{1-2\epsilon}} - \frac{1}{\sqrt{1+2\epsilon}} \right) \notag \\ 
\implies \mathbb{P}\left( \frac{N_1(n)}{n} \geqslant \frac{1}{2} + \epsilon + \delta + \frac{1}{n} \right) &\leqslant \frac{1}{\delta}\sup_{t > n/2}\mathbb{P}\left( W_t \geqslant \frac{1}{\sqrt{1-2\epsilon}} - \frac{1}{\sqrt{1+2\epsilon}} \right). \notag
\end{align}

Since $\epsilon,\delta>0$ are arbitrary, it follows that for $n$ large enough,
\begin{align}
\mathbb{P}\left( \frac{N_1(n)}{n} \geqslant \frac{1}{2} + 2(\epsilon + \delta) \right) &\leqslant \frac{1}{\delta}\sup_{t > n/2}\mathbb{P}\left( W_t \geqslant \frac{1}{\sqrt{1-2\epsilon}} - \frac{1}{\sqrt{1+2\epsilon}} \right). \label{eqn:tail_prob}
\end{align}

Using \eqref{eqn:Wt_as_0} and \eqref{eqn:tail_prob}, we conclude that for any arbitrary $\epsilon,\delta>0$,
\begin{align*}
\limsup_{n\to\infty}\mathbb{P}\left( \frac{N_1(n)}{n} \geqslant \frac{1}{2} + 2(\epsilon + \delta) \right) \leqslant \frac{1}{\delta}\limsup_{n\to\infty}\mathbb{P}\left( W_n \geqslant \frac{1}{\sqrt{1-2\epsilon}} - \frac{1}{\sqrt{1+2\epsilon}} \right) = 0.
\end{align*}
It therefore follows that for any $\epsilon^\prime > 0$, $\lim_{n\to\infty}\mathbb{P}\left( \frac{N_1(n)}{n} \geqslant \frac{1}{2} + \epsilon^\prime \right)=0$; equivalently, $\lim_{n\to\infty}\mathbb{P}\left( \frac{N_2(n)}{n} \leqslant \frac{1}{2} - \epsilon^\prime \right)=0$. Since arm~2 is inferior by assumption, it naturally holds that $\lim_{n\to\infty}\mathbb{P}\left( \frac{N_2(n)}{n} \geqslant \frac{1}{2} + \epsilon^\prime\right)=0$ (The steps in \ref{subsec:closing_the_loop} can be replicated near-identically for rigor). Thus, the stated assertion $\frac{N_i(n)}{n} \xrightarrow[n\to\infty]{p}\frac{1}{2}\ \forall\ i\in\{1,2\}$, follows. \hfill $\square$

\section{Proof of Theorem~\ref{thm:rates} in the ``moderate gap'' regime}
\label{sec:moderate_gap}

Firstly, note that the $\lambda_\rho^{\ast}(\theta)$ that solves \eqref{eqn:limit_eqn}, satisfies the following properties: (i) Continuous and monotone increasing in $\theta\geqslant 0$, (ii) $\lambda_\rho^{\ast}(\theta) \geqslant 1/2$ for all $\theta \geqslant 0$, (iii) $\lambda_\rho^{\ast}(0) = 1/2$ and $\lambda_\rho^{\ast}(\theta) \to 1$ as $\theta\to\infty$. \\

Secondly, because we are only interested in asymptotics, the $\Delta \sim \sqrt{\frac{\theta\log n}{n}}$ condition is as good as $\Delta = \sqrt{\frac{\theta\log n}{n}}$, since for any arbitrarily small $\epsilon^{\prime}>0$, $\Delta\in \left( \sqrt{\frac{\left(\theta-\epsilon^{\prime}\right)\log n}{n}}, \sqrt{\frac{\left(\theta+\epsilon^{\prime}\right)\log n}{n}} \right)$ for $n$ large enough; the stated assertion would follow in the limit as $\epsilon^{\prime}$ approaches $0$. In what follows, we will therefore assume for readability of the proof, and without loss of generality, that $\Delta = \sqrt{\frac{\theta\log n}{n}}$. \\ 

Thirdly, without loss of generality, suppose that arm~1 is optimal, i.e., $\mu_1 \geqslant \mu_2$.

\subsection{Focusing on arm~1}
\label{subsec:arm1}

Consider an arbitrary $\epsilon\in\left( 0, 1 - \lambda_\rho^{\ast}(\theta) \right)$, and define $u(n) := \left\lceil \left( \lambda_\rho^{\ast}(\theta) + \epsilon \right)n \right\rceil$. We know that
\begin{align}
N_1(n) &\leqslant u(n) + \sum_{t=u(n)+1}^{n}\mathbbm{1}\left\lbrace \pi_t = 1,\ N_1(t-1) \geqslant u(n) \right\rbrace\ \ \ \ \ \ \ \ \ \ \ \ \ \ \ \ \ \ \ \ \ \ \ \ \text{(this is always true)} \notag \\
&= u(n) + \sum_{t=u(n)}^{n-1}\mathbbm{1}\left\lbrace \pi_{t+1} = 1,\ N_1(t) \geqslant u(n) \right\rbrace \notag \\
&\leqslant u(n) + \sum_{t=u(n)}^{n-1}\mathbbm{1}\left\lbrace \pi_{t+1} = 1,\ N_1(t) \geqslant u(t) \right\rbrace \notag \\
&\leqslant u(n) + \sum_{t=u(n)}^{n-1}\mathbbm{1}\left\lbrace \bar{X}_1(t) - \bar{X}_2(t) \geqslant \sqrt{\rho\log t}\left( \frac{1}{\sqrt{N_2(t)}} - \frac{1}{\sqrt{N_1(t)}} \right),\ N_1(t) \geqslant u(t) \right\rbrace \notag \\
&= u(n) + \underbrace{\sum_{t=u(n)}^{n-1}\mathbbm{1}\left\lbrace \bar{Y}_1(t) - \bar{Y}_2(t) \geqslant \sqrt{\rho\log t}\left( \frac{1}{\sqrt{N_2(t)}} - \frac{1}{\sqrt{N_1(t)}} \right) - \Delta,\ N_1(t) \geqslant u(t) \right\rbrace}_{=: Z(n)}, \label{eqn:5}
\end{align}
where $\bar{Y}_i(t) := \frac{\sum_{j=1}^{N_i(t)}Y_{i,j}}{N_i(t)}$ with $Y_{i,j} := X_{i,j} - \mu_i$, $i\in\{1,2\}, j\in\mathbb{N}$. Clearly, $Y_{i,j}$'s are independent, zero-mean, and $Y_{i,j} \in \left[ -\mu_i, 1-\mu_i \right]\ \forall\ i\in\{1,2\}, j\in\mathbb{N}$.

\subsubsection{An almost sure lower bound on the arm-sampling rates}

Consider $n$ large enough such that $\sqrt{\frac{\log n}{n}}$ is monotone decreasing in $n$ ($n\geqslant 3$ suffices). This will enable the inequality in step $(\dag)$ below. From \eqref{eqn:5}, we have
\begin{align}
&\mathbb{E}Z(n) \notag \\
=\ &\sum_{t=u(n)}^{n-1}\mathbb{P}\left( \bar{Y}_1(t) - \bar{Y}_2(t) \geqslant \sqrt{\rho\log t}\left( \frac{1}{\sqrt{N_2(t)}} - \frac{1}{\sqrt{N_1(t)}} \right) - \Delta,\ N_1(t) \geqslant u(t) \right) \notag \\
=\ &\sum_{t=u(n)}^{n-1}\mathbb{P}\left( \bar{Y}_1(t) - \bar{Y}_2(t) \geqslant \sqrt{\rho\log t}\left( \frac{1}{\sqrt{N_2(t)}} - \frac{1}{\sqrt{N_1(t)}} \right) - \sqrt{\frac{\theta\log n}{n}},\ N_1(t) \geqslant u(t) \right) \notag \\
\underset{\mathrm{(\dag)}}{\leqslant}\ &\sum_{t=u(n)}^{n-1}\mathbb{P}\left( \bar{Y}_1(t) - \bar{Y}_2(t) \geqslant \sqrt{\rho\log t}\left( \frac{1}{\sqrt{N_2(t)}} - \frac{1}{\sqrt{N_1(t)}} \right) - \sqrt{\frac{\theta\log t}{t}},\ N_1(t) \geqslant u(t) \right) \notag \\
=\ &\sum_{t=u(n)}^{n-1}\mathbb{P}\left( \bar{Y}_1(t) - \bar{Y}_2(t) \geqslant \sqrt{\rho\log t}\left( \frac{1}{\sqrt{N_2(t)}} - \frac{1}{\sqrt{N_1(t)}} - \sqrt{\frac{\theta}{\rho t}}\right),\ N_1(t) \geqslant u(t) \right) \label{eqn:arbit_init} \\
\leqslant\ &\sum_{t=u(n)}^{n-1}\sum_{m=u(t)}^{t-1}\mathbb{P}\left( \frac{\sum_{j=1}^{m}Y_{1,j}}{m} - \frac{\sum_{j=1}^{t-m}Y_{2,j}}{t-m} \geqslant \sqrt{\rho \log t}\left( \frac{1}{\sqrt{t-m}} - \frac{1}{\sqrt{m}} - \sqrt{\frac{\theta}{\rho t}}\right)\right). \label{eqn:arbit}
\end{align}
Notice that in the interval $m\in\left[ u(t), t-1 \right]$,
\begin{align*}
\frac{1}{\sqrt{t-m}} - \frac{1}{\sqrt{m}} - \sqrt{\frac{\theta}{2t}} \geqslant \frac{1}{\sqrt{t-u(t)}} - \frac{1}{\sqrt{u(t)}} - \sqrt{\frac{\theta}{\rho t}} \geqslant \frac{1}{\sqrt{t}} \left( \frac{1}{\sqrt{1-\lambda_\rho^{\ast}(\theta)-\epsilon}} - \frac{1}{\sqrt{\lambda_\rho^{\ast}(\theta)+\epsilon}} - \sqrt{\frac{\theta}{\rho}} \right) > 0,
\end{align*}
where the final inequality follows since $\lambda_\rho^{\ast}(\theta)$ is the solution to \eqref{eqn:limit_eqn}. We can therefore apply the Chernoff-Hoeffding bound (Fact~\ref{hoeffding}) to \eqref{eqn:arbit} to conclude
\begin{align}
&\mathbb{E}Z(n) \notag \\
\leqslant\ &\sum_{t=u(n)}^{n-1}\sum_{m=u(t)}^{t-1}\exp \left[ -2\rho\log t \left( \frac{1}{\sqrt{t-m}} - \frac{1}{\sqrt{m}} - \sqrt{\frac{\theta}{\rho t}} \right)^2 \frac{m(t-m)}{t} \right] \notag \\
=\ &\sum_{t=u(n)}^{n-1}\sum_{m=u(t)}^{t-1}\exp \left[ -2\rho\log t \left( \sqrt{\frac{m}{t}} - \sqrt{1-\frac{m}{t}} - \sqrt{\frac{\theta}{\rho}}\sqrt{\frac{m}{t}\left(1 - \frac{m}{t}\right)} \right)^2 \right] \notag \\
=\ &\sum_{t=u(n)}^{n-1}\sum_{m=u(t)}^{t-1}\exp \left[ -2\rho\log t \left( f\left( {\frac{m}{t}} \right) \right)^2 \right], \label{eqn:arbit2}
\end{align}
where the function $f(x) := \sqrt{x} - \sqrt{1-x} - \sqrt{\theta x(1-x)/\rho}$. Notice that $f(x)$ is monotone increasing over the interval $\left(1/2, 1\right)$ $\left( \because \theta,\rho \geqslant 0 \right)$. Also, note that $1/2 < \lambda_\rho^{\ast}(\theta) + \epsilon < m/t < 1$ in \eqref{eqn:arbit2}. Thus, we have in \eqref{eqn:arbit2} that $f\left( {\frac{m}{t}} \right) \geqslant \min_{x\in\left[ \lambda_\rho^{\ast}(\theta)+\epsilon, 1\right)} f(x) = f\left( \lambda_\rho^{\ast}(\theta)+\epsilon \right)$. An expression for $f\left( \lambda_\rho^{\ast}(\theta)+\epsilon \right)$ is provided in \eqref{eqn:tempo} below. Observe that $f\left( \lambda_\rho^{\ast}(\theta)+\epsilon \right) > 0$; this follows since $\lambda_\rho^{\ast}(\theta)$ is the solution to \eqref{eqn:limit_eqn}. Using these facts in \eqref{eqn:arbit2}, we conclude
\begin{align}
\mathbb{E}Z(n) &\leqslant \sum_{t=u(n)}^{n-1}\sum_{m=u(t)}^{t-1}\exp \left[ -2\rho\log t \left( \min_{x\in\left[ \lambda_\rho^{\ast}(\theta)+\epsilon, 1\right)} f(x) \right)^2 \right] \notag \\
&= \sum_{t=u(n)}^{n-1}\sum_{m=u(t)}^{t-1}\exp \left[ -2\rho\log t \left( f\left( \lambda_\rho^{\ast}(\theta)+\epsilon \right) \right)^2 \right] \notag \\
&\leqslant \sum_{t=u(n)}^{n-1}t^{1 -2\rho\left( f\left( \lambda_\rho^{\ast}(\theta)+\epsilon \right) \right)^2} \notag \\
&= \sum_{t=\left\lceil \left( \lambda_\rho^{\ast}(\theta) + \epsilon \right)n \right\rceil}^{n-1}t^{1 -2\rho\left( f\left( \lambda_\rho^{\ast}(\theta)+\epsilon \right) \right)^2}. \label{eqn:arbit3}
\end{align}

Now consider an arbitrary $\delta>0$. We then have
\begin{align}
\mathbb{P}\left( N_1(n) - u(n) \geqslant \delta n\right) &\leqslant \mathbb{P}\left( Z(n) \geqslant \delta n\right) & \mbox{(using \eqref{eqn:5})} \notag \\
&\leqslant \frac{\mathbb{E}Z(n)}{\delta n} & \mbox{(Markov's inequality)} \notag \\ 
\implies \mathbb{P}\left( N_1(n) \geqslant \left\lceil \left( \lambda_\rho^{\ast}(\theta) + \epsilon \right)n \right\rceil + \delta n \right) &\leqslant \frac{1}{\delta n}\sum_{t=\left\lceil \left( \lambda_\rho^{\ast}(\theta) + \epsilon \right)n \right\rceil}^{n-1}t^{1 -2\rho\left( f\left( \lambda_\rho^{\ast}(\theta)+\epsilon \right) \right)^2}. & \mbox{(using \eqref{eqn:arbit3})} \label{eqn:arbit4} 
\end{align}

Note that $f\left( \lambda_\rho^{\ast}(\theta)+\epsilon \right)$ is given by
\begin{align}
f\left( \lambda_\rho^{\ast}(\theta)+\epsilon \right) = \sqrt{\lambda_\rho^{\ast}(\theta)+\epsilon} - \sqrt{1-\lambda_\rho^{\ast}(\theta)-\epsilon} - \sqrt{\frac{\theta}{\rho}}\sqrt{\left( \lambda_\rho^{\ast}(\theta)+\epsilon \right)\left( 1-\lambda_\rho^{\ast}(\theta)-\epsilon \right)}. \label{eqn:tempo}
\end{align}
Setting $\epsilon=0$ in \eqref{eqn:tempo} yields $f\left( \lambda_\rho^{\ast}(\theta)\right) = 0$ (follows from \eqref{eqn:limit_eqn}), whereas setting $\epsilon=1 - \lambda_\rho^{\ast}(\theta)$ yields $f\left( 1 \right) = 1$. Since $\rho>1$, and $f\left( \lambda_\rho^{\ast}(\theta)+\epsilon \right)$ is continuous and monotone increasing in $\epsilon$, $\exists\ \epsilon_{\theta,\rho}\in\left( 0, 1 - \lambda^{\ast}_\rho(\theta) \right)$ s.t. $f\left( \lambda^{\ast}_\rho(\theta)+\epsilon \right) > 1/\sqrt{\rho}$ for $\epsilon \geqslant \epsilon_{\theta,\rho}$. Substituting $\epsilon = \epsilon_{\theta,\rho}$ in \eqref{eqn:arbit4} and using the aforementioned fact, we obtain
\begin{align}
\mathbb{P}\left( N_1(n) \geqslant \left\lceil \left( \lambda_\rho^{\ast}(\theta) + \epsilon_{\theta,\rho} \right)n \right\rceil + \delta n \right) &\leqslant \frac{1}{\delta n}\sum_{t=\left\lceil \left( \lambda_\rho^{\ast}(\theta) + \epsilon_{\theta,\rho} \right)n \right\rceil}^{n-1}t^{1- 2\rho\left( f\left( \lambda^{\ast}_\rho(\theta)+\epsilon_{\theta,\rho} \right) \right)^2} \notag \\
&\leqslant \left(\frac{2^{2\rho-1}}{\delta}\right)n^{-\left( 2\rho\left( f\left( \lambda^{\ast}_\rho(\theta)+\epsilon_{\theta,\rho} \right) \right)^2-1 \right)}, \label{eqn:yahoo}
\end{align}
where the last inequality follows since $1/\sqrt{\rho} < f\left( \lambda^{\ast}_\rho(\theta)+\epsilon_{\theta,\rho} \right) < 1$, and $\lambda^{\ast}_\rho(\theta)+\epsilon_{\theta,\rho} > 1/2$. Finally since $\delta>0$ is arbitrary, we conclude from \eqref{eqn:yahoo} using the Borel-Cantelli Lemma that 
\begin{align*}
\limsup_{n\to\infty} \frac{N_1(n)}{n} \leqslant \lambda_\rho^{\ast}\left( \theta \right) + \epsilon_{\theta,\rho} < 1\ \ \ \text{w.p. $1$}.
\end{align*}
The above result naturally holds for arm~2 as well, since it is inferior by assumption (we resort to the cop-out that a near-identical argument handles its case). Therefore, in conclusion,
\begin{align}
\liminf_{n\to\infty} \frac{N_i(n)}{n} \geqslant 1 - \lambda_\rho^{\ast}\left( \theta \right) - \epsilon_{\theta,\rho} > 0\ \ \ \text{w.p. $1$}\ \ \ \forall\ i\in\{1,2\}. \label{eqn:3yo}
\end{align}

\subsubsection{Closing the loop}

From \eqref{eqn:arbit_init}, we know that
\begin{align}
&\mathbb{E}Z(n) \notag \\
\leqslant\ &\sum_{t=u(n)}^{n-1}\mathbb{P}\left( \bar{Y}_1(t) - \bar{Y}_2(t) \geqslant \sqrt{\rho\log t}\left( \frac{1}{\sqrt{N_2(t)}} - \frac{1}{\sqrt{N_1(t)}} - \sqrt{\frac{\theta}{\rho t}}\right),\ N_1(t) \geqslant u(t) \right) \notag \\
\leqslant\ &\sum_{t=u(n)}^{n-1}\mathbb{P}\left( \bar{Y}_1(t) - \bar{Y}_2(t) \geqslant \sqrt{\frac{\rho\log t}{t}}\left( \frac{1}{\sqrt{1-\lambda_\rho^{\ast}(\theta)-\epsilon}} - \frac{1}{\sqrt{\lambda_\rho^{\ast}(\theta)+\epsilon}} - \sqrt{\frac{\theta}{\rho}} \right) \right) \notag \\
=\ &\sum_{t=u(n)}^{n-1}\mathbb{P}\left( \underbrace{\sqrt{\frac{t}{\rho\log t}} \left( \bar{Y}_1(t) - \bar{Y}_2(t) \right)}_{=:W_t} \geqslant \frac{1}{\sqrt{1-\lambda_\rho^{\ast}(\theta)-\epsilon}} - \frac{1}{\sqrt{\lambda_\rho^{\ast}(\theta)+\epsilon}} - \sqrt{\frac{\theta}{\rho}} \right), \label{eqn:Wt0arbit}
\end{align}
where we already know that $\frac{1}{\sqrt{1-\lambda_\rho^{\ast}(\theta)-\epsilon}} - \frac{1}{\sqrt{\lambda_\rho^{\ast}(\theta)+\epsilon}} - \sqrt{\frac{\theta}{\rho}}>0$ (since $\lambda_\rho^{\ast}(\theta)$ is the solution to \eqref{eqn:limit_eqn}). Now,
\begin{align}
&\left\lvert W_t \right\rvert \notag \\
\leqslant\ &\sqrt{\frac{t}{\rho\log t}} \left( \left\lvert\frac{\sum_{j=1}^{N_1(t)}Y_{1,j}}{N_1(t)}\right\rvert + \left\lvert\frac{\sum_{j=1}^{N_2(t)}Y_{2,j}}{N_2(t)}\right\rvert \right) \notag \\
=\ &\sqrt{\frac{2t}{\rho\log t}} \left( \sqrt{\frac{\log \log N_1(t)}{N_1(t)}}\left\lvert\frac{\sum_{j=1}^{N_1(t)}Y_{1,j}}{\sqrt{2N_1(t)\log \log N_1(t)}}\right\rvert + \sqrt{\frac{\log \log N_2(t)}{N_2(t)}}\left\lvert\frac{\sum_{j=1}^{N_2(t)}Y_{2,j}}{\sqrt{2N_2(t)\log \log N_2(t)}}\right\rvert \right) \notag \\
\leqslant\ &\sqrt{\frac{2t}{\rho\log t}} \left( \sqrt{\frac{\log \log t}{N_1(t)}}\left\lvert\frac{\sum_{j=1}^{N_1(t)}Y_{1,j}}{\sqrt{2N_1(t)\log \log N_1(t)}}\right\rvert + \sqrt{\frac{\log \log t}{N_2(t)}}\left\lvert\frac{\sum_{j=1}^{N_2(t)}Y_{2,j}}{\sqrt{2N_2(t)\log \log N_2(t)}}\right\rvert \right) \notag \\
=\ &\sqrt{\frac{2\log \log t}{\rho\log t}} \left( \sqrt{\frac{t}{N_1(t)}}\left\lvert\frac{\sum_{j=1}^{N_1(t)}Y_{1,j}}{\sqrt{2N_1(t)\log \log N_1(t)}}\right\rvert + \sqrt{\frac{t}{N_2(t)}}\left\lvert\frac{\sum_{j=1}^{N_2(t)}Y_{2,j}}{\sqrt{2N_2(t)\log \log N_2(t)}}\right\rvert \right). \label{eqn:Wtyo}
\end{align}

We know that $N_i(t)$, for both arms~$i\in\{1,2\}$, can be lower bounded \emph{path-wise} by a deterministic monotone increasing function of $t$, say $g(t)$, that grows to $+\infty$ as $t\to\infty$. This is a trivial consequence of the structure of the canonical UCB policy (Algorithm~\ref{alg:UCB}), and the fact that the rewards are uniformly bounded. Therefore, for any arm~$i\in\{1,2\}$, we have 
\begin{align*}
\left\lvert\frac{\sum_{j=1}^{N_i(t)}Y_{i,j}}{\sqrt{2N_i(t)\log \log N_i(t)}}\right\rvert \leqslant \sup_{m\geqslant g(t)} \left\lvert\frac{\sum_{j=1}^{m}Y_{i,j}}{\sqrt{2m\log \log m}}\right\rvert.
\end{align*}

For a fixed $i\in\{1,2\}$, $\left\lbrace Y_{i,j} : j\in\mathbb{N} \right\rbrace$ is a collection of i.i.d. random variables with $\mathbb{E}Y_{i,1} = 0$ and $\text{Var}\left( Y_{i,1} \right) = \text{Var}\left( X_{i,1} \right) \leqslant 1$. Also, $g(t)$ is a monotone increasing and coercive function of $t$. Therefore, the \emph{Law of the Iterated Logarithm} (see \cite{durrett2019probability}, Theorem~8.5.2) implies
\begin{align}
\limsup_{t\to\infty} \left\lvert\frac{\sum_{j=1}^{N_i(t)}Y_{i,j}}{\sqrt{2N_i(t)\log \log N_i(t)}}\right\rvert \leqslant 1\ \ \ \text{w.p.}\ 1\ \ \forall\ i\in\{1,2\}. \label{eqn:LILyo}
\end{align}

Using \eqref{eqn:3yo}, \eqref{eqn:Wtyo} and \eqref{eqn:LILyo}, we conclude that
\begin{align}
\lim_{t\to\infty} W_t = 0\ \ \ \text{w.p.}\ 1. \label{eqn:Wt_as_0yo}
\end{align}

Now consider an arbitrary $\delta>0$. We have
\begin{align}
\mathbb{P}\left( N_1(n) - u(n) \geqslant \delta n \right) &\leqslant \mathbb{P}\left( Z(n) \geqslant \delta n\right) & \mbox{(using \eqref{eqn:5})} \notag \\
&\leqslant \frac{\mathbb{E}Z(n)}{\delta n} & \mbox{(Markov's inequality)} \notag \\
&\leqslant \frac{1}{\delta n}\sum_{t=u(n)}^{n-1}\mathbb{P}\left( W_t \geqslant \frac{1}{\sqrt{1-\lambda_\rho^{\ast}(\theta)-\epsilon}} - \frac{1}{\sqrt{\lambda_\rho^{\ast}(\theta)+\epsilon}} - \sqrt{\frac{\theta}{\rho}} \right) & \mbox{(using \eqref{eqn:Wt0arbit})} \notag \\
&\leqslant \frac{1}{\delta}\sup_{t > n/2}\mathbb{P}\left( W_t \geqslant \frac{1}{\sqrt{1-\lambda_\rho^{\ast}(\theta)-\epsilon}} - \frac{1}{\sqrt{\lambda_\rho^{\ast}(\theta)+\epsilon}} - \sqrt{\frac{\theta}{\rho}} \right). \label{eqn:tail_probyo}
\end{align}

Using \eqref{eqn:Wt_as_0yo} and \eqref{eqn:tail_probyo}, it follows that
\begin{align*}
\limsup_{n\to\infty}\mathbb{P}\left( N_1(n) - u(n) \geqslant \delta n \right) \leqslant \frac{1}{\delta}\limsup_{n\to\infty}\mathbb{P}\left( W_n \geqslant \frac{1}{\sqrt{1-\lambda_\rho^{\ast}(\theta)-\epsilon}} - \frac{1}{\sqrt{\lambda_\rho^{\ast}(\theta)+\epsilon}} - \sqrt{\frac{\theta}{\rho}} \right) = 0.
\end{align*}
Since $u(n) = \left\lceil \left( \lambda_\rho^{\ast}(\theta) + \epsilon \right)n \right\rceil$ and $\epsilon,\delta>0$ are arbitrary, we conclude that for any $\epsilon>0$, it holds that $\lim_{n\to\infty}\mathbb{P}\left( \frac{N_1(n)}{n} \geqslant \lambda_\rho^{\ast}(\theta) + \epsilon\right)=0$. Equivalently, $\lim_{n\to\infty}\mathbb{P}\left( \frac{N_2(n)}{n} \leqslant 1-\lambda_\rho^{\ast}(\theta) - \epsilon\right)=0$ holds for any $\epsilon>0$. \hfill $\square$

\subsection{Focusing on arm~2 and concluding}

We will essentially replicate here the proof for arm~1 given in \ref{subsec:arm1}, albeit with a few subtle modifications to account for the fact that arm~2 is inferior. Consistent with previous approach and notation, we consider an arbitrary $\epsilon\in\left( 0, \lambda_\rho^{\ast}(\theta) \right)$ and set $u(n) := \left\lceil \left( 1 - \lambda_\rho^{\ast}(\theta) + \epsilon \right)n \right\rceil$, where $\lambda_\rho^{\ast}(\theta)$ is the solution to \eqref{eqn:limit_eqn} (Note that the definition of $u(n)$ here is different from the one used in the proof for arm~1.). We know that
\begin{align}
N_2(n) &\leqslant u(n) + \sum_{t=u(n)+1}^{n}\mathbbm{1}\left\lbrace \pi_t = 2,\ N_2(t-1) \geqslant u(n) \right\rbrace\ \ \ \ \ \ \ \ \ \ \ \ \ \ \ \ \ \ \ \ \ \ \ \ \text{(this is always true)} \notag \\
&= u(n) + \sum_{t=u(n)}^{n-1}\mathbbm{1}\left\lbrace \pi_{t+1} = 2,\ N_2(t) \geqslant u(n) \right\rbrace \notag \\
&\leqslant u(n) + \sum_{t=u(n)}^{n-1}\mathbbm{1}\left\lbrace \bar{X}_2(t) - \bar{X}_1(t) \geqslant \sqrt{\rho\log t}\left( \frac{1}{\sqrt{N_1(t)}} - \frac{1}{\sqrt{N_2(t)}} \right),\ N_2(t) \geqslant u(n) \right\rbrace \notag \\
&= u(n) + \underbrace{\sum_{t=u(n)}^{n-1}\mathbbm{1}\left\lbrace \bar{Y}_2(t) - \bar{Y}_1(t) \geqslant \sqrt{\rho\log t}\left( \frac{1}{\sqrt{N_1(t)}} - \frac{1}{\sqrt{N_2(t)}} \right) + \Delta,\ N_2(t) \geqslant u(n) \right\rbrace}_{=: Z(n)}, \label{eqn:arm2_init}
\end{align}
where $\bar{Y}_i(t) := \frac{\sum_{j=1}^{N_i(t)}Y_{i,j}}{N_i(t)}$ with $Y_{i,j} := X_{i,j} - \mu_i$, $i\in\{1,2\}, j\in\mathbb{N}$ (Notice that these definitions of $\bar{Y}_i(t)$ and $Y_{i,j}$ are identical to their counterparts from the proof for arm~1.). From \eqref{eqn:arm2_init}, it follows that 
\begin{align}
&\mathbb{E}Z(n) \notag \\
=\ &\sum_{t=u(n)}^{n-1}\mathbb{P}\left( \bar{Y}_2(t) - \bar{Y}_1(t) \geqslant \sqrt{\rho\log t}\left( \frac{1}{\sqrt{N_1(t)}} - \frac{1}{\sqrt{N_2(t)}} \right) + \Delta,\ N_2(t) \geqslant u(n) \right) \notag \\
=\ &\sum_{t=u(n)}^{n-1}\mathbb{P}\left( \bar{Y}_2(t) - \bar{Y}_1(t) \geqslant \sqrt{\rho\log t}\left( \frac{1}{\sqrt{N_1(t)}} - \frac{1}{\sqrt{N_2(t)}} \right) + \sqrt{\frac{\theta\log n}{n}},\ N_2(t) \geqslant u(n) \right) \notag \\
\leqslant\ &\sum_{t=u(n)}^{n-1}\mathbb{P}\left( \bar{Y}_2(t) - \bar{Y}_1(t) \geqslant \sqrt{\rho\log t}\left( \frac{1}{\sqrt{t-u(n)}} - \frac{1}{\sqrt{u(n)}} \right) + \sqrt{\frac{\theta\log n}{n}} \right) \notag \\
\leqslant\ &\sum_{t=u(n)}^{n-1}\mathbb{P}\left( \bar{Y}_2(t) - \bar{Y}_1(t) \geqslant \sqrt{\rho\log t}\left( \frac{1}{\sqrt{n-u(n)}} - \frac{1}{\sqrt{u(n)}} \right) + \sqrt{\frac{\theta\log t}{n}} \right) \notag \\
\leqslant\ &\sum_{t=u(n)}^{n-1}\mathbb{P}\left( \bar{Y}_2(t) - \bar{Y}_1(t) \geqslant \sqrt{\frac{\rho\log t}{n}}\left( \frac{1}{\sqrt{\lambda_\rho^{\ast}(\theta) - \epsilon}} - \frac{1}{\sqrt{1 - \lambda_\rho^{\ast}(\theta) + \epsilon}} + \sqrt{\frac{\theta}{\rho}}\right) \right), \notag
\end{align}
where $\frac{1}{\sqrt{\lambda_\rho^{\ast}(\theta) - \epsilon}} - \frac{1}{\sqrt{1 - \lambda_\rho^{\ast}(\theta) + \epsilon}} + \sqrt{\frac{\theta}{\rho}} > 0$ is guaranteed since $\lambda_\rho^{\ast}(\theta)$ is the solution to \eqref{eqn:limit_eqn}. Also, $t \geqslant u(n) = \left\lceil \left( 1 - \lambda_\rho^{\ast}(\theta) + \epsilon \right)n \right\rceil \implies n \leqslant \frac{t}{1 - \lambda_\rho^{\ast}(\theta) + \epsilon}$. Therefore,
\begin{align}
&\mathbb{E}Z(n) \notag \\
\leqslant\ &\sum_{t=u(n)}^{n-1}\mathbb{P}\left( \bar{Y}_2(t) - \bar{Y}_1(t) \geqslant \sqrt{1 - \lambda_\rho^{\ast}(\theta) + \epsilon}\sqrt{\frac{\rho\log t}{t}}\left( \frac{1}{\sqrt{\lambda_\rho^{\ast}(\theta) - \epsilon}} - \frac{1}{\sqrt{1 - \lambda_\rho^{\ast}(\theta) + \epsilon}} + \sqrt{\frac{\theta}{\rho}}\right) \right) \notag \\
=\ &\sum_{t=u(n)}^{n-1}\mathbb{P}\left( \underbrace{\sqrt{\frac{t}{\rho\log t}}\left( \bar{Y}_2(t) - \bar{Y}_1(t) \right)}_{=:W_t} \geqslant \underbrace{\sqrt{1 - \lambda_\rho^{\ast}(\theta) + \epsilon}\left( \frac{1}{\sqrt{\lambda_\rho^{\ast}(\theta) - \epsilon}} - \frac{1}{\sqrt{1 - \lambda_\rho^{\ast}(\theta) + \epsilon}} + \sqrt{\frac{\theta}{\rho}}\right)}_{=:\varepsilon(\theta,\rho,\epsilon)\ \left(\text{Note that}\ \varepsilon(\theta,\rho,\epsilon)>0\ \text{for any}\ \epsilon\in\left( 0, \lambda_\rho^{\ast}(\theta) \right) \right)} \right). \label{eqn:arm2_1}
\end{align}

Recall that we have already handled $W_t$ (albeit a negated version thereof) in the proof for arm~1 in \eqref{eqn:Wt0arbit} and shown that $W_t \to 0$ almost surely in \eqref{eqn:Wt_as_0yo}. Now consider an arbitrary $\delta>0$. We then have
\begin{align}
\mathbb{P}\left( N_2(n) - u(n) \geqslant \delta n \right) &\leqslant \mathbb{P}\left( Z(n) \geqslant \delta n\right) & \mbox{(using \eqref{eqn:arm2_init})} \notag \\
&\leqslant \frac{\mathbb{E}Z(n)}{\delta n} & \mbox{(Markov's inequality)} \notag \\
&\leqslant \frac{1}{\delta n}\sum_{t=u(n)}^{n-1}\mathbb{P}\left( W_t \geqslant \varepsilon(\theta,\rho,\epsilon) \right) & \mbox{(using \eqref{eqn:arm2_1})} \notag \\
&\leqslant \frac{1}{\delta}\sup_{t > \left( 1 - \lambda_\rho^{\ast}(\theta) \right)n}\mathbb{P}\left( W_t \geqslant \varepsilon(\theta,\rho,\epsilon) \right). \label{eqn:arm2_tail_probyo}
\end{align}

Taking limits on both sides of \eqref{eqn:arm2_tail_probyo}, we obtain
\begin{align*}
\limsup_{n\to\infty}\mathbb{P}\left( N_2(n) - u(n) \geqslant \delta n \right) \leqslant \frac{1}{\delta}\limsup_{n\to\infty}\mathbb{P}\left( W_n \geqslant \varepsilon(\theta,\rho,\epsilon) \right) = 0,
\end{align*}
where the final conclusion follows since $W_n \to 0$ almost surely, and hence also in probability. Now since $u(n) = \left\lceil \left( 1 - \lambda_\rho^{\ast}(\theta) + \epsilon \right)n \right\rceil$ and $\epsilon,\delta>0$ are arbitrary, it follows that for any $\epsilon>0$, we have $\lim_{n\to\infty}\mathbb{P}\left( \frac{N_2(n)}{n} \geqslant 1 - \lambda_\rho^{\ast}(\theta) + \epsilon\right)=0$. From the proof for arm~$1$, we already know that $\lim_{n\to\infty}\mathbb{P}\left( \frac{N_2(n)}{n} \leqslant 1 - \lambda_\rho^{\ast}(\theta) - \epsilon\right)=0$ holds for any $\epsilon>0$. Therefore, it must be the case that $\frac{N_2(n)}{n}\xrightarrow[n\to\infty]{p}1-\lambda_\rho^{\ast}(\theta)$ and $\frac{N_1(n)}{n}\xrightarrow[n\to\infty]{p}\lambda_\rho^{\ast}(\theta)$, as desired. \hfill $\square$

\section{Proof of Theorem~\ref{thm:zero-gap-K-arm}}
\label{proof:K-arm}

We will prove this result in two parts; the preamble in \ref{subsec:preamble} below will prove a meta-result stating that $N_i(n)/n > 1/\left(2\left\lvert \mathcal{I} \right\rvert \right)$ with high probability (approaching $1$ as $n\to\infty$) for any arm~$i\in\mathcal{I}$. We will then leverage this meta-result to prove the assertions of the theorem in \ref{subsec:main}. 

\subsection{Preamble}
\label{subsec:preamble}

Let $L := |\mathcal{I}|$. If $L=1$, the result follows trivially from the standard logarithmic bound for the expected regret (Theorem~7 in \cite{audibert2009exploration}), followed by Markov's inequality. Therefore, without loss of generality, suppose that $|\mathcal{I}| \geqslant 2$, and fix an arbitrary arm $i\in\mathcal{I}$. Then, we know that the following is true for any integer $u>1$:
\begin{align*}
N_i(n) &\leqslant u + \sum_{t=u}^{n-1}\mathbbm{1} \left\lbrace \pi_{t+1} = i,\ N_i(t) \geqslant u \right\rbrace \\
&\leqslant u + \sum_{t=u}^{n-1}\mathbbm{1} \left\lbrace \pi_{t+1} = i,\ N_i(t) \geqslant u,\ \sum_{j\in\mathcal{I}\backslash\{i\}}N_j(t) \leqslant t-u\right\rbrace,
\end{align*}
where $\pi_{t+1}\in[K]$ indicates the arm played at time $t+1$. In particular, the above holds also for $u= \left\lceil \left( \frac{1}{L}+\epsilon\right) n \right\rceil$, where $\epsilon\in\left( 0, \frac{L-1}{L} \right)$ is arbitrarily chosen. We will fix this $u$ going forward, even though we may not always express its value explicitly for readability of the analysis that follows. We thus have
\begin{align}
N_i(n) &\leqslant u +  \sum_{t=u}^{n-1}\mathbbm{1} \left\lbrace B_{i,N_i(t),t} \geqslant \max_{\hat{j}\in[K]\backslash\{i\}}B_{\hat{j},N_{\hat{j}}(t),t},\ N_i(t) \geqslant u,\ \sum_{j\in\mathcal{I}\backslash\{i\}}N_j(t) \leqslant t-u\right\rbrace \notag \\
&\leqslant u +  \sum_{t=u}^{n-1}\mathbbm{1} \left\lbrace B_{i,N_i(t),t} \geqslant \max_{\hat{j}\in\mathcal{I}\backslash\{i\}}B_{\hat{j},N_{\hat{j}}(t),t},\ N_i(t) \geqslant u,\ \sum_{j\in\mathcal{I}\backslash\{i\}}N_j(t) \leqslant t-u\right\rbrace, \label{eqn:hihihahaha}
\end{align}
where $B_{k,s,t} := \hat{X}_k(s) + \sqrt{(\rho\log t)/s}$ for $k\in[K]$, and \emph{$\hat{X}_k(s):= {\sum_{l=1}^{s}X_{k,l}}/{s}$ denotes the empirical mean reward from the ``first $s$ plays'' of arm $k$} (Note the distinction from $\bar{X}_k(s)$, which has been defined before as the empirical mean reward of arm~$k$ ``\emph{at time $s$},'' i.e., mean over its ``first $N_k(s)$ plays''). Now observe that
\begin{align}
\left\lbrace \sum_{j\in\mathcal{I}\backslash\{i\}}N_j(t) \leqslant t-u \right\rbrace \subseteq \left\lbrace \exists\ j\in\mathcal{I}\backslash\{i\} : N_j(t) \leqslant \frac{t-u}{L-1} \right\rbrace \subseteq \left\lbrace \exists\ j\in\mathcal{I}\backslash\{i\} : N_j(t) \leqslant \left( \frac{1}{L} - \frac{\epsilon}{L-1}\right)t \right\rbrace, \label{eqn:huhuhahaha}
\end{align}
where the last inclusion follows using $u= \left\lceil \left( \frac{1}{L}+\epsilon\right) n \right\rceil$ and $n\geqslant t$. Combining \eqref{eqn:hihihahaha} and \eqref{eqn:huhuhahaha} using the Union bound, we obtain
\begin{align}
N_i(n) &\leqslant u +  \sum_{t=u}^{n-1}\sum_{j\in\mathcal{I}\backslash\{i\}}\mathbbm{1} \left\lbrace B_{i,N_i(t),t} \geqslant \max_{\hat{j}\in\mathcal{I}\backslash\{i\}}B_{\hat{j},N_{\hat{j}}(t),t},\ N_i(t) \geqslant u,\ N_j(t) \leqslant \left( \frac{1}{L} - \frac{\epsilon}{L-1}\right)t\right\rbrace \notag \\
&\leqslant u +  \sum_{t=u}^{n-1}\sum_{j\in\mathcal{I}\backslash\{i\}}\mathbbm{1} \left\lbrace B_{i,N_i(t),t} \geqslant B_{j,N_j(t),t},\ N_i(t) \geqslant u,\ N_j(t) \leqslant \left( \frac{1}{L} - \frac{\epsilon}{L-1}\right)t\right\rbrace \notag \\
&\leqslant u +  \underbrace{\sum_{t=u}^{n-1}\sum_{j\in\mathcal{I}\backslash\{i\}}\mathbbm{1} \left\lbrace B_{i,N_i(t),t} \geqslant B_{j,N_j(t),t},\ N_i(t) \geqslant \left( \frac{1}{L} + \epsilon\right)t,\ N_j(t) \leqslant \left( \frac{1}{L} - \frac{\epsilon}{L-1}\right)t\right\rbrace}_{=: Z_n}, \label{eqn:Zn_def}
\end{align}
where the last inequality again uses $u= \left\lceil \left( \frac{1}{L}+\epsilon\right) n \right\rceil$ and $n\geqslant t$. Define the events: \\ $E_i := \left\lbrace N_i(t) \geqslant \left( \frac{1}{L}+\epsilon\right)t \right\rbrace$, and $E_j := \left\lbrace N_j(t) \leqslant \left( \frac{1}{L} - \frac{\epsilon}{L-1}\right)t \right\rbrace$. Now,
\begin{align}
&\mathbb{E}Z_n \notag \\
= &\sum_{t=u}^{n-1}\sum_{j\in\mathcal{I}\backslash\{i\}}\mathbb{P} \left( B_{i,N_i(t),t} \geqslant B_{j,N_j(t),t},\ E_i,\ E_j\right) \notag \\
= &\sum_{t=u}^{n-1}\sum_{j\in\mathcal{I}\backslash\{i\}}\mathbb{P} \left( \hat{Y}_{i}\left( N_i(t) \right) - \hat{Y}_{j}\left( N_j(t) \right) \geqslant \sqrt{\rho\log t}\left( \frac{1}{\sqrt{N_j(t)}} - \frac{1}{\sqrt{N_i(t)}} \right),\ E_i,\ E_j \right), \label{eqn:continue_from_here}
\end{align}
where $\hat{Y}_k(s) := {\sum_{l=1}^{s}Y_{k,l}}/{s}$ and $Y_{k,l} := X_{k,l} - \mathbb{E}X_{k,l}$ for $k\in[K]$, $s\in\mathbb{N}$, $l\in\mathbb{N}$. The last equality above follows since $i,j\in\mathcal{I}$ and the mean rewards of arms in $\mathcal{I}$ are equal. Thus,
\begin{align*}
\mathbb{E}Z_n \leqslant \sum_{t=u}^{n-1}\sum_{j\in\mathcal{I}\backslash\{i\}}\sum_{m_i = \left\lceil \left( \frac{1}{L} + \epsilon\right)t \right\rceil}^{t}\sum_{m_j=1}^{\left\lfloor \left( \frac{1}{L} - \frac{\epsilon}{L-1} \right)t \right\rfloor}\mathbb{P} \left( \hat{Y}_{i}\left( m_i \right) - \hat{Y}_{j}\left( m_j \right) \geqslant \sqrt{\rho\log t}\left( \frac{1}{\sqrt{m_j}} - \frac{1}{\sqrt{m_i}} \right) \right).
\end{align*}
Since $\mathbb{E}\left[ \hat{Y}_{i}\left( m_i \right) - \hat{Y}_{j}\left( m_j \right) \right]=0$, and $m_j<m_i$ over the range of the summation above, we can use the Chernoff-Hoeffding bound (Fact~\ref{hoeffding}) to obtain
\begin{align}
\mathbb{E}Z_n \leqslant \sum_{t=u}^{n-1}\sum_{j\in\mathcal{I}\backslash\{i\}}\sum_{m_i = \left\lceil \left( \frac{1}{L} + \epsilon\right)t \right\rceil}^{t}\sum_{m_j=1}^{\left\lfloor \left( \frac{1}{L} - \frac{\epsilon}{L-1} \right)t \right\rfloor} \exp \left[ -2\rho\left( 1 - 2\sqrt{\frac{m_im_j}{\left( m_i+m_j \right)^2}} \right)\log t \right]. \label{eqn:sumtoyu}
\end{align}
Let $\gamma := m_i/(m_i+m_j)$. Then, $\gamma \geqslant \frac{\frac{1}{L}+\epsilon}{\frac{2}{L}+\left(\frac{L-2}{L-1}\right)\epsilon} > 1/2$ over the range of the summation in \eqref{eqn:sumtoyu}. Consequently, $\gamma(1-\gamma)$ is maximized at $\gamma = \frac{\frac{1}{L}+\epsilon}{\frac{2}{L}+\left(\frac{L-2}{L-1}\right)\epsilon}$, and therefore, $m_im_j/\left(m_i+m_j\right)^2 = \gamma(1-\gamma) \leqslant \left( f(\epsilon,L) \right)^2 < 1/4$ in \eqref{eqn:sumtoyu}, where $f(\epsilon,L)$ as defined as:
\begin{align}
f(\epsilon,L) := \sqrt{\frac{(L-1)(1+\epsilon L)\left( L-1-\epsilon L \right)}{\left( 2(L-1) + L(L-2)\epsilon \right)^2}}. \label{eqn:f_def}
\end{align}
Combining \eqref{eqn:sumtoyu} and \eqref{eqn:f_def}, we obtain
\begin{align}
\mathbb{E}Z_n \leqslant (L-1)\sum_{t=u}^{n-1} t^{-2\left( \rho - 1 - 2\rho f(\epsilon,L) \right)}. \label{eqn:EZnnew}
\end{align}
Now consider an arbitrary $\delta>0$. From \eqref{eqn:Zn_def}, we have
\begin{align}
\mathbb{P}\left( N_i(n) \geqslant u + \delta n \right) \leqslant \mathbb{P}\left( Z_n \geqslant \delta n \right) \underset{\mathrm{(\star)}}{\leqslant} \frac{\mathbb{E}Z_n}{\delta n} \underset{\mathrm{(\dag)}}{\leqslant} \left(\frac{L-1}{\delta n}\right) \sum_{t=u}^{n-1} t^{-2\left( \rho\left(1 - 2 f(\epsilon,L) \right)- 1 \right)}, \label{eqn:finalyoyu}
\end{align}
where $(\star)$ is due to Markov's inequality, and $(\dag)$ follows using \eqref{eqn:EZnnew}. Observe from \eqref{eqn:f_def} that $f(\epsilon,L)$ is monotone decreasing in $\epsilon$ over the interval $\epsilon \in \left[ 0, \frac{L-1}{L} \right]$, with $f(0,L) = 1/2$ and $f\left( \frac{L-1}{L}, L \right) = 0$. Therefore, $1 - 2f(\epsilon,L)>0$ in the interval $\epsilon \in \left( 0, \frac{L-1}{L} \right]$. Thus, for $\rho$ large enough, the exponent of $t$ in \eqref{eqn:finalyoyu} can be made arbitrarily small. That is, $\exists$ $\rho_0\in\mathbb{R}_+$ s.t. for all $\rho\geqslant \rho_0$, we have $2\left( \rho\left(1 - 2 f(\epsilon,L) \right)- 1 \right) > 0$ $\forall\ \epsilon\in\left[ \frac{1}{2L(L-1)}, \frac{L-1}{L} \right]$. Now supposing $\rho \geqslant \rho_0$, plug in $\epsilon = \frac{1}{2L(L-1)}$ in \eqref{eqn:finalyoyu} (this includes substituting $u=\left\lceil \left( \frac{1}{L} + \frac{1}{2L(L-1)} \right)n \right\rceil$). Then since $\delta>0$ and $i\in\mathcal{I}$ are arbitrary, it follows that for any $\delta>0$ and $i\in\mathcal{I}$,
\begin{align}
\lim_{n\to\infty}\mathbb{P}\left( {N_i(n)} \geqslant \left( \frac{1}{L} + \frac{1}{2L(L-1)} + \delta \right)n \right) = \frac{L^{2\rho}}{\delta}\lim_{n\to\infty}n^{-2\left( \rho\left(1 - 2 f\left( \frac{1}{2L(L-1)}, L \right) \right)- 1 \right)} = 0. \label{eqn:result_is_here}
\end{align}
Notice that for any $\delta>0$ and $i\in\mathcal{I}$,
\begin{align*}
\mathbb{P}\left( N_i(n) + \sum_{j\in{[K]}\backslash\mathcal{I}}N_j(n) \leqslant \left( \frac{1}{2L} - (L-1)\delta\right)n \right) &= \mathbb{P}\left( \sum_{j\in\mathcal{I}\backslash\{i\}}N_j(n) \geqslant \left(L-1\right)\left( \frac{1}{L} + \frac{1}{2L(L-1)} + \delta \right) n \right) \\
&\leqslant \sum_{j\in\mathcal{I}\backslash\{i\}}\mathbb{P}\left( N_j(n) \geqslant \left( \frac{1}{L} + \frac{1}{2L(L-1)} + \delta \right) n \right),
\end{align*}
where the last inequality follows using the Union bound. Taking limits on both sides above, we conclude using \eqref{eqn:result_is_here} that for any $\delta>0$ and $i\in\mathcal{I}$,
\begin{align}
\lim_{n\to\infty}\mathbb{P}\left( N_i(n) + \sum_{j\in{[K]}\backslash\mathcal{I}}N_j(n) \leqslant \left( \frac{1}{2L} - (L-1)\delta\right)n \right) = 0. \label{eqn:result_final_here}
\end{align}
If $\mathcal{I}=[K]$, the conclusion that $N_i(n)/n > 1/(2L) = 1/(2K)$ with high probability (approaching $1$ as $n\to\infty$) for all $i\in[K]$, is immediate from \eqref{eqn:result_final_here}. If $\mathcal{I}\neq [K]$, then $\sum_{j\in[K]\backslash\mathcal{I}}\mathbb{E}\left({N_j(n)}/{n}\right) \leqslant CK\rho\left[ \left( \frac{1}{\Delta_{\min}^2} \right) \left(\frac{\log n}{n}\right) + \frac{1}{(\rho-1)n} \right]$ for some absolute constant $C>0$ follows from \cite{audibert2009exploration}, Theorem~7. Consequently if $\Delta_{\min} = \omega\left( \sqrt{\frac{\log n}{n}} \right)$, Markov's inequality implies that $\sum_{j\in[K]\backslash\mathcal{I}}N_j(n)/n = o_p(1)$. Thus, it again follows using \eqref{eqn:result_final_here} that $N_i(n)/n > 1/(2L)$ with high probability (approaching $1$ as $n\to\infty$) for all $i\in\mathcal{I}$. \hfill $\square$

\subsection{Proof of part (I) and (II)}
\label{subsec:main}

Note that the following holds for any integer $u>1$ and any arm~$i\in[K]$:
\begin{align*}
N_i(n) \leqslant u + \sum_{t=u+1}^{n}\mathbbm{1} \left\lbrace \pi_t = i,\ N_i(t-1) \geqslant u \right\rbrace,
\end{align*}
where $\pi_t\in[K]$ indicates the arm played at time $t$. In particular, the above is true also for $u= N_j(n)+\left\lceil \epsilon n \right\rceil$, where $j\in[K]\backslash\{i\}$ and $\epsilon>0$ are arbitrarily chosen. Without loss of generality, suppose that $|\mathcal{I}| \geqslant 2$ (the result is trivial for $|\mathcal{I}|=1$), and fix two arbitrary arms $i,j\in\mathcal{I}$. Then,
\begin{align*}
N_i(n) &\leqslant N_j(n) + \left\lceil \epsilon n \right\rceil +  \sum_{t=N_j(n)+\left\lceil \epsilon n \right\rceil+1}^{n}\mathbbm{1} \left\lbrace \pi_t = i,\ N_i(t-1) \geqslant N_j(n)+\left\lceil \epsilon n \right\rceil \right\rbrace \\
&\leqslant N_j(n) + \left\lceil \epsilon n \right\rceil +  \sum_{t=\left\lceil \epsilon n \right\rceil+1}^{n}\mathbbm{1} \left\lbrace \pi_t = i,\ N_i(t-1) \geqslant N_j(n) + \epsilon n \right\rbrace \\
&\leqslant N_j(n) + \left\lceil \epsilon n \right\rceil +  \sum_{t=\left\lceil \epsilon n \right\rceil+1}^{n}\mathbbm{1} \left\lbrace B_{i,N_i(t-1),t-1} \geqslant B_{j,N_j(t-1),t-1},\ N_i(t-1) \geqslant N_j(n) + \epsilon n \right\rbrace,
\end{align*}
where $B_{k,s,t} := \hat{X}_k(s) + \sqrt{(\rho\log t)/s}$ for $k\in[K]$, with $\hat{X}_k(s)$ denoting the empirical mean reward from the first $s$ plays of arm~$k$. Then,
\begin{align}
N_i(n) &\leqslant N_j(n) + \left\lceil \epsilon n \right\rceil +  \sum_{t=\left\lceil \epsilon n \right\rceil}^{n-1}\mathbbm{1} \left\lbrace B_{i,N_i(t),t} \geqslant B_{j,N_j(t),t},\ N_i(t) \geqslant N_j(n) + \epsilon n \right\rbrace \notag \\
&\leqslant N_j(n) + \left\lceil \epsilon n \right\rceil +  \sum_{t=\left\lceil \epsilon n \right\rceil}^{n-1}\mathbbm{1} \left\lbrace B_{i,N_i(t),t} \geqslant B_{j,N_j(t),t},\ N_i(t) \geqslant N_j(t) + \epsilon t \right\rbrace \notag \\
&\leqslant N_j(n) + \epsilon n + 1 + Z_n, \label{eqn:N1nho}
\end{align}
where $Z_n := \sum_{t=\left\lceil \epsilon n \right\rceil}^{n-1}\mathbbm{1} \left\lbrace B_{i,N_i(t),t} \geqslant B_{j,N_j(t),t},\ N_i(t) \geqslant N_j(t) + \epsilon t \right\rbrace$. Now,
\begin{align}
&\mathbb{E}Z_n \notag \\
=\ &\sum_{t=\left\lceil \epsilon n \right\rceil}^{n-1}\mathbb{P} \left\lbrace B_{i,N_i(t),t} \geqslant B_{j,N_j(t),t},\ N_i(t) \geqslant N_j(t) + \epsilon t \right\rbrace \notag \\
=\ &\sum_{t=\left\lceil \epsilon n \right\rceil}^{n-1}\mathbb{P} \left\lbrace \hat{X}_i\left(N_i(t)\right) - \hat{X}_j\left(N_j(t)\right) \geqslant \sqrt{\rho\log t}\left( \frac{1}{\sqrt{N_j(t)}} - \frac{1}{\sqrt{N_i(t)}} \right),\ N_i(t) \geqslant N_j(t) + \epsilon t \right\rbrace \notag \\
=\ &\sum_{t=\left\lceil \epsilon n \right\rceil}^{n-1}\mathbb{P} \left\lbrace \hat{Y}_i\left(N_i(t)\right) - \hat{Y}_j\left(N_j(t)\right) \geqslant \sqrt{\rho\log t}\left( \frac{1}{\sqrt{N_j(t)}} - \frac{1}{\sqrt{N_i(t)}} \right),\ N_i(t) \geqslant N_j(t) + \epsilon t \right\rbrace, \notag 
\end{align}
where $\hat{Y}_k(s) := {\sum_{l=1}^{s}Y_{k,l}}/{s}$ for $k\in[K]$, $s\in\mathbb{N}$, with $Y_{k,l} := X_{k,l} - \mathbb{E}X_{k,l}$ for $l\in\mathbb{N}$. The last equality above follows since $i,j\in\mathcal{I}$ and the mean rewards of arms in $\mathcal{I}$ are equal. Thus,
\begin{align}
&\mathbb{E}Z_n \notag \\
=\ &\sum_{t=\left\lceil \epsilon n \right\rceil}^{n-1}\mathbb{P} \left( \hat{Y}_i\left(N_i(t)\right) - \hat{Y}_j\left(N_j(t)\right) \geqslant \sqrt{\frac{\rho\log t}{N_i(t)}}\left( \sqrt{\frac{N_i(t)}{N_j(t)}} - 1 \right),\ N_i(t) \geqslant N_j(t) + \epsilon t \right) \notag \\ 
\leqslant\ &\sum_{t=\left\lceil \epsilon n \right\rceil}^{n-1}\mathbb{P} \left( \hat{Y}_i\left(N_i(t)\right) - \hat{Y}_j\left(N_j(t)\right) \geqslant \sqrt{\frac{\rho\log t}{t}}\left( \sqrt{1+\epsilon} - 1 \right),\ N_i(t) \geqslant N_j(t) + \epsilon t \right) \notag \\ 
\leqslant\ &\sum_{t=\left\lceil \epsilon n \right\rceil}^{n-1}\mathbb{P} \left( W_t \geqslant \sqrt{1+\epsilon} - 1 \right), \label{eqn:EZn2ho}
\end{align}
where $W_t := \sqrt{\frac{t}{\rho\log t}} \left( \frac{\sum_{l=1}^{N_i(t)}Y_{i,l}}{N_i(t)} - \frac{\sum_{l=1}^{N_j(t)}Y_{j,l}}{N_j(t)} \right)$. Now,
\begin{align}
&\left\lvert W_t \right\rvert \notag \\
\leqslant\ &\sqrt{\frac{t}{\rho\log t}} \left( \left\lvert\frac{\sum_{l=1}^{N_i(t)}Y_{i,l}}{N_i(t)}\right\rvert + \left\lvert\frac{\sum_{l=1}^{N_j(t)}Y_{j,l}}{N_j(t)}\right\rvert \right) \notag \\
=\ &\sqrt{\frac{2t}{\rho\log t}} \left( \sqrt{\frac{\log \log N_i(t)}{N_i(t)}}\left\lvert\frac{\sum_{l=1}^{N_i(t)}Y_{i,l}}{\sqrt{2N_i(t)\log \log N_i(t)}}\right\rvert + \sqrt{\frac{\log \log N_j(t)}{N_j(t)}}\left\lvert\frac{\sum_{l=1}^{N_j(t)}Y_{j,j}}{\sqrt{2N_j(t)\log \log N_j(t)}}\right\rvert \right) \notag \\
\leqslant\ &\sqrt{\frac{2t}{\rho\log t}} \left( \sqrt{\frac{\log \log t}{N_i(t)}}\left\lvert\frac{\sum_{l=1}^{N_i(t)}Y_{i,l}}{\sqrt{2N_i(t)\log \log N_i(t)}}\right\rvert + \sqrt{\frac{\log \log t}{N_j(t)}}\left\lvert\frac{\sum_{l=1}^{N_j(t)}Y_{j,l}}{\sqrt{2N_j(t)\log \log N_j(t)}}\right\rvert \right) \notag \\
=\ &\sqrt{\frac{2\log \log t}{\rho\log t}} \left( \sqrt{\frac{t}{N_i(t)}}\left\lvert\frac{\sum_{l=1}^{N_i(t)}Y_{i,l}}{\sqrt{2N_i(t)\log \log N_i(t)}}\right\rvert + \sqrt{\frac{t}{N_j(t)}}\left\lvert\frac{\sum_{l=1}^{N_j(t)}Y_{j,l}}{\sqrt{2N_j(t)\log \log N_j(t)}}\right\rvert \right). \label{eqn:Wtho}
\end{align}

\noindent We know that $N_k(t)$, for any arm~$k\in[K]$, can be lower-bounded \emph{path-wise} by a deterministic monotone increasing \emph{coercive} function of $t$, say $h(t)$. This follows as a trivial consequence of the structure of the policy, and the fact that the rewards are uniformly bounded. Therefore, we have for any arm~$k\in\mathcal{I}$ that
\begin{align}
&\left\lvert\frac{\sum_{l=1}^{N_k(t)}Y_{k,l}}{\sqrt{2N_k(t)\log \log N_k(t)}}\right\rvert \leqslant \sup_{m\geqslant h(t)} \left\lvert\frac{\sum_{l=1}^{m}Y_{k,l}}{\sqrt{2m\log \log m}}\right\rvert \notag \\
\implies &\limsup_{t\to\infty}\left\lvert\frac{\sum_{l=1}^{N_k(t)}Y_{k,l}}{\sqrt{2N_k(t)\log \log N_k(t)}}\right\rvert \leqslant \limsup_{t\to\infty}\left\lvert\frac{\sum_{l=1}^{t}Y_{k,l}}{\sqrt{2t\log \log t}}\right\rvert. \label{eqn:randoum}
\end{align}
For any $k\in\mathcal{I}$, we know that $\left\lbrace Y_{k,l} : l\in\mathbb{N} \right\rbrace$ is a collection of i.i.d. random variables with $\mathbb{E}Y_{k,1} = 0$ and $\text{Var}\left( Y_{k,1} \right) = \text{Var}\left( X_{k,1} \right) \leqslant 1$. Therefore, we conclude using the Law of the Iterated Logarithm (see Theorem~8.5.2 in \cite{durrett2019probability}) in \eqref{eqn:randoum} that 
\begin{align}
\limsup_{t\to\infty} \left\lvert\frac{\sum_{l=1}^{N_k(t)}Y_{k,l}}{\sqrt{2N_k(t)\log \log N_k(t)}}\right\rvert \leqslant 1\ \ \ \text{w.p.}\ 1\ \ \forall\ k\in\mathcal{I}. \label{eqn:LILho}
\end{align}
Using \eqref{eqn:Wtho}, \eqref{eqn:LILho}, and the meta-result from the preamble in \ref{subsec:preamble} that $N_k(t)/t > 1/\left( 2\left\lvert \mathcal{I} \right\rvert \right)$ with high probability (approaching $1$ as $t\to\infty$) for any arm~$k\in\mathcal{I}$, we conclude that
\begin{align}
W_t \xrightarrow{p} 0\ \ \ \text{as}\ t\to\infty. \label{eqn:Wt_as_0ho}
\end{align}
Now,
\begin{align*}
\mathbb{P}\left( \frac{N_i(n)-N_j(n)}{n} \geqslant 2\epsilon\right) \underset{\mathrm{(\dag)}}{\leqslant} \mathbb{P}(1+Z_n \geqslant \epsilon n) &\underset{\mathrm{(\ddag)}}{\leqslant} \frac{1+\mathbb{E}Z_n}{\epsilon n} \underset{\mathrm{(\star)}}{\leqslant} \frac{1}{\epsilon n} + \frac{1}{\epsilon n}\sum_{t=\left\lceil \epsilon n \right\rceil}^{n-1}\mathbb{P} \left( W_t > \sqrt{1+\epsilon} - 1 \right),
\end{align*}
where $(\dag)$ follows using \eqref{eqn:N1nho}, $(\ddag)$ using Markov's inequality, and $(\star)$ from \eqref{eqn:EZn2ho}. Therefore,
\begin{align}
\mathbb{P}\left( \frac{N_i(n)-N_j(n)}{n} \geqslant 2\epsilon\right) \leqslant \frac{1}{\epsilon n} + \left(\frac{1 - \epsilon}{\epsilon}\right)\sup_{t \geqslant \epsilon n}\mathbb{P} \left( W_t > \sqrt{1+\epsilon} - 1 \right). \label{eqn:tail_probho}
\end{align}
Since $\epsilon>0$ is arbitrary, we conclude using \eqref{eqn:Wt_as_0ho} and \eqref{eqn:tail_probho} that for any $\epsilon>0$,
\begin{align}
\lim_{n\to\infty}\mathbb{P}\left( \frac{N_i(n)-N_j(n)}{n} \geqslant 2\epsilon\right) = 0. \label{eqn:hoho4}
\end{align}
Our proof is symmetric w.r.t. the labels $i,j$, therefore, an identical result holds also with the labels interchanged in \eqref{eqn:hoho4}. Thus, we have $N_i(n)/n - N_j(n)/n\xrightarrow{p}0$. Since $i,j$ are arbitrary in $\mathcal{I}$, the aforementioned convergence holds for any pair of arms in $\mathcal{I}$. Now if $\mathcal{I}=[K]$, we are done. If $\mathcal{I}\neq[K]$, then $\sum_{i\in[K]\backslash\mathcal{I}}\mathbb{E}\left(N_i(n)/n\right) \leqslant CK\rho\left[ \left( \frac{1}{\Delta_{\min}^2} \right) \left(\frac{\log n}{n}\right) + \frac{1}{(\rho-1)n} \right]$ for some absolute constant $C>0$ follows from Theorem~7 in \cite{audibert2009exploration}. Consequently if $\Delta_{\min} = \omega\left( \sqrt{\frac{\log n}{n}} \right)$, it would follow from Markov's inequality that $\sum_{i\in[K]\backslash\mathcal{I}}N_i(n)/n = o_p(1)$. Thus, any arm~$i\in\mathcal{I}$ must satisfy $N_i(n)/n \xrightarrow{p} 1/|\mathcal{I}|$. \hfill $\square$

\section{Proof of Theorem~\ref{thm:TS-BP-00}}
\label{appendix:thm5}

\subsection{Proof of part (I)}

Let $\theta_k,\tilde{\theta}_k$ be Beta$(1, k+1)$-distributed, with $\theta_k,\tilde{\theta}_l$ independent $\forall\ k,l\in\mathbb{N}\cup\{0\}$. In the two-armed bandit with deterministic $0$ rewards, at any time $n+1$, the probability of playing arm~$1$ conditioned on the entire history up to that point, is given by $\mathbb{P}\left( \pi_{n+1} = 1\ |\ \mathcal{F}_n\right) = \mathbb{P}\left( \theta_{N_1(n)} > \tilde{\theta}_{N_2(n)}\ |\ \mathcal{F}_n \right) = \frac{n - N_1(n)+1}{n+2}$ (using Fact~\eqref{fact:1}). Since the arms are identical, and $N_1(n)+N_2(n)=n$, we must have $\mathbb{E}\left( N_1(n)/n \right) = 1/2\ \forall\ n\in\mathbb{N}$ by symmetry. Define $Z_n := N_1(n)/n$. Then, $Z_n$ evolves according to the following Markovian rule:
\begin{align*}
Z_{n+1} = \left(\frac{n}{n+1}\right)Z_n + \frac{Y\left( Z_n, n, \xi_n \right)}{n+1},
\end{align*}
where $\{\xi_n\}$ is an independent noise process that is such that $Y\left( Z_n, n, \xi_n \right) | Z_n$ is distributed as Bernoulli$\left( \frac{n(1-Z_n)+1}{n+2} \right)$. Note that $Y\left( \cdot, \cdot, \cdot \right)\in\{0,1\}$. Then,
\begin{align*}
Z_{n+1}^2 &= \left(\frac{n}{n+1}\right)^2Z_n^2 + \left(\frac{Y\left( Z_n, n, \xi_n \right)}{n+1}\right)^2 + \frac{2nZ_nY\left( Z_n, n, \xi_n \right)}{(n+1)^2} \\
&= \left(\frac{n}{n+1}\right)^2Z_n^2 + \frac{Y\left( Z_n, n, \xi_n \right)}{(n+1)^2} + \frac{2nZ_nY\left( Z_n, n, \xi_n \right)}{(n+1)^2}.
\end{align*}
Solving the recursion for $Z_{n+1}^2$, we obtain
\begin{align*}
Z_{n+1}^2 = \frac{Z_1^2 + \sum_{t=1}^{n}\left[ Y\left( Z_t, t, \xi_t \right) + 2tZ_tY\left( Z_t, t, \xi_t \right) \right] }{(n+1)^2} = \frac{Z_1 + \sum_{t=1}^{n}\left[ Y\left( Z_t, t, \xi_t \right) + 2tZ_tY\left( Z_t, t, \xi_t \right) \right] }{(n+1)^2},
\end{align*}
where the last equality follows since $Z_1\in\{0,1\}$. Taking expectations and using the fact that $\mathbb{E}Z_t=1/2\ \forall\ t\in\mathbb{N}$, yields
\begin{align*}
\mathbb{E}Z_{n+1}^2 = \frac{\frac{1}{2} + \sum_{t=1}^{n} \left( \frac{1}{2} + 2t\mathbb{E}\left[ \frac{tZ_t(1-Z_t)+Z_t}{t+2} \right] \right) }{(n+1)^2}.
\end{align*}
Using $Z_t(1-Z_t) \leqslant 1/4$, we get the relation
\begin{align*}
\mathbb{E}Z_{n+1}^2 &\leqslant \frac{1 + \sum_{t=1}^{n} \left( 1 + t\mathbb{E}\left[ \frac{t + 4Z_t}{t+2} \right] \right) }{2(n+1)^2} = \frac{n+1 + \sum_{t=1}^{n} t}{2(n+1)^2} = \frac{n+2}{4(n+1)}.
\end{align*}

\noindent Thus, $\text{Var}\left( \frac{N_1(n)}{n} \right) = \text{Var}\left( Z_n \right) \leqslant \frac{n+1}{4n} - \frac{1}{4} = \frac{1}{4n}$. Since $\mathbb{E}\left( \frac{N_1(n)}{n} \right) = \frac{1}{2}$, we conclude using Chebyshev's inequality that $\frac{N_1(n)}{n}\to\frac{1}{2}$ in probability as $n\to\infty$. \hfill $\square$

\subsection{Proof of part (II)}

Our proof of this part is essentially pivoted on showing the stronger result that $\mathbb{P}\left( N_1(n) = m \right) = \frac{1}{n+1}$ for any $m\in\{0,...,n\}$ and $n\in\mathbb{N}$. To this end, for an arbitrary $m$ in said interval, let $\mathrm{S}_{m}$ be the set of sample-paths of length $n$ such that $N_1(n)=m$ on each sample-path $\mathrm{s}_m\in\mathrm{S}_{m}$. Clearly, $\left\lvert \mathrm{S}_{m} \right\rvert = {{n}\choose{m}}$. Let $i\left( \mathrm{s}_m, t \right)\in\{1,2\}$ denote the index of the arm pulled at time $t\in\{1,...,n\}$ on $\mathrm{s}_m$, and let $\tilde{N}_j(t)$ denote the number of pulls of arm~$j\in\{1,2\}$ up to (and including) time $t$ on $\mathrm{s}_m$ (with $\tilde{N}_1(0)=\tilde{N}_2(0):=0$). Note that $i\left( \mathrm{s}_m, t \right)$, $\tilde{N}_1(t)$ and $\tilde{N}_2(t)$ are deterministic for all $t\in\{1,...,n\}$, once $\mathrm{s}_m$ is fixed. Let $\theta_k,\tilde{\theta}_k$ be Beta$(k+1,1)$-distributed, with $\theta_k,\tilde{\theta}_l$ independent $\forall\ k,l\in\mathbb{N}\cup\{0\}$. It then follows that
\begin{align*}
\mathbb{P}\left( N_1(n) = m \right) &= \sum_{\mathrm{s}_m\in\mathrm{S}_{m}}\prod_{t=1}^{n} \mathbb{P}\left( \theta_{\tilde{N}_{i\left( \mathrm{s}_m, t \right)}(t-1)} > \tilde{\theta}_{\tilde{N}_{\{1,2\}\backslash i\left( \mathrm{s}_m, t \right)}(t-1)} \right) \\
&= \sum_{\mathrm{s}_m\in\mathrm{S}_{m}} \prod_{t=1}^{n}\left( \frac{\tilde{N}_{i\left( \mathrm{s}_m, t \right)}(t-1)+1}{t+1} \right) & \mbox{(using Fact~\eqref{fact:2})} \\
&= \frac{1}{(n+1)!} \sum_{\mathrm{s}_m\in\mathrm{S}_{m}} \prod_{t=1}^{n}\left( \tilde{N}_{i\left( \mathrm{s}_m, t \right)}(t-1)+1 \right) \\
&= \frac{1}{(n+1)!} \sum_{\mathrm{s}_m\in\mathrm{S}_{m}} m!(n-m)!,
\end{align*}
where the last equality follows since $\tilde{N}_1(n)=m$, $\tilde{N}_2(n)=n-m$, $\tilde{N}_1(0)=\tilde{N}_2(0)=0$ on $\mathrm{s}_m$. Therefore, we have for all $m\in\{0,...,n\}$ that
\begin{align}
\mathbb{P}\left( N_1(n) = m \right) = \frac{{{n}\choose{m}}m!(n-m)!}{(n+1)!} = \frac{n!}{(n+1)!} = \frac{1}{n+1}. \label{eqn:lastyo}
\end{align}
This, in fact, proves a stronger result that $N_1(n)/n$ is uniformly distributed on $\left\lbrace 0, \frac{1}{n}, \frac{2}{n}, ..., \frac{n-1}{n}, 1 \right\rbrace$ for any $n\in\mathbb{N}$. The desired result now follows as a corollary in the limit $n\to\infty$; for an arbitrary $x\in[0,1]$, consider
\begin{align*}
\mathbb{P}\left( \frac{N_1(n)}{n} \leqslant x \right) = \sum_{m=0}^{\left\lfloor xn \right\rfloor}\mathbb{P}\left( N_1(n) = m \right) = \frac{\left\lfloor xn \right\rfloor + 1}{n+1},
\end{align*}
where the last equality follows using \eqref{eqn:lastyo}. Thus, we have $\lim_{n\to\infty} \mathbb{P}\left( \frac{N_1(n)}{n} \leqslant x \right) = x$ for any $x\in[0,1]$, i.e., $\frac{N_1(n)}{n}$ converges in law to the Uniform distribution on $[0,1]$. \hfill $\square$

\section{Proof of Theorem~\ref{cor}}
\label{sec:cor}

We essentially need to bound the growth rate of $R^{\pi}_n$ under the policy $\pi$ given by Algorithm~\ref{alg:UCB} with $\rho>1$, in three (exhaustive) regimes, viz., (i) $\Delta = o\left( \sqrt{\frac{\log n}{n}} \right)$ (``small gap''), (ii) $\Delta = \omega\left( \sqrt{\frac{\log n}{n}} \right)$ (``large gap''), and (iii) $\Delta = \Theta\left( \sqrt{\frac{\log n}{n}} \right)$ (``moderate gap''). We handle the three cases below separately.

\subsection{The ``small gap'' regime}

Here, we have 
\begin{align*}
\frac{\mathbb{E}R^{\pi}_n}{\sqrt{n\log n}} \leqslant \frac{\Delta n}{\sqrt{n\log n}} = \sqrt{\frac{\Delta^2 n}{\log n}}.
\end{align*}
Since $\Delta = o\left( \sqrt{\frac{\log n}{n}} \right)$, it follows that $\mathbb{E}R_n^\pi = o\left( \sqrt{n\log n} \right)$. Therefore, we conclude using Markov's inequality that $R_n^\pi = o_p\left( \sqrt{n\log n} \right)$ whenever $\Delta = o\left( \sqrt{\frac{\log n}{n}} \right)$.

\subsection{The ``large gap'' regime}

In this regime, we have
\begin{align*}
\frac{\mathbb{E}R^{\pi}_n}{\sqrt{n\log n}} \leqslant \frac{C\rho\left(\frac{\log n}{\Delta} + \frac{\Delta}{\rho-1}\right)}{\sqrt{n\log n}},
\end{align*}
where $C$ is some absolute constant (follows from \cite{audibert2009exploration}, Theorem~7). Since $\Delta = \omega\left( \sqrt{\frac{\log n}{n}} \right)$ and $\Delta\leqslant 1$ (rewards bounded in $[0,1]$), it follows that $\mathbb{E}R_n^\pi = o\left( \sqrt{n\log n} \right)$. Thus, we again conclude using Markov's inequality that $R_n^\pi = o_p\left( \sqrt{n\log n} \right)$ whenever $\Delta = \omega\left( \sqrt{\frac{\log n}{n}} \right)$.

\subsection{The ``moderate gap'' regime}

Since $\Delta = \Theta\left( \sqrt{\frac{\log n}{n}} \right)$, there exists some $\theta\in\mathbb{R}_+$ and a diverging sequence of natural numbers $\left\lbrace n_k \right\rbrace_{k\in\mathbb{N}}$ such that $\Delta$ scales with the horizon of play $n_k$ along this sequence as $\Delta = \sqrt{\frac{\theta\log n_k}{n_k}}$. Without loss of generality, suppose that arm~1 is optimal, i.e., $\mu_1 \geqslant \mu_2$. We then have
\begin{align*}
\frac{R^\pi_{n_k}}{\sqrt{n_k\log n_k}} &= \frac{\mu_1n_k - \sum_{j=1}^{N_1\left(n_k\right)}X_{1,j} - \sum_{j=1}^{N_2\left(n_k\right)}X_{2,j}}{\sqrt{n_k\log n_k}} & \mbox{(using \eqref{eqn:stochastic_regret})} \\
&= \frac{\mu_1n_k - \mu_1N_1\left(n_k\right) - \mu_2N_2\left(n_k\right) - \sum_{j=1}^{N_1\left(n_k\right)}Y_{1,j} - \sum_{j=1}^{N_2\left(n_k\right)}Y_{2,j}}{\sqrt{n_k\log n_k}},
\end{align*}
where $Y_{i,j} := X_{i,j} - \mu_i$, $i\in\{1,2\}$, $j\in\mathbb{N}$. Therefore,
\begin{align}
\frac{R^\pi_{n_k}}{\sqrt{n_k\log n_k}} &= \frac{\Delta N_2\left(n_k\right) - \sum_{j=1}^{N_1\left(n_k\right)}Y_{1,j} - \sum_{j=1}^{N_2\left(n_k\right)}Y_{2,j}}{\sqrt{n_k\log n_k}} \notag \\
&= \Delta\sqrt{\frac{n_k}{\log n_k}}\left(\frac{N_2\left(n_k\right)}{n_k}\right) - \left( \frac{\sum_{j=1}^{N_1\left(n_k\right)}Y_{1,j} + \sum_{j=1}^{N_2\left(n_k\right)}Y_{2,j}}{\sqrt{n_k\log n_k}} \right) \notag \\
&= \sqrt{\theta}\left(\frac{N_2\left(n_k\right)}{n_k}\right) - \left( \frac{\sum_{j=1}^{N_1\left(n_k\right)}Y_{1,j} + \sum_{j=1}^{N_2\left(n_k\right)}Y_{2,j}}{\sqrt{n_k\log n_k}} \right). \label{eqn:c1}
\end{align}
Consider the summation terms above. We have
\begin{align}
\left\lvert \frac{\sum_{j=1}^{N_1\left(n_k\right)}Y_{1,j} + \sum_{j=1}^{N_2\left(n_k\right)}Y_{2,j}}{\sqrt{n_k\log n_k}} \right\rvert &\leqslant \left\lvert \frac{\sum_{j=1}^{N_1\left(n_k\right)}Y_{1,j}}{\sqrt{n_k\log n_k}} \right\rvert + \left\lvert \frac{\sum_{j=1}^{N_2\left(n_k\right)}Y_{2,j}}{\sqrt{n_k\log n_k}} \right\rvert \notag \\
&\leqslant \left\lvert \frac{\sum_{j=1}^{N_1\left(n_k\right)}Y_{1,j}}{\sqrt{N_1\left(n_k\right)\log N_1\left(n_k\right)}} \right\rvert + \left\lvert \frac{\sum_{j=1}^{N_2\left(n_k\right)}Y_{2,j}}{\sqrt{N_2\left(n_k\right)\log N_2\left(n_k\right)}} \right\rvert. \label{eqn:c2}
\end{align}
Since $N_i\left(n_k\right)$, for each $i\in\{1,2\}$, can be lower-bounded \emph{path-wise} by a deterministic monotone increasing coercive function of $k$, say $f\left(k\right)$, we have 
\begin{align}
\left\lvert \frac{\sum_{j=1}^{N_i\left(n_k\right)}Y_{i,j}}{\sqrt{N_i\left(n_k\right)\log N_i\left(n_k\right)}} \right\rvert \leqslant \sup_{m \geqslant f\left(k\right)} \left\lvert \frac{\sum_{j=1}^{m}Y_{i,j}}{\sqrt{m\log m}} \right\rvert\ \ \ \ \forall\ i\in\{1,2\}. \label{eqn:c3}
\end{align}
Since $Y_{i,j}$'s are independent, zero-mean, bounded random variables, it follows from the Law of the Iterated Logarithm (see \cite{durrett2019probability}, Theorem~8.5.2) that
\begin{align}
\sup_{m \geqslant f\left(k\right)} \left\lvert \frac{\sum_{j=1}^{m}Y_{i,j}}{\sqrt{m\log m}} \right\rvert \xrightarrow[k\to\infty]{\text{w.p.}\ 1} 0\ \ \ \ \forall\ i\in\{1,2\}. \label{eqn:c4}
\end{align} 
Combining \eqref{eqn:c1}, \eqref{eqn:c2}, \eqref{eqn:c3} and \eqref{eqn:c4}, we conclude
\begin{align*}
\frac{R^\pi_{n_k}}{\sqrt{n_k\log n_k}} = \sqrt{\theta}\left(\frac{N_2\left(n_k\right)}{n_k}\right) + o_p(1).
\end{align*}
From Theorem~\ref{thm:rates}, we know that when $\Delta \sim \sqrt{\frac{\theta\log n_k}{n_k}}$, $\frac{N_2\left(n_k\right)}{n_k} \xrightarrow[k\to\infty]{p}1-\lambda_\rho^\ast(\theta)$. Thus, it follows that
\begin{align*}
\frac{R^\pi_{n_k}}{\sqrt{n_k\log n_k}} \xrightarrow[k\to\infty]{p} \sqrt{\theta}\left( 1-\lambda_\rho^\ast(\theta) \right) = h_\rho(\theta).
\end{align*}
Since $\theta\in\mathbb{R}_+$ is arbitrary, the worst-case regret in the $\Delta = \Theta\left( \sqrt{\frac{\log n}{n}} \right)$ regime corresponds to the choice of $\theta$ given by $\theta_\rho^\ast = \arg\max_{\theta\geqslant 0} h_\rho(\theta)$. Since we already know that $R_n^\pi = o_p \left( \sqrt{n\log n} \right)$ in the other two regimes (``small'' and ``large'' gaps), it must be that the $\theta_\rho^\ast$ so obtained indeed corresponds to the global (in $\Delta$) worst-case regret of Algorithm~\ref{alg:UCB}. \hfill $\square$

\section{Proof of Theorem~\ref{thm:diffusion}}
\label{sec:diffusion}

\textbf{Notation.} Let $\mathcal{C}$ be the space of continuous functions $[0,1]\mapsto\mathbb{R}^2$, endowed with the uniform metric. Let $\mathcal{D}$ be the space of right-continuous functions with left limits, mapping $[0,1]\mapsto\mathbb{R}^2$, and endowed with the Skorohod metric (see \cite{billingsley2013convergence}, Chapters 2 and 3, for an overview). Let $\mathcal{D}_0$ be the set of elements of $\mathcal{D}$ of the form $\left(\phi_1,\phi_2\right)$, where $\phi_i$ is a non-decreasing real-valued function satisfying $0 \leqslant \phi_i(t) \leqslant 1$ for $i\in\{1,2\}$ and $t\in[0,1]$. For $t\in[0,1]$, denote the identity map by $\mathfrak{e}(t) := t$. \\

For $i\in\{1,2\}$ and $t\in[0,1]$, define $\Psi_{i,n}(t) := \frac{\sum_{j=1}^{\left\lfloor nt\right\rfloor}X_{i,j} - \mu nt}{\sqrt{n}}$. Then, $\left( {\Psi}_{1,n}, {\Psi}_{2,n} \right)\in\mathcal{D}$. Also for $i\in\{1,2\}$ and $t\in[0,1]$, define $W_{i}(t) := \theta_i t + \sigma_iB^{\prime}_i(t)$, where $B^{\prime}_1$ and $B^{\prime}_2$ are independent standard Brownian motions in $\mathbb{R}$. Note that $\mathbb{P}\left( W_i \in \mathcal{C} \right) = 1$ for $i\in\{1,2\}$. Since $\left( X_{i,j} \right)_{i\in\{1,2\}, j\in\mathbb{N}}$'s are independent random variables (i.i.d. within and independent across sequences), we know from Donsker's Theorem (see \cite{billingsley2013convergence}, Section 14, for details) that as $n\to\infty$,
\begin{align*}
\left( {\Psi}_{1,n}, {\Psi}_{2,n} \right) \Rightarrow \left( W_1, W_2 \right)\ \text{in}\ \mathcal{D}.
\end{align*}

For $i\in\{1,2\}$ and $t\in[0,1]$, define $\phi_{i,n}(t) := \frac{N_i(\left\lfloor nt\right\rfloor)}{n}$. Thus, $\left( {\phi}_{1,n}, {\phi}_{2,n} \right)\in\mathcal{D}_0$, and it follows from the result for the ``small gap'' regime in Theorem~\ref{thm:rates} that as $n\to\infty$,
\begin{align*}
\left( {\phi}_{1,n}, {\phi}_{2,n} \right) \xrightarrow{p} \left( \frac{\mathfrak{e}}{2}, \frac{\mathfrak{e}}{2} \right)\ \text{in}\ \mathcal{D}_0.
\end{align*}
Thus, we have convergence in the product space (see \cite{billingsley2013convergence}, Theorem 3.9), i.e., as $n\to\infty$,
\begin{align*}
\left( {\Psi}_{1,n}, {\Psi}_{2,n}, {\phi}_{1,n}, {\phi}_{2,n} \right) \Rightarrow \left( W_1, W_2, \frac{\mathfrak{e}}{2}, \frac{\mathfrak{e}}{2} \right)\ \text{in}\ \mathcal{D}\times\mathcal{D}_0.
\end{align*}

For $i\in\{1,2\}$ and $t\in[0,1]$, define the composition $\left({\Psi}_{i,n}\circ{\phi}_{i,n}\right)(t) := {\Psi}_{i,n}\left( {\phi}_{i,n}(t) \right)$, and $\left(W_i\circ \frac{\mathfrak{e}}{2}\right)(t) := W_i\left( \frac{\mathfrak{e}(t)}{2} \right) = W_i \left( \frac{t}{2} \right)$. Since $W_1,W_2,\mathfrak{e}\in\mathcal{C}$ w.p.~$1$, it follows from the \emph{random time-change lemma} (see \cite{billingsley2013convergence}, Section 14, for details) that as $n\to\infty$
\begin{align*}
\left( {\Psi}_{1,n}\circ{\phi}_{1,n}, {\Psi}_{2,n}\circ{\phi}_{2,n} \right) \Rightarrow \left( W_1\circ\frac{\mathfrak{e}}{2}, W_2\circ\frac{\mathfrak{e}}{2} \right)\ \text{in}\ \mathcal{D}.
\end{align*}

The stated assertion on cumulative rewards now follows by recognizing for $i\in\{1,2\}$ and $t\in[0,1]$ that $\left({\Psi}_{i,n}\circ{\phi}_{i,n}\right)(t) = \frac{\tilde{S}_{i,\left\lfloor nt \right\rfloor}}{\sqrt{n}}$, and defining $B_i(t) := \sqrt{2}B^{\prime}_i\left(\frac{t}{2}\right)$. To prove the assertion on regret, assume without loss of generality that arm~1 is optimal, i.e., $\theta_1\geqslant \theta_2$. Then, the result follows after a direct application of the Continuous Mapping Theorem (see \cite{billingsley2013convergence}, Theorem~2.7), to wit,
\begin{align*}
R^\pi_{\left\lfloor nt \right\rfloor} = \left(\mu + \frac{\theta_1}{\sqrt{n}}\right)\left\lfloor nt \right\rfloor - S_{1,\left\lfloor nt \right\rfloor} - S_{2,\left\lfloor nt \right\rfloor} = \frac{\theta_1\left\lfloor nt \right\rfloor}{\sqrt{n}} - \left( \tilde{S}_{1,\left\lfloor nt \right\rfloor} + \tilde{S}_{2,\left\lfloor nt \right\rfloor} \right),
\end{align*}
and therefore as $n\to\infty$,
\begin{align*}
\left(\frac{R^\pi_{\left\lfloor nt \right\rfloor}}{\sqrt{n}}\right)_{t\in[0,1]} \Rightarrow \left( \theta_1t - \left( \left(\frac{\theta_1+\theta_2}{2}\right)t + \frac{\sigma_1}{\sqrt{2}}B_1(t) + \frac{\sigma_2}{\sqrt{2}}B_2(t) \right) \right)_{t\in[0,1]} = \left(\frac{\Delta_0t}{2} + \sqrt{\frac{\sigma_1^2+\sigma_2^2}{2}}\tilde{B}(t)\right)_{t\in[0,1]},
\end{align*}
where $\tilde{B}(t) := -\sqrt{\frac{\sigma_1^2}{\sigma_1^2+\sigma_2^2}}B_1(t) - \sqrt{\frac{\sigma_2^2}{\sigma_1^2+\sigma_2^2}}B_2(t)$. \hfill $\square$

\section{Proof of Fact~\ref{fact:1} and Fact~\ref{fact:2}}
\label{facts}

\subsection{Fact~\ref{fact:1}}
\label{fact:1pf}

Since $\theta_k$ is Beta$(1,k+1)$-distributed, its PDF, say $f_k(\cdot)$, is given by 
\begin{align}
f_k(x) = (k+1)(1-x)^k;\ \ \ x\in[0,1]. \label{eqn:k_density_1}
\end{align}
Thus,
\begin{align*}
\mathbb{P}\left( \theta_k > \tilde{\theta}_l \right) &= \int_{0}^{1}\left( \int_{y}^{1} f_k(x) dx \right) f_l(y)dy \\
&= \int_{0}^{1}\left( \int_{y}^{1} (k+1)(1-x)^k dx \right) (l+1)(1-y)^ldy & \mbox{(using \eqref{eqn:k_density_1})} \\
&= \int_{0}^{1}(l+1)(1-y)^{k+l+1} dy \\
&= \frac{l+1}{k+l+2}.
\end{align*} \hfill $\square$

\subsection{Fact~\ref{fact:2}}
\label{fact:2pf}

Since $\theta_k$ is Beta$(k+1,1)$-distributed, its PDF, say $f_k(\cdot)$, is given by 
\begin{align}
f_k(x) = (k+1)x^k;\ \ \ x\in[0,1]. \label{eqn:k_density_2}
\end{align}
Thus,
\begin{align*}
\mathbb{P}\left( \theta_k > \tilde{\theta}_l \right) &= \int_{0}^{1}\left( \int_{y}^{1} f_k(x) dx \right) f_l(y)dy \\
&= \int_{0}^{1}\left( \int_{y}^{1} (k+1)x^k dx \right) (l+1)y^ldy & \mbox{(using \eqref{eqn:k_density_2})} \\
&= \int_{0}^{1}(l+1)\left(1-y^{k+1}\right)y^l dy \\
&= 1 - \frac{l+1}{k+l+2} \\
&= \frac{k+1}{k+l+2}.
\end{align*} \hfill $\square$

\bibliographystyle{agsm} 
\bibliography{ref}

\end{document}